%% file: main.tex
\PassOptionsToPackage{table}{xcolor}
\documentclass{pinepaper}

\usepackage{amssymb,tabularx,pifont}
\usepackage{algorithm}
\usepackage{algpseudocode}
\usepackage{multirow}
\usepackage{subcaption}
\usepackage{xspace}
\usepackage{xurl}
\usepackage{float}
\usepackage{placeins}
\usepackage{needspace}
\usepackage[capitalize, noabbrev]{cleveref}

\definecolor{tablerow}{HTML}{ECF4F5}
\newcommand{\best}[1]{\textbf{#1}}
\newcommand{\second}[1]{\underline{#1}}
\definecolor{missingred}{rgb}{0.80,0.08,0.08}
\definecolor{unexpcolor}{rgb}{0.55,0.05,0.65}

\newcommand{\method}{\textnormal{\textsc{CauSyncVLA}}\xspace}
\newcommand{\wm}{\mathcal{W}_\theta}
\newcommand{\cache}{\mathcal{Q}}
\newcommand{\hist}{\mathcal{H}}
\DeclareMathAlphabet{\amsbb}{U}{msb}{m}{n}

\newcommand{\MTFiveStockMean}{60.01}
\newcommand{\MTFiveStockTail}{60.20}
\newcommand{\MTFiveStockRate}{16.66}
\newcommand{\MTFiveStockSupply}{83.32}

\newcommand{\MTFiveStockOneMean}{36.00}
\newcommand{\MTFiveStockOneTail}{36.32}
\newcommand{\MTFiveStockOneRate}{27.78}
\newcommand{\MTFiveStockOneSupply}{138.88}

\newcommand{\TenSeven}{91.20}
\newcommand{\OneSeven}{90.80}

\newcommand{\SyncTFWSSuccess}{93.00}
\newcommand{\SyncTFWSMean}{35.54}
\newcommand{\SyncTFWSPninetyfive}{35.94}
\newcommand{\SyncTFWSRate}{28.14}
\newcommand{\SyncTFWSSupply}{140.70}

\newcommand{\FinalSuccess}{93.80}

\newcommand{\FinalStaleError}{0.10355}

\newcommand{\FinalStaleKV}{0.42430}

\newcommand{\FinalSharedMean}{\FinalPending}
\newcommand{\FinalSharedPninetyfive}{\FinalPending}
\newcommand{\FinalSharedRate}{\FinalPending}
\newcommand{\FinalSharedSupply}{\FinalPending}

\newcommand{\FinalSharedGain}{\FinalPending}

\renewcommand{\FinalSharedMean}{11.91}
\renewcommand{\FinalSharedRate}{83.96}
\renewcommand{\FinalSharedSupply}{419.82}
\renewcommand{\FinalSharedGain}{5.04}
\renewcommand{\FinalSharedPninetyfive}{13.38}

\newcommand{\GrootMs}{\hlpurple{--}}
\newcommand{\GrootRate}{\hlpurple{--}}
\newcommand{\GrootSuccess}{\hlpurple{--}}

\newcommand{\OftMs}{\hlpurple{--}}
\newcommand{\OftRate}{\hlpurple{--}}
\newcommand{\OftSuccess}{\hlpurple{--}}

\newcommand{\GrootPninetyfive}{\hlpurple{--}}
\newcommand{\GrootSupply}{\hlpurple{--}}
\newcommand{\OftPninetyfive}{\hlpurple{--}}
\newcommand{\OftSupply}{\hlpurple{--}}

\renewcommand{\GrootMs}{59.37}
\renewcommand{\GrootPninetyfive}{59.90}
\renewcommand{\GrootRate}{16.84}
\renewcommand{\GrootSupply}{84.21}
\renewcommand{\OftSuccess}{89.20}
\renewcommand{\OftMs}{62.71}
\renewcommand{\OftPninetyfive}{62.91}
\renewcommand{\OftRate}{15.95}
\renewcommand{\OftSupply}{79.73}
\renewcommand{\GrootSuccess}{85.80}

\newcommand{\kvexampleimages}[1]{%
\captionsetup[subfigure]{font={scriptsize},skip=2pt}
\makebox[\textwidth][l]{\footnotesize\textbf{External camera}}\par\vspace{2pt}
\begin{subfigure}[t]{.19\textwidth}\centering
\includegraphics[width=\linewidth]{figures/kv/pair#1_front_anchor.pdf}
\caption{Anchor image.}
\end{subfigure}\hfill
\begin{subfigure}[t]{.19\textwidth}\centering
\includegraphics[width=\linewidth]{figures/kv/pair#1_front_real_anchor.pdf}
\caption{KV anchor.}
\end{subfigure}\hfill
\begin{subfigure}[t]{.19\textwidth}\centering
\includegraphics[width=\linewidth]{figures/kv/pair#1_front_predicted.pdf}
\caption{Predicted KV memory.}
\end{subfigure}\hfill
\begin{subfigure}[t]{.19\textwidth}\centering
\includegraphics[width=\linewidth]{figures/kv/pair#1_front_real_target.pdf}
\caption{Fresh KV memory.}
\end{subfigure}\hfill
\begin{subfigure}[t]{.19\textwidth}\centering
\includegraphics[width=\linewidth]{figures/kv/pair#1_front_target.pdf}
\caption{Target image.}
\end{subfigure}\par\vspace{6pt}
\makebox[\textwidth][l]{\footnotesize\textbf{Wrist camera}}\par\vspace{2pt}
\begin{subfigure}[t]{.19\textwidth}\centering
\includegraphics[width=\linewidth]{figures/kv/pair#1_wrist_anchor.pdf}
\caption{Anchor image.}
\end{subfigure}\hfill
\begin{subfigure}[t]{.19\textwidth}\centering
\includegraphics[width=\linewidth]{figures/kv/pair#1_wrist_real_anchor.pdf}
\caption{KV anchor.}
\end{subfigure}\hfill
\begin{subfigure}[t]{.19\textwidth}\centering
\includegraphics[width=\linewidth]{figures/kv/pair#1_wrist_predicted.pdf}
\caption{Predicted KV memory.}
\end{subfigure}\hfill
\begin{subfigure}[t]{.19\textwidth}\centering
\includegraphics[width=\linewidth]{figures/kv/pair#1_wrist_real_target.pdf}
\caption{Fresh KV memory.}
\end{subfigure}\hfill
\begin{subfigure}[t]{.19\textwidth}\centering
\includegraphics[width=\linewidth]{figures/kv/pair#1_wrist_target.pdf}
\caption{Target image.}
\end{subfigure}\par\smallskip
}

\title{\textsc{CauSyncVLA}: Causal Latent World Modeling for Asynchronous Robot Control}

\author[1]{Xu~Pan}
\author[2]{Chubin~Zhang}
\author[3]{Xingrui~Yu}
\author[1]{Runmin~Lan}
\author[1]{Zihao~Wang}
\author[1]{Wenkai~Guo}
\author[1]{Bofang~Jia}
\author[2]{Bo~An}
\author[1,4]{Ziwei~Wang\textsuperscript{\textdagger}}

\affil[1]{PINE Lab, School of Electrical and Electronic Engineering, Nanyang Technological University, Singapore}
\affil[2]{College of Computing and Data Science, Nanyang Technological University, Singapore}
\affil[3]{CFAR and IHPC, Agency for Science, Technology and Research (A*STAR), Singapore}
\affil[4]{Embodied AI Center, Singapore}

\date{\small \textsuperscript{\textdagger}Corresponding author.}
\papervenue{Preprint}
\paperdate{\today}
\projectpage{https://xupan.top/projects/causyncvla}
\correspondence{\href{mailto:ziwei.wang@ntu.edu.sg}{ziwei.wang@ntu.edu.sg}}

\begin{document}
\makepinetitle

\begin{pineabstract}
Vision--Language--Action (VLA) policies must react while visual inference is still running.
Overlapping computation and execution reduces waiting, yet leaves perceptual memory stale and successive action chunks disconnected.
We present \method, an asynchronous dual-system controller that aligns perceptual-memory prediction and trajectory correction with actual robot execution.
Rather than forecasting pixels or physical states, a \textbf{Causal Latent World Model} predicts current perceptual memory from delayed visual anchors and executed state--action history, supervised through action-sensitive distillation to retain the frozen expert's control response.
Conditioned on this updated memory, \textbf{Predictive Flow Warm-Starting (PFWS)} aligns the unexecuted action horizon for single-step flow correction without extra training.
On LIBERO-10, \method reaches 93.8\% success at \FinalSharedGain$\times$ inference throughput, improving closed-loop robustness and robot motion smoothness.
\end{pineabstract}

\keywords{
    Vision-Language-Action Models,
    Asynchronous Robot Control,
    Latent World Models,
    Flow Matching,
    Robotic Manipulation}

\section{Introduction}

\begin{figure}[t]
\centering
\includegraphics[width=\textwidth]{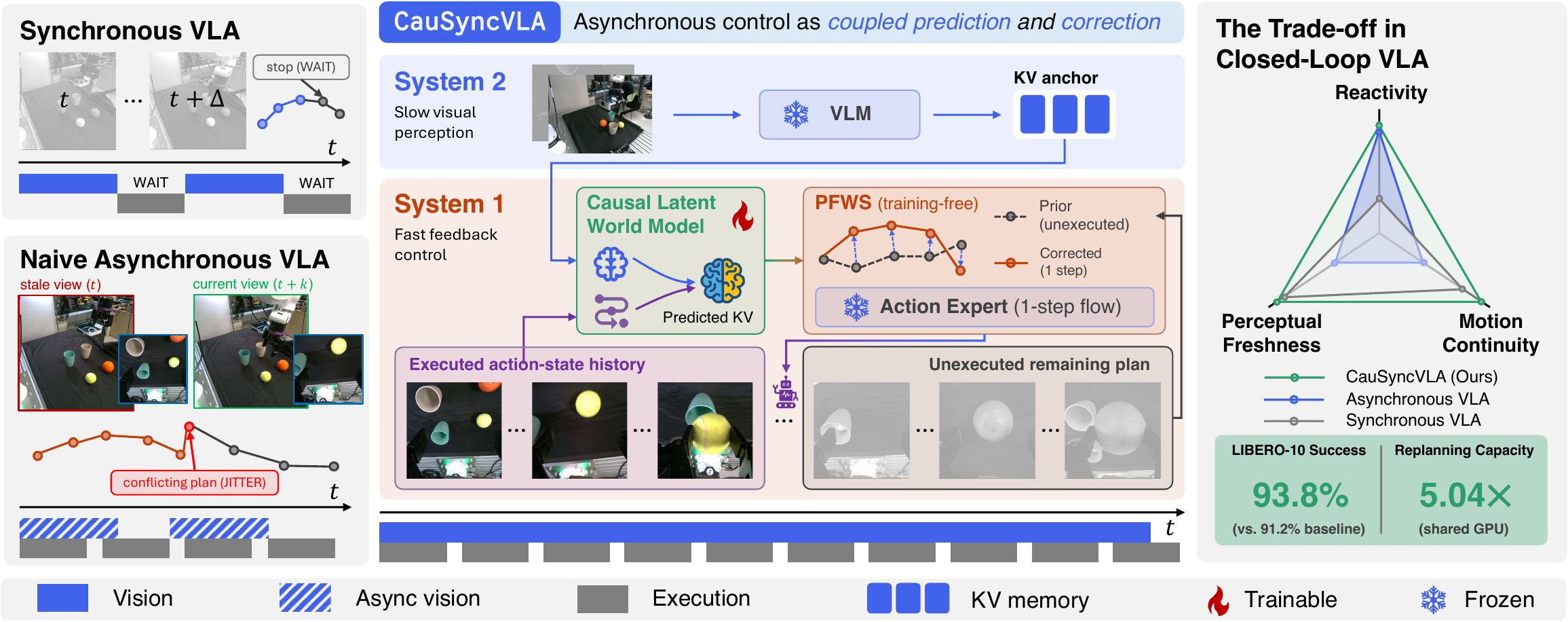}
\caption{\textbf{Coupled prediction and correction in asynchronous control.}
\textbf{Left:} Synchronous control stalls for vision, whereas naive asynchronous control acts on stale views with boundary jitter.
\textbf{Center:} System~2 publishes KV anchors while System~1 propagates KV memory from executed history and refines plans in one frozen-expert PFWS step.
\textbf{Right:} Control trade-offs and LIBERO-10 gains.}
\label{fig:teaser}
\end{figure}

Vision--Language--Action (VLA) policies transfer pretrained multimodal knowledge to robotic manipulation~\citep{brohan2023rt2,kim2024openvla,octo2024}.
They pair foundation backbones with generative action heads to synthesize continuous motion~\citep{chi2023diffusion,black2025pi0,black2025pi05}. 
During physical deployment, however, the environment does not pause for model inference. 
Approaching, grasping, and placing continuously alter the robot--scene configuration and demand timely closed-loop decisions grounded in the evolving physical state.

Meeting this requirement is difficult because heavy visual encoding cannot match the rate of fast physical execution, stalling synchronous closed-loop feedback. 
Action chunking~\citep{zhao2023act,chi2023diffusion} and asynchronous execution~\citep{black2025rtc,shukor2025smolvla} amortize this latency by overlapping inference with ongoing motion, yet leave perception and action temporally misaligned because replaying open-loop tails or stitching boundaries keeps intermediate queries conditioned on delayed perceptual memory when contacts shift mid-chunk. 
Existing methods address stitching or latent extrapolation~\citep{black2025rtc,liu2026latentbridge,yang2026eaglevla}, motivating a common execution reference for perceptual conditioning and action refinement. 
As illustrated in Figure~\ref{fig:teaser}, asynchronous control must reconcile reactivity, perceptual freshness, and trajectory continuity across mismatched sensing and control timescales.

To resolve this misalignment, we observe that actual robot execution supplies a shared temporal reference to align perception and action. 
Observed state transitions and applied commands record physical evolution since the latest visual frame, while the unexecuted tail of the preceding plan provides a structured motion prior. 
Rather than predicting pixels or physical states independently of control, we formulate asynchronous control as coupled perceptual-memory prediction and trajectory correction under one physical clock, updating prefix key--value (KV) memory from execution history before refining remaining actions.

We instantiate this principle in \method, an asynchronous dual-system controller~\citep{kahneman2011thinking}.
A slow visual stream periodically publishes KV memory anchors from observed images, while a fast control stream predicts current perceptual memory and refines subsequent actions. 
Specifically, a Causal Latent World Model predicts current KV memory from the latest delayed anchor and executed state--action history under action-sensitive distillation. 
Conditioned on this predicted memory, Predictive Flow Warm-Starting (PFWS) aligns the unexecuted action horizon to the current step and refines it in one flow step. 
Training only the lightweight predictor keeps the backbone and action expert frozen, retaining pretrained competence while supporting rapid closed-loop replanning between visual updates.

On LIBERO-10, \method raises task success from 91.2\% to 93.8\% over the ten-step synchronous policy while delivering \FinalSharedGain$\times$ higher model replanning capacity on a shared GPU. 
Ablations, perturbation tests, and physical trials isolate the complementary roles of causal perceptual-memory prediction and trajectory correction. 
Our principal technical contributions are organized around three axes.

\begin{itemize}\itemsep=1.0pt\parskip=0pt
    \item \textbf{Execution-grounded asynchronous control.} We formulate asynchronous control as coupled perceptual-memory prediction and trajectory correction, evaluating execution-aligned replanning across simulation and physical trials, reducing feedback latency with frozen VLAs.
    \item \textbf{Causal, action-relevant latent world modeling.} We propagate delayed KV anchors along realized state--action history in latent prefix space, supervising residual updates via action-sensitive distillation, which recovers control-relevant perceptual conditioning.
    \item \textbf{Training-free predictive trajectory correction.} We propose PFWS to reuse unexecuted action horizons as intermediate flow priors, refining trajectories in one frozen-expert Euler step, which suppresses boundary mode switching across perturbed trials.
\end{itemize}

\section{Related Works}

\paragraph{Asynchronous Vision--Language--Action Execution.}
Generalist policies pair multimodal pretraining in RT-1~\citep{brohan2022rt1}, RT-2~\citep{brohan2023rt2}, OpenVLA~\citep{kim2024openvla}, and Octo~\citep{octo2024} with generative control in $\pi_0$~\citep{black2025pi0}, $\pi_{0.5}$~\citep{black2025pi05}, and CogACT~\citep{li2024cogact}.
Later works extend this paradigm through spatial grounding~\citep{pan2026savla}, mobile manipulation~\citep{wu2025momanipvla}, and reinforcement learning~\citep{lu2025vlarl}. 
For higher closed-loop speed, OpenVLA-OFT~\citep{kim2025fine} accelerates decoding, while EfficientVLA and HBVLA reduce model cost through compression and low-bit quantization~\citep{yang2025efficientvla,yan2026hbvla}.
Action chunking~\citep{zhao2023act,chi2023diffusion} further amortizes computation across fixed horizons or adaptive horizons in AAC~\citep{liang2026aac}. 
SmolVLA~\citep{shukor2025smolvla} overlaps prediction with execution, whereas RTC~\citep{black2025rtc} and Acting While Understanding~\citep{yan2026acting} reconcile chunk boundaries via action inpainting or time-misalignment training. 
Still, overlapping or stitching chunks leaves intermediate queries conditioned on stale anchors between visual passes, requiring execution-grounded updates over frozen backbones.

\paragraph{Latent World Models for Visuomotor Control.}
Compensating for delayed perception links asynchronous VLAs to latent dynamics models in DreamerV3~\citep{hafner2023dreamerv3} and TD-MPC~\citep{hansen2022tdmpc}, later extended to manipulation and whole-body control in ManiGaussian~\citep{lu2024manigaussian} and SuperMimic~\citep{liao2026supermimic}. 
Within VLAs, MAP-VLA~\citep{li2025mapvla} injects execution memory, Latent Bridge~\citep{liu2026latentbridge} predicts inter-pass feature or KV deltas, and EagleVLA~\citep{yang2026eaglevla} forecasts future representations for asynchronous inference. 
Action-informed belief learning also links latent memory to control through an inverse-dynamics objective~\citep{wang2026actioninformed}. 
Our Causal Latent World Model advances delayed KV memory anchors along realized state--action transitions inside prefix cache space, supervised through the frozen flow expert via action-sensitive distillation to maintain control fidelity.

\begin{figure}[t]
\centering
\includegraphics[width=\linewidth]{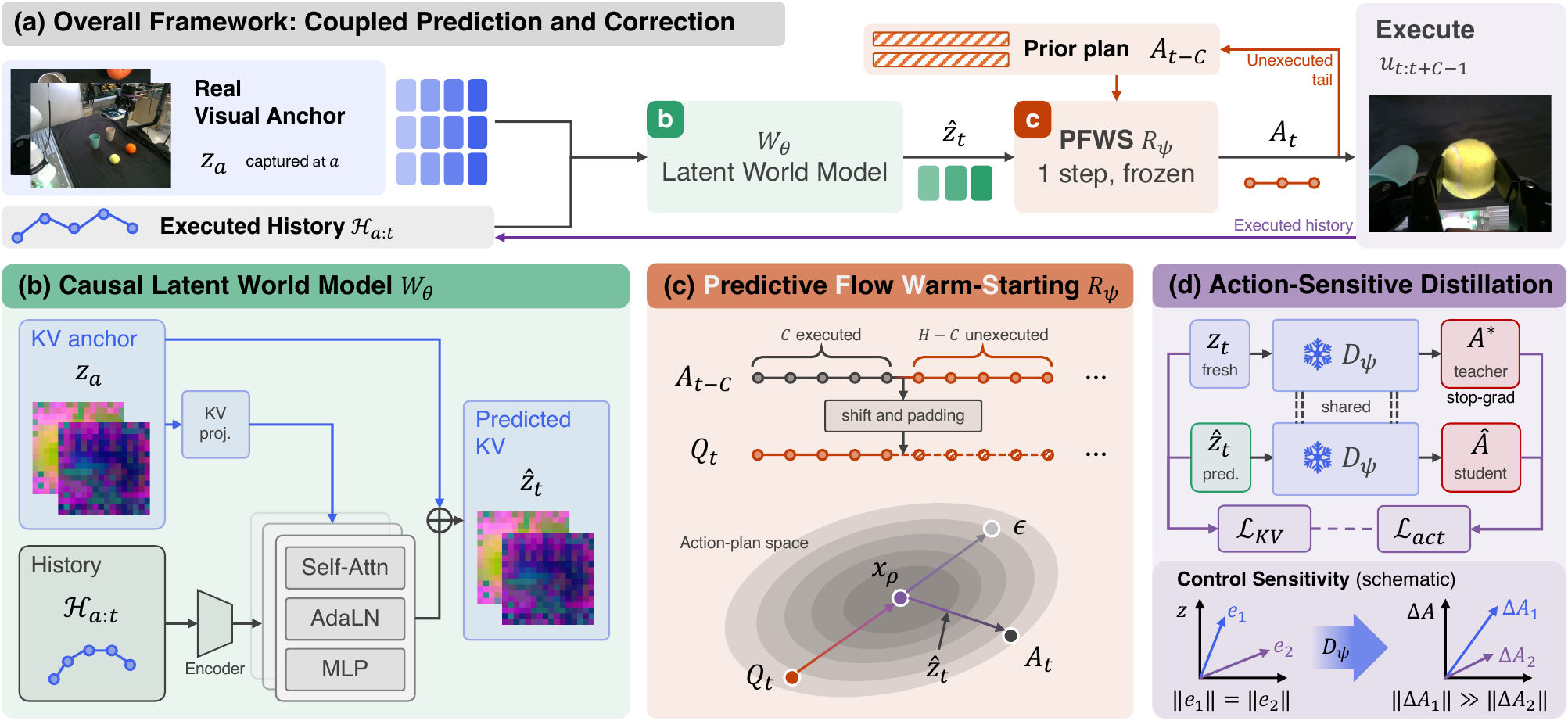}
\caption{\textbf{Architecture and training of \method.}
\textbf{(a)} Executed history $\hist_{a:t}$ propagates delayed KV anchor $z_a$ into predicted memory $\hat z_t$ via $W_\theta$, while prior plan $A_{t-C}$ initializes one-step frozen-expert refinement $\mathcal R_\psi$.
\textbf{(b)} History $\hist_{a:t}$ modulates KV tokens through AdaLN blocks, and adding residual $\delta z_t$ to $z_a$ gives $\hat z_t$.
\textbf{(c)} Shifted prior $\cache_t$ interpolates with noise $\epsilon$ into $x_\rho$ for one Euler step under $\hat z_t$.
\textbf{(d)} Fresh $z_t$ and predicted $\hat z_t$ query frozen endpoint map $D_\psi$ under shared noise, flow time, and state to train $W_\theta$ via $\mathcal L_{\mathrm{KV}}$ and $\mathcal L_{\mathrm{act}}$.}
\label{fig:method}
\end{figure}

\paragraph{Trajectory Priors and Fast Flow Sampling.}
Even with refreshed perceptual memory, resampling trajectories from Gaussian noise at each cycle risks cross-chunk mode switching and multi-step overhead in diffusion~\citep{chi2023diffusion} and flow~\citep{lipman2023flow,albergo2023building} policies. 
Prior speedups rely on Consistency Policy~\citep{prasad2024consistency} distillation, real-time iteration~\citep{duan2025rti} warm-starts, history-informed initialization in A2A~\citep{jia2026action}, history transport in $\Delta$-flow~\citep{zuo2026deltaflow}, or adaptive tail reuse in VLA-Rolling~\citep{he2026vlarolling}. 
Building on trajectory reuse, PFWS keeps expert weights fixed, shifts the unexecuted horizon past consumed commands, rebases state-relative coordinates to the latest proprioceptive state, and applies one Euler refinement under causally predicted memory to improve alignment and continuity across successive chunks.

\section{Method}
\label{sec:method}

Overlapping visual encoding with physical actuation mitigates inference stalls, yet still leaves perceptual conditioning stale and consecutive action chunks spatially misaligned. 
To address this challenge, \method couples perceptual-memory prediction and trajectory correction around the frozen visual--language backbone $B_\phi$ and action expert $v_\psi$~\citep{black2025pi0,black2025pi05} (Figure~\ref{fig:method}a). 
At each replanning boundary, our Causal Latent World Model $W_\theta$ integrates the latest visual anchor with realized state--action history to predict current KV memory. 
Predictive Flow Warm-Starting then refines the prior plan's unexecuted tail under this memory in one frozen-expert Euler step, while action-sensitive distillation guides $W_\theta$.

\subsection{Execution-Grounded Asynchronous Control}
\label{sec:dual_stream}

At environment tick $t$, the robot observes visual frames $o_t$ and proprioceptive state $s_t\in\mathbb R^{d_s}$ while executing command $u_t\in\mathbb R^{d_u}$. 
Perceptual memory is represented by the frozen VLM's prefix key--value tensors, termed KV memory. System~2 encodes image $o_a$ and instruction $\ell$ into $z_a=B_\phi(o_a,\ell)$, the latest published KV memory anchor. Its capture tick is $a=a(t)$ and its delay is $k=t-a$, turning $z_a$ into delayed perceptual memory as physical motion proceeds. 
In parallel, System~1 synthesizes $H$-step action plans and replans every $C$ executed commands $u_{t:t+C-1}$. 
Each fast decision reads two execution-aligned records. 
First, execution history $\hist_{a:t}$ pairs $(s_a,s_t)$ with realized transitions $(s_i-s_a,u_{i-1})$ for $i=a+1,\ldots,t$, recording both realized motion and applied commands since anchor capture. 
Second, trajectory prior $\cache_t$ aligns the unexecuted tail of $A_{t-C}$ with the $C$ consumed commands. 
Thus, realized transitions in $\hist_{a:t}$ advance perceptual memory while the unexecuted action prior $\cache_t$ initializes trajectory refinement.

Combining both execution records gives the closed-loop control law at replanning boundary $t$ as
\begin{equation}
A_t
=
\mathcal R_\psi\!\left(
\cache_t,\,
W_\theta(z_a,\hist_{a:t})
\right),
\label{eq:coupled_control}
\end{equation}
where $W_\theta$ predicts current KV memory $\hat z_t$ and $\mathcal R_\psi$ refines the aligned prior through frozen expert $v_\psi$ without auxiliary weights, omitting noise and state inputs. 
Equation~\ref{eq:coupled_control} formalizes dual-stream coupling by letting execution feedback update perceptual conditioning before trajectory revision. 
History $\hist_{a:t}$ contains observed states and applied commands, while capture tick $a$ and the preceding replanning boundary $t-C$ can differ. 
Completed visual passes publish new KV anchors asynchronously, whereas intermediate fast cycles advance anchor $z_a$ across varying slow-to-fast ratios.

\subsection{Causal Latent World Model}
\label{sec:latent_world_model}

Between visual updates, physical motion renders cached prefix keys and values out of date before System~2 finishes encoding a new frame. 
Rather than synthesizing intermediate RGB images, the Causal Latent World Model $W_\theta$ operates directly inside the frozen VLM's prefix space as shown in Figure~\ref{fig:method}b, preserving fused vision--language semantics via an anchor-conditioned residual
\begin{equation}
\hat z_t
=
W_\theta(z_a,\hist_{a:t})
=
z_a+\amsbb{I}[k>0]\,\delta z_t,
\label{eq:world_model}
\end{equation}
where $\delta z_t$ is a validity-masked, scale-restored KV residual conditioned on $\hist_{a:t}$, and the age gate $\amsbb{I}[k>0]$ preserves unmodified anchors when $k=0$. 
Rather than forecasting future physical states or claiming causal identification of external scene dynamics, $W_\theta$ operates causally in decision-time information, conditioning strictly on the latest real KV memory anchor $z_a$ and realized state--action history $\hist_{a:t}$ up to tick $t$ until the next completed visual pass refreshes $z_a$.

To compute $\delta z_t$, $W_\theta$ standardizes keys and values using per-layer RMS scales~\citep{zhang2019root} $\sigma^{\mathrm{KV}}$ to equalize depth-wise magnitudes, concatenates layer features per prefix slot, and projects them into tokens with learned positional embeddings. 
In parallel, a trajectory MLP encodes normalized anchor age, endpoints, net displacement, and padded transitions of normalized state deltas and native-coordinate commands. 
This encoding aggregates observed displacement and commanded motion relative to the same anchor while keeping valid task prompt slots intact. 
Bidirectional self-attention and MLP blocks then inject this history embedding via adaptive LayerNorm~\citep{peebles2023dit}. 
Finally, an output projection restores the post-RoPE~\citep{su2024roformer} KV layout under residual scales $\sigma^\Delta$ and prefix mask $m$, zeroing padded slots before adding $\delta z_t$ to $z_a$. 
Zero-initialized projections configure $W_\theta$ as an identity map at initialization, while exceeding history capacity triggers a synchronous refresh matching the horizon settings in Table~\ref{tab:config}.

\subsection{Predictive Flow Warm-Starting}
\label{sec:flow_continuation}

Predicted KV memory $\hat z_t$ updates action conditioning, but cold-start sampling discards the preceding trajectory and can switch between modes at chunk boundaries. 
Figure~\ref{fig:method}c instantiates $\cache_t$ and $\mathcal R_\psi$ in Eq.~\ref{eq:coupled_control} through Predictive Flow Warm-Starting (PFWS), drawing on warm-started optimization~\citep{duan2025rti} and horizon reuse~\citep{he2026vlarolling}. 
After $C$ executed steps, shifting the remaining $H-C$ entries of $A_{t-C}\in\mathbb R^{H\times D}$ forward and repeating the terminal action $C$ times forms the prior $\cache_t=\operatorname{ShiftRepeat}(A_{t-C},C)$. 
State-relative coordinates are rebased to $s_t$ to align $\cache_t$ with predicted KV memory $\hat z_t$.

Rather than denoising from pure noise, PFWS interpolates $\cache_t$ with Gaussian noise $\epsilon\sim\mathcal N(0,I)$ at intermediate flow time $\rho\in(0,1]$ as
\begin{equation}
\begin{aligned}
x_\rho
&=(1-\rho)\cache_t+\rho\epsilon,\\
\mathcal R_\psi(\cache_t,z)
&=x_\rho-\rho\,v_\psi(x_\rho,\rho\mid z),
\end{aligned}
\label{eq:one_step}
\end{equation}
where flow time $\rho$ proceeds from noise at $1$ to actions at $0$~\citep{black2025pi0}. 
Evaluating the same frozen expert $v_\psi$ at $z=\hat z_t$ refines the rebased prior in one Euler step, so that trajectory continuation and memory updates share one execution-aligned reference $(s_t,\hat z_t)$.
Together with a fixed-length history representation, this single pass maintains constant fast-loop complexity independent of anchor age.
The factor $\rho\in(0,1)$ trades continuity against reactivity, where smaller $\rho$ keeps commands close to $\cache_t$ to suppress chatter, moderate $\rho$ preserves headroom for $\hat z_t$-guided target correction, and $\rho=1$ recovers cold-start sampling at episode onset.
Caching endpoint $A_t$ before unnormalization and dispatching its leading $C$ commands $u_{t:t+C-1}$ sustains cross-chunk smoothness without modifying expert weights.

\subsection{Action-Sensitive Distillation}
\label{sec:objective}

Pretrained VLM KV memory encodes broad scene semantics, but equal-sized memory errors can have different effects on action prediction.
KV regression alone does not capture this sensitivity, whereas downstream control depends on how $\hat z_t$ steers the frozen expert in Eq.~\ref{eq:one_step}. 
We supervise $W_\theta$ through the one-step output of this same expert alongside KV regression, as shown in Figure~\ref{fig:method}d. 
Given a training pair $(a,t)$ with history $\hist_{a:t}$ and fresh KV memory $z_t=B_\phi(o_t,\ell)$, we define $D_\psi(z;x_\rho,\rho)$ to extract the $d_u$ normalized control dimensions of $x_\rho-\rho v_\psi(x_\rho,\rho\mid z)$ using the same frozen expert as PFWS. 
Shared $(x_\rho,\rho)$ and robot states yield detached teacher target $A^\star=\operatorname{sg}[D_\psi(z_t;x_\rho,\rho)]$ and student $\hat A=D_\psi(\hat z_t;x_\rho,\rho)$ under this composite loss
\begin{equation}
\mathcal L(\theta)
=
\underbrace{\left\|
\frac{\hat z_t-z_t}{\sigma^\Delta}
\right\|_m^2}_{\mathcal L_{\mathrm{KV}}}
+
\lambda\underbrace{\left\|\hat A-A^\star\right\|_w^2}_{\mathcal L_{\mathrm{act}}},
\label{eq:total_loss}
\end{equation}
where $\mathcal L_{\mathrm{KV}}$ averages normalized error across valid KV entries and $\mathcal L_{\mathrm{act}}$ averages coordinate-weighted error under weight vector $w$ across $H d_u$ control dimensions. 
Sharing $(x_\rho,\rho)$ aligns memory-induced control errors and distills the fresh-memory control response~\citep{rusu2016distillation} into $W_\theta$ via frozen-expert KV gradients without modifying $v_\psi$.

\begin{table}[t]\centering\footnotesize
\caption{\textbf{Task competence and inference capacity on LIBERO-10.} Metrics span 500 official starts under a 520-action budget alongside batch-one model inference throughput, while Table~\ref{tab:related_execution} evaluates 20\,Hz timed closed-loop execution. Bold and underline mark the best and second-best verified values.}
\label{tab:final_native}
\setlength{\tabcolsep}{2.5pt}
\begin{tabularx}{\textwidth}{llcr*{4}{>{\raggedleft\arraybackslash}X}}\toprule
 & & & & \multicolumn{2}{c}{Latency (ms) $\downarrow$} & \multicolumn{2}{c}{Model capacity $\uparrow$}\\
\cmidrule(lr){5-6}\cmidrule(l){7-8}
Controller & Decoder & $H/C$ & Success (\%) $\uparrow$ & Mean & P95 & Plans/s & Actions/s\\\midrule
$\pi_{0.5}$ & 10-step flow & 10/5 & \TenSeven & \MTFiveStockMean & \MTFiveStockTail & \MTFiveStockRate & \MTFiveStockSupply\\
$\pi_{0.5}$ & 1-step flow & 10/5 & \OneSeven & \MTFiveStockOneMean & \MTFiveStockOneTail & \MTFiveStockOneRate & \MTFiveStockOneSupply\\
$\pi_{0.5}$ + PFWS & 1-step flow & 10/5 & \SyncTFWSSuccess & \second{\SyncTFWSMean} & \second{\SyncTFWSPninetyfive} & \second{\SyncTFWSRate} & \second{\SyncTFWSSupply}\\
GR00T N1.7~\citep{nvidia2026grootlibero} & 4-step flow & 16/5 & \GrootSuccess & \GrootMs & \GrootPninetyfive & \GrootRate & \GrootSupply\\
OpenVLA-OFT~\citep{kim2025fine} & Parallel L1 & 8/5 & \OftSuccess & \OftMs & \OftPninetyfive & \OftRate & \OftSupply\\
RTC~\citep{black2025rtc} & 10-step + guidance & 10/5 & 91.80 & 179.82 & 182.66 & 5.56 & 27.81\\
Latent Bridge R0~\citep{liu2026latentbridge} & 2-step + bridge & 10/5 & 91.80 & 63.06 & 78.29 & 15.86 & 79.29\\
VLA-Rolling~\citep{he2026vlarolling} & Adaptive 1--10 & 50/10 & \second{93.40} & 166.47 & 202.06 & 6.01 & 60.07\\
\midrule
\rowcolor{tablerow}\method{} & 1-step flow & 10/5 & \best{\FinalSuccess} & \best{\FinalSharedMean} & \best{\FinalSharedPninetyfive} & \best{\FinalSharedRate} & \best{\FinalSharedSupply}\\
\bottomrule
\end{tabularx}\end{table}

A first-order expansion of the expert response in KV error $e=\hat z_t-z_t$, with Jacobian $J_D=\nabla_zD_\psi(z_t;x_\rho,\rho)$, yields a quadratic approximation to the action loss
\begin{equation}
\mathcal L_{\mathrm{act}}
\approx
\frac{1}{H d_u}
\left\|\Omega^{1/2}J_D e\right\|_2^2,
\label{eq:sensitivity}
\end{equation}
where $\Omega=I_H\otimes\operatorname{diag}(w)$. 
Thus, equal-norm KV errors induce unequal action shifts across expert input directions. 
Minimizing $\mathcal L_{\mathrm{act}}$ induces a pullback metric $J_D^\top \Omega J_D$ weighting KV directions by control impact without forming $J_D$, while $\mathcal L_{\mathrm{KV}}$ regularizes $\operatorname{ker}(J_D)$ toward fresh VLM statistics to support repeated execution-conditioned updates across delayed visual intervals. 
Mixing cold $\rho=1$ and warm $0<\rho<1$ queries aligns $W_\theta$ across onset generation and mid-trajectory refinement per the training schedule in Section~\ref{sec:exp_setup} and Table~\ref{tab:config}.

\section{Experiments}
Our evaluation separates individual module contributions from whole-system asynchronous execution while testing whether decoupled visual encoding accelerates closed-loop reaction without sacrificing competence, whether causal latent prediction preserves conditioning between visual refreshes, and whether PFWS sustains smoothness and robustness. We further ablate each component and training objective within matched cohorts before validating closed-loop physical deployment on a Franka FR3 arm.

\subsection{Experimental Setup}
\label{sec:exp_setup}
Built on a frozen $\pi_{0.5}$ backbone and flow action expert, the Causal Latent World Model is a 24.27M-parameter AdaLN transformer trained for 12k updates with simulation horizon $H=10$, stride $C=5$, one Euler step, and PFWS factor $\rho=0.5$, while baselines retain public weights and official inference heads. Specifically, RTC and Latent Bridge R0 share the OpenPI $\pi_{0.5}$ base at $H/C=10/5$, whereas VLA-Rolling retains its LeRobot backbone at $H/C=50/10$. Full hyperparameters and training datasets appear in Appendix~\ref{sec:method_details} and Table~\ref{tab:config}.

Competence and robustness span 500 native starts on LIBERO-10~\citep{liu2023libero} and 2,519 paired perturbations on LIBERO-Plus~\citep{fei2025liberoplus}, while timed closed-loop execution evaluates 100 shared starts at 20\,Hz under a 26\,s budget on an exclusive RTX~5090 GPU alongside scheduling analyses in Appendix~\ref{sec:pretrained_protocol}. Physical manipulation is evaluated on a 7-DoF Franka FR3 arm with dual D405 cameras at $H/C=10/2$ under a 100\,Hz dispatch ceiling and live 30\,Hz perception, deploying an offline timestamp-adapted updater trained on 10\,Hz demonstrations across altered cup stacking and held-out fruit placement and vegetable sorting per the protocol and trial counts in Appendix~\ref{sec:real_robot_supp}.

\subsection{Task Competence and Closed-Loop Latency}
\label{sec:exp_speed}
Decoupling fast action generation from slow visual encoding must overcome single-step Euler truncation without relying on stale visual features. Across LIBERO-10 in Table~\ref{tab:final_native} and Table~\ref{tab:supp_final}, compressing synchronous $\pi_{0.5}$ from ten steps to one accelerates inference yet erodes competence because coarse integration from Gaussian noise amplifies discretization drift across contact-rich subgoals. Warm-starting one-step flow with PFWS from Section~\ref{sec:flow_continuation} and Eq.~\ref{eq:one_step} recovers accuracy by initializing near the shifted prior chunk, whereas RTC~\citep{black2025rtc} and Latent Bridge R0~\citep{liu2026latentbridge} retain ten-step guided inpainting or bridge overhead, and VLA-Rolling sustains high accuracy only at 166.47\,ms mean plan latency. Propagating delayed visual anchors along realized execution feedback per Eq.~\ref{eq:coupled_control} removes System~2 stalls from the control path while keeping conditioning aligned with motion, yielding the native point estimates in Table~\ref{tab:final_native} across comparison cohorts separated in Appendix~\ref{sec:review_evidence}.

Beyond raw forward speed, practical deployment requires that model capacity translates into lower closed-loop response latency and timely task completion under fixed wall-clock budgets. As shown in Figure~\ref{fig:capacity} and Table~\ref{tab:final_timing}, reducing flow steps cuts shared-GPU latency from 60.01 to 36.00\,ms, while offloading visual encoding lowers fast-loop latency further to 11.91\,ms, yielding fivefold shared-GPU and over twelvefold isolated capacity gains while $W_\theta$ preserves conditioning quality rather than accelerating compute. Under co-located single-GPU 20\,Hz execution in Table~\ref{tab:related_execution}, this decoupled loop from Section~\ref{sec:dual_stream} halves response latency relative to synchronous $\pi_{0.5}$ from 176.73 to 83.04\,ms and increases realized replanning, lifting timed completion to 91.00\% ahead of deadline-bound schedules, with separate-GPU RTC timing and scheduling diagnostics reported in Appendix~\ref{sec:pretrained_protocol} and Tables~\ref{tab:deadline_detail}, \ref{tab:clock_audit}, and~\ref{tab:rtc_isolated_execution}.

\begin{table}[!t]\centering\footnotesize
\caption{\textbf{Single-GPU closed-loop execution within 26 seconds.} Policy inference and simulation rendering share one RTX~5090. Timed success, observation-to-first-action response P50/P95, and realized dispatch span 100 shared 20\,Hz starts.}
\label{tab:related_execution}
\setlength{\tabcolsep}{9pt}
\input{tables/pretrained_baselines.tex}
\end{table}

\subsection{Action-Relevant Latent Prediction}
\label{sec:exp_memory}
Higher replanning speed alone cannot guide contact-rich manipulation unless predicted perceptual memory tracks evolving scene geometry between visual refreshes. Inside frozen cross-attention layers, outdated KV caches force the action expert to synthesize velocities relative to outdated end-effector and object poses, creating systematic spatial lag. Across 320 held-out windows in Table~\ref{tab:history}, stale KV anchors degrade predictions, whereas propagating delayed anchors via $W_\theta$ in Section~\ref{sec:latent_world_model} and Eq.~\ref{eq:world_model} cuts downstream action MSE over tenfold.

\begin{table}[!t]\centering\small
\caption{\textbf{Execution-history ablations in memory prediction.} Fresh-target errors across 320 held-out windows under masked or permuted records preserving state--action pairings.}
\label{tab:history}
\setlength{\tabcolsep}{9pt}\input{tables/claim_history_table.tex}
\end{table}

\begin{figure}[!t]\centering
\begin{minipage}[t]{0.48\linewidth}\centering\vspace{0pt}
\includegraphics[width=\linewidth]{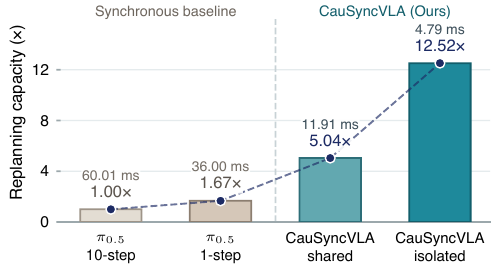}
\caption{\textbf{Normalized model replanning capacity.} Inverse batch-one forward latency at $C=5$ relative to ten-step synchronous $\pi_{0.5}$, comparing shared single-GPU visual encoding against isolated System~1 execution.}
\label{fig:capacity}
\end{minipage}\hfill
\begin{minipage}[t]{0.48\linewidth}\centering\vspace{0pt}
\includegraphics[width=\linewidth]{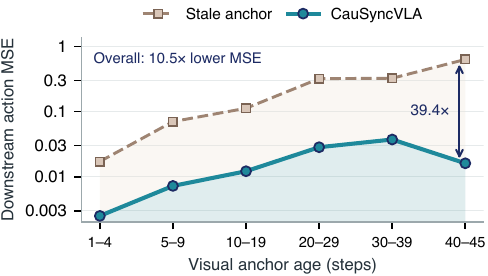}
\caption{\textbf{Downstream action error across anchor staleness.} Logarithmic action MSE of stale and $W_\theta$-predicted conditioning relative to fresh-anchor outputs across 320 windows.}
\label{fig:age}
\end{minipage}
\end{figure}

Simply knowing elapsed time or start--end gripper positions cannot distinguish curved trajectories or intermediate contact events between camera frames. Progressively ablating conditioning signals in Table~\ref{tab:history} isolates how ordered execution history resolves this ambiguity, reflecting empirical alignment within the expert interface rather than causal identification of external dynamics. Elapsed age encodes drift duration without directional geometry, while endpoints $(s_a, s_t)$ collapse intermediate trajectories into net offsets. Unordered increments ground traversed geometry to $z_a$ via $(s_i-s_a, u_{i-1})$ and cut error nearly fourfold, whereas causal ordering preserves sequential command--state coupling and lowers action MSE over tenfold. Across staleness horizons up to 45 steps in Figure~\ref{fig:age} and Table~\ref{tab:supp_age}, $W_\theta$ keeps action error an order of magnitude below stale anchors until integration drift warrants a System~2 refresh. Dual-camera PCA projections in Figures~\ref{fig:kv_mugs}--\ref{fig:kv_drawer} and Table~\ref{tab:kv_examples} confirm lower full-dimensional KV error alongside localized representation shifts across dynamic scene views.

\subsection{Trajectory Correction with PFWS}
\label{sec:exp_tfws}
Restoring scene semantics via $W_\theta$ resolves spatial lag, yet Gaussian resampling at chunk boundaries can still trigger mode switching even under synchronous fresh visual conditioning. Anchoring generation to prior plans via PFWS from Section~\ref{sec:flow_continuation} and Eq.~\ref{eq:one_step}, illustrated in Figure~\ref{fig:method}c and detailed in Appendix~\ref{sec:pfws_details}, suppresses reversals in Table~\ref{tab:tfws_main} and Table~\ref{tab:smoothness}, nearly halving boundary flips and eliminating boundary-versus-interior flip excess across full and matched episodes. While native success gains are modest with inconclusive paired evidence in Appendix~\ref{sec:review_evidence}, trajectory continuity provides its primary gain.

\begin{table}[!t]\centering\footnotesize
\caption{\textbf{PFWS competence and trajectory continuity.} Episode-averaged flip rates over 500 native starts and 2,519 LIBERO-Plus variants, where zero boundary excess denotes interior-step parity.}
\label{tab:tfws_main}
\setlength{\tabcolsep}{9pt}
\input{tables/pfws_comparison_table.tex}
\end{table}

\begin{figure}[!t]\centering
\begin{minipage}[t]{0.48\linewidth}\centering\vspace{0pt}
\parbox[c][0.62\linewidth][c]{\linewidth}{\centering\includegraphics[width=\linewidth]{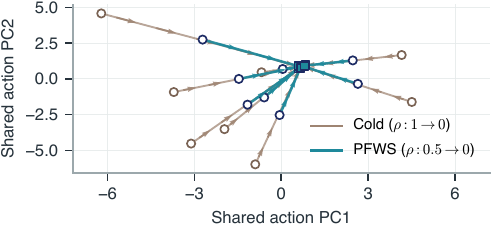}}
\caption{\textbf{Cold and PFWS frozen-expert flow paths.} Shared PCA projections across eight matched seeds tracing $v_\psi$ from initial latents (circles) to dense endpoints (squares).}
\label{fig:flow_geometry}
\end{minipage}\hfill
\begin{minipage}[t]{0.48\linewidth}\centering\vspace{0pt}
\parbox[c][0.62\linewidth][c]{\linewidth}{\centering\includegraphics[width=\linewidth]{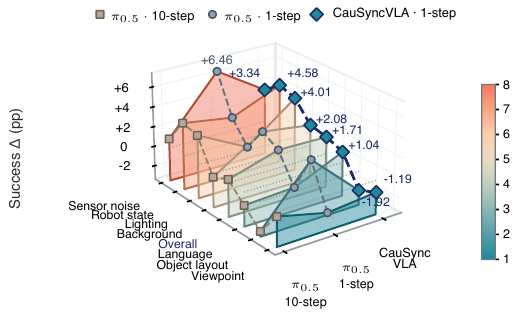}}
\caption{\textbf{PFWS differentials across LIBERO-Plus.} Percentage-point gains over cold starts across 2,519 variants with fixed $W_\theta$ in \method and synchronous vision in $\pi_{0.5}$.}
\label{fig:plus}
\end{minipage}
\end{figure}

Supporting single-step refinement around rebased action tails without fine-tuning $v_\psi$ depends on the pretrained flow field preserving consistent velocity directions over the remaining interval. Projecting frozen-expert trajectories onto a shared PCA basis in Figure~\ref{fig:flow_geometry} and Appendix~\ref{sec:flow_geometry} reveals low-curvature streamlines whose cold and warm ODE paths exceed unit path-to-chord length by about 0.3\%, while a matched $[0.5, 0]$ cold suffix mirrors warm accuracy. Seeding integration from shifted prior tails halves the remaining flow distance along aligned vector fields, reducing one-step endpoint error by 19.7\% without modifying expert weights.

This trajectory reliance defines an operating boundary across the 2,519 LIBERO-Plus rollouts in Figure~\ref{fig:plus} and Table~\ref{tab:plus_full}. Within both synchronous $\pi_{0.5}$ and \method, PFWS improves robustness under sensor noise and robot-state shifts where rebased tails $\cache_t$ filter jitter, whereas viewpoint and layout jumps displace grasp attractors and favor cold starts. Because \method still trails synchronous PFWS overall in Table~\ref{tab:tfws_main}, trajectory reuse acts as a within-controller regularizer rather than unconditional perturbation robustness.

\subsection{Component and Objective Ablations}
\label{sec:exp_ablation}
We next test why coupling memory propagation with trajectory reuse outperforms each mechanism in isolation, evaluating every ablation within its matched cohort in Table~\ref{tab:cohorts}. Disabling $W_\theta$ while keeping PFWS in Table~\ref{tab:ablation_main} and Table~\ref{tab:controls} drops competence to 80.20\% because warm-started refinement then steers continuous motion toward stale anchors, whereas dropping PFWS retains success yet nearly doubles boundary flips and coupling both preserves smoothness at 93.40\%. Paired discordant tests in Appendix~\ref{sec:review_evidence} confirm significant competence gains from memory prediction, while PFWS acts primarily through continuity rather than a standalone accuracy leap.

\begin{table}[!t]\centering\footnotesize\renewcommand{\arraystretch}{0.98}
\caption{\textbf{Component and objective ablations on LIBERO-10.} Module removals over 500 starts (top) and uniform KV loss versus action distillation across matched 12k-update predictors (bottom).}
\label{tab:ablation_main}\label{tab:modules_main}\label{tab:loss_main}
\setlength{\tabcolsep}{9pt}
\begin{tabular}{lrrr}\toprule
Controller & Success (\%) $\uparrow$ & \multicolumn{2}{r}{Pooled flips (\%) $\downarrow$}\\\midrule
w/o latent world model & 80.20 & \multicolumn{2}{r}{2.30}\\
w/o PFWS & 92.00 & \multicolumn{2}{r}{3.69}\\
Sync. PFWS (fresh KV) & 92.00 & \multicolumn{2}{r}{\best{1.26}}\\\midrule
\rowcolor{tablerow}Full (w/ latent WM + PFWS) & \best{93.40} & \multicolumn{2}{r}{\second{1.86}}\\\midrule
Objective & Success (\%) $\uparrow$ & KV error $\downarrow$ & Action MSE $\downarrow$\\\midrule
w/o action distillation & 85.60 & \best{0.28461} & 0.02018\\\midrule
\rowcolor{tablerow}w/ action distillation & \best{92.80} & 0.31958 & \best{0.00983}\\\bottomrule
\end{tabular}\end{table}

\begin{table}[!t]\centering\footnotesize\renewcommand{\arraystretch}{0.96}
\caption{\textbf{Closed-loop competence, feedback latency, and motion smoothness on a physical Franka FR3 robot.} Evaluated at $H=10$ and $C=2$ with cohort counts detailed in Appendix~\ref{sec:real_robot_supp}, reporting all-trial medians for dispatch, latency, and 100\,ms EEF motion alongside mean Time over solved runs.}
\label{tab:fr3_closed_loop}
\setlength{\tabcolsep}{3pt}\input{tables/fr3_final_table.tex}
\end{table}

We next test whether unconstrained KV reconstruction suffices to preserve the frozen expert's control response. In the lower panel of Table~\ref{tab:ablation_main}, adding action distillation via Eq.~\ref{eq:total_loss} in Section~\ref{sec:objective} cuts action MSE from 0.02018 to 0.00983 and raises success from 85.60\% to 92.80\% despite increasing raw KV error from 0.28461 to 0.31958, establishing a statistically significant paired objective gain in Appendix~\ref{sec:review_evidence}. This metric reversal reflects the sensitivity geometry of Eq.~\ref{eq:sensitivity}, where uniform KV distance weights all cache coordinates equally regardless of Jacobian sensitivity $J_D$, whereas matching frozen-expert outputs focuses supervision on perturbations that alter generated actions. Tracking optimization across update budgets in Figure~\ref{fig:supp_training} and Table~\ref{tab:validation} likewise favors checkpoint selection by downstream control accuracy over KV regression alone.

\subsection{Real-Robot Evaluation}
\label{sec:real_robot}
\begin{figure}[!t]\centering
\captionsetup[subfigure]{font={scriptsize},skip=2pt}
\makebox[\linewidth][l]{\footnotesize\textbf{$\pi_{0.5}$}}\par\vspace{1pt}
\begin{subfigure}[t]{.244\linewidth}\centering
\includegraphics[width=\linewidth]{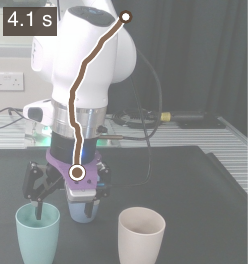}
\caption{Grasp 1.}\label{fig:fr3_baseline_1}
\end{subfigure}\hfill
\begin{subfigure}[t]{.244\linewidth}\centering
\includegraphics[width=\linewidth]{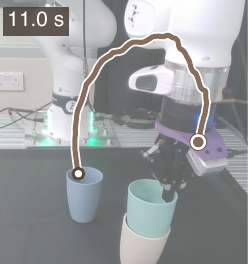}
\caption{Place 1.}\label{fig:fr3_baseline_2}
\end{subfigure}\hfill
\begin{subfigure}[t]{.244\linewidth}\centering
\includegraphics[width=\linewidth]{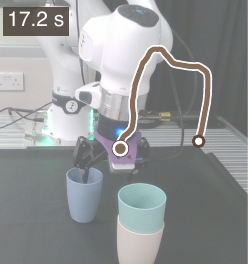}
\caption{Grasp 2.}\label{fig:fr3_baseline_3}
\end{subfigure}\hfill
\begin{subfigure}[t]{.244\linewidth}\centering
\includegraphics[width=\linewidth]{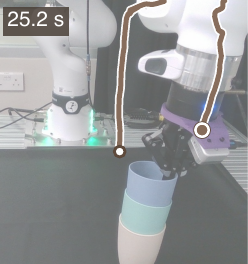}
\caption{Place 2.}\label{fig:fr3_baseline_4}
\end{subfigure}\par\vspace{5pt}
\makebox[\linewidth][l]{\footnotesize\textbf{\method}}\par\vspace{1pt}
\begin{subfigure}[t]{.244\linewidth}\centering
\includegraphics[width=\linewidth]{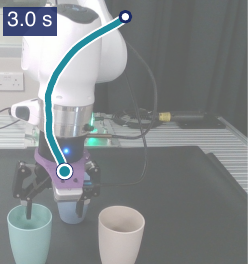}
\caption{Grasp 1.}\label{fig:fr3_ours_1}
\end{subfigure}\hfill
\begin{subfigure}[t]{.244\linewidth}\centering
\includegraphics[width=\linewidth]{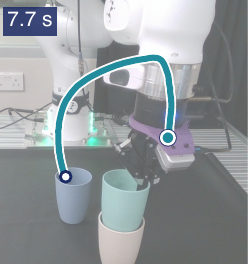}
\caption{Place 1.}\label{fig:fr3_ours_2}
\end{subfigure}\hfill
\begin{subfigure}[t]{.244\linewidth}\centering
\includegraphics[width=\linewidth]{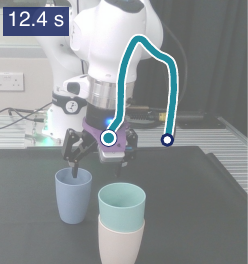}
\caption{Grasp 2.}\label{fig:fr3_ours_3}
\end{subfigure}\hfill
\begin{subfigure}[t]{.244\linewidth}\centering
\includegraphics[width=\linewidth]{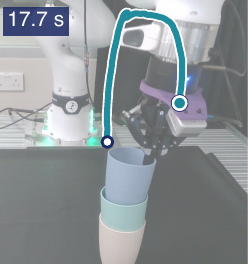}
\caption{Place 2.}\label{fig:fr3_ours_4}
\end{subfigure}\par\vspace{2pt}

\caption{\textbf{Stage-aligned cup stacking on Franka FR3.} Aligned three-cup grasp--place milestones for $\pi_{0.5}$ (top) and \method (bottom) under shared crops and masking, tracing gripper motion from circles to endpoints.}
\label{fig:fr3_selected}
\end{figure}

Physical manipulation exposes contact compliance, camera transport jitter, and surface friction that turn minor perceptual phase lag into rim collisions or object slip when visual encoding stalls. Across Franka FR3 evaluations in Table~\ref{tab:fr3_closed_loop} and Tables~\ref{tab:fr3_final_timing}--\ref{tab:fr3_final_windows}, synchronous control waits over 72\,ms at replanning boundaries and leaves terminal errors uncorrected during grasp closure and release, whereas cutting median response fivefold to 13.17\,ms via Eq.~\ref{eq:coupled_control} more than doubles active dispatch to 43.13\,actions/s and lifts pooled success from 72.5\% to 87.0\%. Normalized execution indexing in Sections~\ref{sec:latent_world_model} and~\ref{sec:exp_memory} transfers offline 10\,Hz adaptation directly to live 30\,Hz cup stacking under altered layouts, while zero-shot improvements carry over to held-out fruit placement and deformable vegetable sorting where timely feedback prevents slip and crushing.

Faster hardware replanning typically risks actuator chatter when independent chunks switch modes or amplify sensor noise. Consistent with Sections~\ref{sec:exp_tfws}--\ref{sec:exp_ablation}, coupling propagated KV anchors with PFWS from Eq.~\ref{eq:one_step} lets higher dispatch coincide with lower P95 tail latency, speed variation, and acceleration RMS across all three suites in Table~\ref{tab:fr3_closed_loop}, cutting mean solved duration from 25.16 to 17.93\,s. Inside uninterrupted System~2 intervals, Table~\ref{tab:fr3_final_windows} confirms multi-plan intra-window updates in 35.00\% of slow passes, while unsmoothed traces and frames in Figure~\ref{fig:fr3_selected}, Table~\ref{tab:fr3_selected_pair}, and Figures~\ref{fig:fr3_selected_motion}--\ref{fig:fr3_eef_coordinates} show synchronous $\pi_{0.5}$ stalling near cup rims with velocity spikes upon frame arrival whereas \method sustains continuous approach arcs.

Because physical trials reset scenes independently and active dispatch co-varies with fast-loop latency while both controllers pause and clear priors at shared gripper barriers, Table~\ref{tab:fr3_closed_loop} measures deployed whole-system performance rather than fixed-frequency module effects, with active rates excluding shared waits retained in total solve time and Table~\ref{tab:fr3_final_timing}. Per-suite trial counts, Wilson confidence intervals, and single-backbone $\pi_{0.5}$ scope limits appear in Appendices~\ref{sec:real_robot_supp} and~\ref{sec:review_evidence}.

\section{Conclusion}
We presented \method, an execution-grounded asynchronous controller that reconciles perceptual freshness with cross-chunk continuity around frozen VLA policies without retraining. Coupling history-conditioned prefix KV prediction with single-step flow warm-starting keeps rapid trajectory revisions aligned with actual robot motion and clarifies a continuity--adaptation boundary between local stabilization and large exogenous shifts. Operating strictly through cached prefix KV memory and chunked flow priors makes the design directly compatible with single-system flow VLAs, while multi-backbone hardware validation, event-triggered visual refreshes, and exogenous prior resets in Appendix~\ref{sec:limitations} mark natural next steps toward broad physical autonomy.

\bibliography{main}

\clearpage
\raggedbottom
\appendix
\makeatletter
\setlength{\@fptop}{0pt}
\setlength{\@fpsep}{1.5em}
\setlength{\@fpbot}{0pt plus 1fil}
\makeatother
\pdfbookmark[0]{Appendix}{appendix}
\noindent{\sffamily\bfseries\Large\color{PineLink}Appendix}\par\vspace{0.5em}

\section{Method Details and Theoretical Analysis}
\label{sec:method_details}
\subsection{Experimental Setup and Architecture}
This appendix expands the main text's three evaluations of feedback response, action-relevant memory, and trajectory continuity. Table~\ref{tab:config} and Algorithm~\ref{alg:dual} summarize the controller architecture and training hyperparameters; simulation environments run in robosuite~\citep{zhu2022robosuite} on the MuJoCo physics engine~\citep{mujocoDocs}. Subsequent sections document protocols, analytical scope, and physical trials.

\begin{table}[ht]
\centering\small
\setlength{\tabcolsep}{3pt}
\caption{\textbf{Controller architecture and training setup.} A global update combines two devices and two gradient-accumulation steps per device. KV and action dimensions match the frozen expert.}
\label{tab:config}
\begin{tabularx}{0.88\textwidth}{>{\raggedright\arraybackslash}Xl}\toprule
Hyperparameter & Value \\\midrule
Updater parameters & 24,273,024 \\
Native KV Dimension per Token & $18 \text{ layers} \times 256 \times 2 = 9{,}216$ \\
Compact Prefix Slots & 544 ($512 \text{ visual} + 32 \text{ language}$) \\
Trajectory Conditioner Input Dim & 745 ($1 + 3\times 8 + 48\times(8+7)$) \\
Transformer Depth / Width / Heads & 6 blocks / 384 dim / 4 heads \\
Transformer MLP Intermediate Dim & 1,536 \\
Trajectory Conditioner MLP & $745 \to 768 \to 384$ (GELU) \\
State / Flow Latent / Control Dim & 8 / 32 / 7 \\
Action Chunk Horizon / Consumed ($H / C$) & 10 / 5 \\
Warm Restart / Maximum Anchor Age ($\rho_{\mathrm{warm}} / K$) & 0.5 / 48 \\
Inference Solver NFE & 1 NFE (Euler step) \\
Training devices & 2 GPUs \\
Pairs per GPU microbatch / global update & 8 / 32 \\
Anchors / targets per anchor per microbatch & 2 / 4 \\
Selected checkpoint / completed run & 12,000 / 18,000 updates \\
Peak Learning Rate / Warmup & $5\times10^{-5}$ / 200 steps (cosine decay) \\
Weight Decay / Gradient Clipping & 0.01 / 1.0 \\
Loss Weights ($\mathcal{L}_{\mathrm{KV}} / \mathcal{L}_{\mathrm{act}} / w_{\mathrm{grip}}$) & 1.0 / 2.0 / 0.1 \\
Latency Age Support & 1--10 (80\%); 11--45 (20\%) \\
Training runtime & PyTorch 2.8.0 / CUDA 12.8 \\
Simulation Environment & robosuite 1.4.1 / MuJoCo 2.3.7 \\\bottomrule
\end{tabularx}
\end{table}

Simulation uses robosuite 1.4.1 with MuJoCo 2.3.7 as summarized in Table~\ref{tab:config}.

The transformer block follows adaptive layer-normalization conditioning used in diffusion transformers~\citep{peebles2023dit}. Let $x$ denote the projected KV tokens and $c$ the encoded execution history. Each transformer block computes
\begin{align}
(\gamma_1,\beta_1,\alpha_1,\gamma_2,\beta_2,\alpha_2)&=M(c),\\
h&=\mathrm{LN}(x)\odot(1+\gamma_1)+\beta_1,\\
x'&=x+\alpha_1\odot\mathrm{Attention}(h),\\
h'&=\mathrm{LN}(x')\odot(1+\gamma_2)+\beta_2,\\
x''&=x'+\alpha_2\odot\mathrm{MLP}(h').
\end{align}
Attention masks exclude inactive tokens. The compact updater operates on 512 visual and 32 language slots, retaining post-RoPE key coordinates~\citep{su2024roformer}, and restores the 968-slot expert layout after prediction. Masked slots receive zero residual. Both modulation heads and the final residual projection are zero-initialized. Independently of initialization, an explicit $\amsbb{I}[k>0]$ gate zeroes the residual at age zero so fresh anchors remain unchanged throughout training and deployment. Input K/V tokens use per-layer RMS scales $\sigma^{\mathrm{KV}}_{X,l}$; output residuals and KV regression use separate delta RMS scales $\sigma^\Delta_{X,l}$. All normalization statistics are fitted on training data and inherited by subsequent checkpoints.

Physical tick $t$, wall-clock decision time $q_t$, and flow time $\rho$ are distinct. The latest available anchor has capture tick $a(t)$ and staleness age $k=t-a(t)$. The history $\mathcal H_{a:t}$ pairs states $s_{a+1},\ldots,s_t$ with executed commands $u_a,\ldots,u_{t-1}$. The receding endpoint prior is $\cache_t$ with restart time $\rho=0.5$.

\subsubsection{Exact History and Projection Interfaces}
\label{sec:exact_interfaces}
The simulation history conditioner uses a fixed-length chronologically ordered vector so runtime latency remains invariant to elapsed steps. Let $d_s=8$, $d_u=7$, history capacity $K=48$, $k=t-a$, and $\bar s=(s-\mu_s)\oslash\sigma_s$. With chronological zero-padded rows,
\begin{equation}
\begin{split}
h_{a:t}=\big[&k/K,\ \bar s_a,\ \bar s_t,\ (s_t-s_a)\oslash\sigma_s,\\
&\operatorname{vec}(\Delta S),\ \operatorname{vec}(U)\big],
\end{split}
\end{equation}
where $\Delta S_i=(s_{a+i+1}-s_a)\oslash\sigma_s$ and $U_i=u_{a+i}$ for $0\le i<k$, and entries for $i\ge k$ are zero-padded. State increments are flattened before executed commands, with actions retaining native control units for a combined dimensionality of $1+3d_s+K(d_s+d_u)=745$. Each post-step state pairs with its preceding issued command, excluding speculative future plans; anchor ages exceeding capacity trigger a synchronous visual refresh.

The history MLP maps $745\!\to\!768\!\to\!384$ with GELU activations. For prefix token $j$, keys and values across all 18 layers are standardized and concatenated:
\begin{equation}
q_j=\operatorname{concat}\!\left(
\{K_{\ell j}/\sigma^{\rm KV}_{K,\ell}\}_{\ell=1}^{18},
\{V_{\ell j}/\sigma^{\rm KV}_{V,\ell}\}_{\ell=1}^{18}\right).
\end{equation}
An affine $9216\!\to\!384$ projection, learned positional embeddings, and a history projection initialize the token stream across six AdaLN blocks, followed by LayerNorm and a $384\!\to\!9216$ output projection. Reshaping the predicted residual back into layer-wise K/V tensors, rescaling by $\sigma^\Delta$, and gating by active tokens and $\amsbb{I}[k>0]$ prepares the additive update to $z_a$. Token attention is bidirectional across prefix slots, whereas causal conditioning refers to indexing strictly realized states and executed commands up to tick $t$.

The frozen $\pi_{0.5}$ backbone~\citep{black2025pi05} exposes a 968-slot prefix comprising two active 256-token camera views, one masked third camera view, and up to 200 text tokens. In simulation, $W_\theta$ operates on the 512 active visual slots and first 32 language slots and scatters predicted residuals back into the native 968-slot layout, injecting proprioceptive history without enabling VLM state tokenization.

\begin{algorithm}[!htbp]
\caption{\method at a replanning boundary}
\label{alg:dual}
\begin{algorithmic}[1]
\Require Frozen $B_\phi$, $v_\psi$; updater $\wm$; consumed length $C$; restart $\rho_{\mathrm{warm}}$; age limit $K$.
\State Initialize real anchor $(a,z_a)$; cache $Y\leftarrow\varnothing$.
\Loop
\State Append observed states and executed commands.
\State Adopt the latest completed real anchor $(a,z_a)$.
\State Refresh synchronously if $t-a>K$.
\State $\hat z_t\leftarrow\wm(z_a,\hist_{a:t})$.
\State Sample fresh $\epsilon\sim\mathcal{N}(0,I)$.
\If{$Y\ne\varnothing$}
\State $Q\leftarrow\mathrm{ShiftRepeat}(Y,C)$; $\rho\leftarrow\rho_{\mathrm{warm}}$.
\State Rebase $Q$ when state-relative coordinates apply.
\State $x_\rho\leftarrow(1-\rho)Q+\rho\epsilon$.
\Else
\State $\rho\leftarrow1$; $x_\rho\leftarrow\epsilon$.
\EndIf
\State $Y\leftarrow x_\rho-\rho v_\psi(x_\rho,\rho\mid\hat z_t)$.
\State Execute the first $C$ commands; record each step.
\EndLoop
\end{algorithmic}
\end{algorithm}


\FloatBarrier
\subsection{Action Distillation}
\label{sec:supp_objective}
Output matching follows the policy-distillation principle~\citep{rusu2016distillation}, applied here to the input memory of an unchanged expert. The teacher and student queries share frozen action expert $v_\psi$, current-state input, initial flow latent $x_\rho$, and flow time $\rho$. The map $D_\psi$ denotes its one-step control endpoint, as in Section~\ref{sec:objective}. The teacher reads fresh KV memory $z_t$; the student reads predicted KV memory $\hat z_t$. Both outputs are normalized one-step endpoints restricted to the seven control dimensions. Their difference supervises the updater, while the expert parameters remain fixed.

Let $b,l,j,d$ index training pairs, layers, prefix slots, and KV channels. For $X\in\{K,V\}$, define the normalized error
\begin{equation}
E^X_{bljd}=
\frac{\hat X_{bljd}-X^\star_{bljd}}{\sigma^\Delta_{X,l}}.
\end{equation}
With prefix mask $m_{bj}$, $L$ layers and $d_z$ channels, the masked norm in Eq.~\ref{eq:total_loss} is
\begin{equation}
\mathcal L_{\mathrm{KV}}=
\frac{\sum_{X,b,l,j,d}m_{bj}(E^X_{bljd})^2}
{2Ld_z\sum_{b,j}m_{bj}}.
\end{equation}
Thus each active KV entry has equal weight after per-layer delta normalization. Input normalization $\sigma^{\mathrm{KV}}$ is used by the predictor; the regression loss uses the separate residual scale $\sigma^\Delta$.

For batch size $B$, horizon $H$, and control index $d$, the action norm is
\begin{equation}
\mathcal L_{\mathrm{act}}=
\frac{1}{7BH}\sum_{b=1}^{B}\sum_{h=1}^{H}\sum_{d=1}^{7}
w_d(\hat A_{bhd}-A^\star_{bhd})^2,
\end{equation}
where $w_{1:6}=1$ and $w_7=0.1$. The denominator is $7BH$, so downweighting the gripper does not rescale the arm terms. The final objective is $\mathcal L_{\mathrm{KV}}+2\mathcal L_{\mathrm{act}}$. Targets are detached, but the student expert query remains in the differentiation graph. Gradients with respect to its KV input train only the updater parameters $\theta$.

Fix the shared action query and let $e=\hat z_t-z_t$. A first-order expansion of the expert gives
\begin{equation}
D_\psi(z_t+e)-D_\psi(z_t)=J_De+o(\|e\|).
\end{equation}
For one pair, let $\Omega=I_H\otimes\operatorname{diag}(w)$ and vectorize the chunk. Then
\begin{equation}
\mathcal L_{\mathrm{act}}=
\frac{e^\top J_D^\top \Omega J_De}{7H}+o(\|e\|^2).
\end{equation}
The positive-semidefinite form $J_D^\top \Omega J_D$ assigns larger penalty to memory errors that change the expert's action response. Its nullspace may contain errors that KV regression still penalizes. This is a local interpretation of output matching: training backpropagates the action loss and never constructs the Jacobian explicitly.


\paragraph{Distillation gradient path and warm-query proxy.}
For one training sample, write $\hat A=D_\psi(W_\theta(z_a,h))$ and detach teacher target $A^\star=D_\psi(z_t)$. Holding robot state and flow initializer $(x_\rho,\rho)$ fixed gives
\begin{equation}
\nabla_\theta\mathcal L_{\rm act}
=\frac{2}{7H}J_{W_\theta}^{\top}J_D^{\top}\Omega(\hat A-A^\star).
\end{equation}
Although expert weights remain frozen, gradients flow with respect to the predicted KV inputs into the action expert. Sharing $(x_\rho,\rho)$ between teacher and student isolates conditioning-induced action differences from sampling noise. Warm training queries are drawn with probability 0.5 whenever a preceding boundary exists by shifting and renoising the fresh-memory endpoint at $t-C$; this provides a tractable offline proxy for recursive warm-started deployment while KV regression and action distillation retain their separate microbatch normalizations.

\subsection{Causality and Local Error Guarantees}
\label{sec:causal_proofs}
\paragraph{Information available at decision tick $t$.}
Let $\mathcal F_t$ denote the filtration generated by observations acquired up to tick $t$, commands issued strictly before $t$, completed visual encodings, and sampler randomness drawn by $t$. An asynchronous anchor becomes eligible for fast updates only upon encoding completion and only for capture ticks $a \le t$.
\paragraph{Causal measurability.}
Because the history buffer appends only realized state--action transitions $(s_{i+1},u_i)$, visual anchor publication is atomic, and $W_\theta$ reads only published real anchors without ever feeding predicted KV back as an anchor, both $\hat z_t$ and $A_t$ are $\mathcal F_t$-measurable by induction over execution steps. This guarantees causal information flow in asynchronous execution, whereas contact behavior still rests on model accuracy.
\paragraph{Exact fresh-anchor identity.}
Because the residual branch is gated as $\hat z=z_a+\amsbb{I}[k>0]\,\delta z_t$, the identity $\hat z=z_a$ holds identically whenever $k=0$, and zero-initialized residual projections further guarantee identity across all ages at initialization.
\paragraph{Conditional action-error bound.}
For a fixed state, flow noise, and initializer, assume $D_\psi$ is differentiable along the line segment connecting $z_t$ to $\hat z_t=z_t+e$ with bounded weighted Jacobian norm $\sup_{u\in[0,1]}\|\Omega^{1/2}J_D(z_t+ue)\|_2\le L$. By the fundamental theorem of calculus,
\begin{equation}
\Omega^{1/2}[D_\psi(z_t+e)-D_\psi(z_t)]
=\int_0^1 \Omega^{1/2}J_D(z_t+ue)e\,du,
\end{equation}
which yields the local bound $\mathcal L_{\rm act}\le L^2\|e\|_2^2/(7H)$. Consistent with the quadratic sensitivity model in Appendix~\ref{sec:supp_objective} and Eq.~\ref{eq:sensitivity}, minimizing $\mathcal L_{\mathrm{act}}$ aligns prediction error with control sensitivity beyond isotropic KV distance.

\subsection{PFWS Alignment and One-Step Integration}
\label{sec:pfws_details}
Consumed-tail alignment shifts the remaining $H-C$ steps forward and repeats the terminal action $C$ times:
\begin{equation}
Q_{t,h}=A_{t-C,\min(h+C,H-1)},\qquad h=0,\ldots,H-1.
\end{equation}
Partial renoising interpolates $Q_t$ with fresh Gaussian innovation $\epsilon\sim\mathcal N(0,I)$ at $x_\rho=(1-\rho)Q_t+\rho\epsilon$, followed by the frozen-expert Euler update $x_0^{\rm E}=x_\rho-\rho v_\psi(x_\rho,\rho\mid\hat z_t)$ integrating from flow time $\rho$ toward the action manifold~\citep{lipman2023flow} ($\rho=1$ at cold episode onset). Hardware execution applies shift--repeat updates on rebased state axes.

\paragraph{Local Euler discretization error.}
Fix the conditioning memory and initializer, and assume the exact ODE velocity $g(r)=v_\psi(x(r),r\mid z)$ is differentiable on $[0,\rho]$ with $\|g'(r)\|\le M$. Comparing the single Euler step $x_0^{\rm E}$ against the exact integral $x_0=x_\rho-\int_0^\rho g(r)\,dr$ gives
\begin{equation}
\begin{split}
\|x_0^{\rm E}-x_0\|
&\le\int_0^\rho\|g(\rho)-g(r)\|\,dr\\
&\le\int_0^\rho M(\rho-r)\,dr=\tfrac12M\rho^2.
\end{split}
\end{equation}
Because cold and warm initializers traverse distinct regions of the flow field with different local curvature bounds $M$, Appendix~\ref{sec:flow_geometry} directly measures dense streamline geometry and one-step truncation error on evaluation contexts.

\paragraph{Exogenous scene identifiability.}
If two environments yield identical anchor and proprioceptive-history inputs while an external object moves independently between camera frames, any deterministic execution-conditioned predictor produces the same $\hat z_t$ for distinct true memories $z_t^{(1)}$ and $z_t^{(2)}$, giving the triangle-inequality lower bound
\begin{equation}
\max_{i\in\{1,2\}}\|\hat z_t-z_t^{(i)}\|
\ge\tfrac12\|z_t^{(1)}-z_t^{(2)}\|.
\end{equation}
Consequently, execution history captures robot-induced and contact-mediated state evolution, whereas purely exogenous visual changes between frames are resolved upon the next synchronous visual anchor refresh.

\subsection{Clean Expert-Only Training}
\label{sec:supp_expert}
The updater's residual output and optimizer moments are initialized to zero. Normalization is fitted on the 341 training demonstrations and is preserved for every checkpoint. A fixed holdout of 38 demonstrations and 100 original-policy trajectories supplies 320 validation pairs, balanced across the ten tasks with demonstration/expert probabilities 20/80. Validation examples remain excluded from training and normalization.

The four expert training pools contain 600/300/300/300 trajectories and are all available at update zero. Demonstrations receive sampling probability 0.20 and these pools receive 0.32/0.16/0.16/0.16. We sample an eligible episode within the selected task and source, then an anchor and causal targets. Every retained row is available at stride one. Expert recordings retain every executed simulator action; ages in preprocessed demonstrations index retained rows. An epoch count would not describe this sampler: the selected 12k updates produce 384,000 target-pair draws, with temporal pairs sampled repeatedly from trajectories.

Table~\ref{tab:config} specifies the microbatch construction and optimizer. The cosine learning-rate schedule repeats every 6k updates, while optimizer moments, RNG, and pools remain continuous up to the selected 12k weights.

Temporal sampling follows the age distribution in Table~\ref{tab:config}. Each target uses executed state--action pairs and freshly encoded target memory. Half of eligible action queries use a shifted fresh-teacher endpoint from the preceding boundary, whereas cold queries share random initializers across teacher and student to isolate memory effects.

\FloatBarrier
\subsection{Checkpoint Selection and Hardware Protocols}
The deployed controller uses the 12k-update checkpoint trained with seed 42 from zero-residual initialization around the official frozen $\pi_{0.5}$ LIBERO weights, evaluated across 500 official LIBERO starts~\citep{liu2023libero} at seed 7 after selecting the training-budget peak in Table~\ref{tab:cohorts}.

\begin{table}[!htbp]\centering\footnotesize
\caption{\textbf{Hardware and evaluation protocols.} Model timing measures inference with resident inputs; fixed-clock and unpaced evaluations include environment feedback. Literature references retain their original configurations.}
\label{tab:hardware}
\setlength{\tabcolsep}{4pt}
\begin{tabularx}{\textwidth}{lll>{\raggedright\arraybackslash}X}\toprule
Study & Hardware & Software / inference & Protocol\\\midrule
Updater training & 2 RTX 5090 & Torch 2.8 / CUDA 12.8 & Batch 32, selected 12k\\
Checkpoint sweep & 4 A100 workers & Torch 2.6 / CUDA 12.4 & $H=10,C=5$, 500 starts/arm\\
Objective control & 2 RTX 5090 & Torch 2.8 / CUDA 12.8 & Paired objectives on each task/GPU\\
Module controls & 1--4 A100 per task shard & Torch 2.6 / CUDA 12.4 & Four arms on each task/GPU\\
Model-only capacity & 1 RTX 5090 & BF16, compiled & 15 warmups, 100 timed calls\\
Deadline grid & 2 A100 & Separate policy / renderer & $20/50$\,Hz, $C=1/5$\\
Backbone deadline references & 2 PRO6000 & Separate policy / renderer & 20\,Hz, $C=5$, 10 starts/arm\\
LIBERO-Plus robustness & A100 / RTX 5090 / PRO6000 & Compiled; paired GPU/variant & 2,519 variants/arm\\
FR3 updater pretraining & 2 RTX 5090 & Torch 2.8 / CUDA 12.8 & Batch 32, 6k updates\\
FR3 time adaptation & 2 L40S & Frozen original expert & Batch 32, 4k updates; select 2.5k\\
Physical FR3 evaluation & 1 RTX 5090 & Compiled; front/wrist D405 & $H=10,C=2$, 100\,Hz ceiling\\\midrule
RTC reference & RTX 4090 & 5-step inference & Unguided / guided\\
Latent Bridge reference & A100 & Control cycle & Mean cycle latency\\
EagleVLA reference & Thor / Orin & Reaction latency & Device-specific values\\\bottomrule
\end{tabularx}\end{table}

GR00T~\citep{nvidia2026grootlibero} native success uses one A100 with its official Torch 2.9 / CUDA 12.8 runtime; its model timing uses RTX 5090. The original-policy native references use standard A100 evaluation, and their timing is separately measured on RTX 5090. A100 success measurements are not used to derive 5090 speed. The native experiment has a 520-action horizon and no imposed wall-clock command deadline. The deadline grid has a 26-second clock budget, with 520 ticks at 20\,Hz or 1,300 ticks at 50\,Hz; it is a capability test rather than the native LIBERO benchmark.

Model timing uses resident inputs and \texttt{max-autotune-no-cudagraphs}, with 100 synchronized calls after 15 warmups. Original one-step and synchronous PFWS native references use eager inference; the matched module controls use the same compiled backend within each task. Published RTC, Latent Bridge and EagleVLA timings retain RTX 4090, A100 and Thor/Orin hardware, respectively, while OpenVLA-OFT~\citep{kim2025fine} executes deterministically.


\FloatBarrier
\section{Discussion, Analytical Scope, and Limitations}
\label{sec:discussion}

\subsection{What Is Added Beyond Asynchronous Scheduling?}
Asynchronous execution by itself can overlap perception and command supply, as demonstrated by RTC~\citep{black2025rtc}. While feature or KV prediction has precedent in Latent Bridge~\citep{liu2026latentbridge} and tail reuse in VLA-Rolling~\citep{he2026vlarolling}, our learned contribution conditions a published real visual anchor on ordered execution transitions and supervises that update through the frozen expert's action sensitivity, paired with training-free tail continuation under PFWS. Writing the frozen one-step endpoint as $F_\psi(\cache,z)$ while suppressing fixed state, noise, and restart time, adding and subtracting $F_\psi(\cache,z^\star)$ around reference inputs $(\cache^\star,z^\star)$ yields
\begin{equation}
\begin{split}
F_\psi(\cache,\hat z)-F_\psi(\cache^\star,z^\star)
={}&[F_\psi(\cache,\hat z)-F_\psi(\cache,z^\star)]\\
&+[F_\psi(\cache,z^\star)-F_\psi(\cache^\star,z^\star)].
\end{split}
\end{equation}
The first term isolates the conditioning discrepancy targeted by action distillation at a shared initializer, and the second captures the effect of the prior initializer under fixed conditioning, explaining why a smooth prior cannot replace current memory and why KV reconstruction alone does not constrain boundary sampling. Matched controls in Table~\ref{tab:controls} test module necessity within this implementation without claiming additive task-success gains or superadditive interaction over a complete multi-factor runtime grid.

\subsection{What Do the Theoretical Arguments Establish?}
Appendix~\ref{sec:causal_proofs} proves causal information flow, exact age-zero identity, and a conditional local action-error bound. It does not prove task success, global stability, or universal transfer across control frequencies. The history interventions are tests of a trained model, not separately trained reduced-input baselines. The KV-only objective comparison, in contrast, changes supervision at a matched training budget.

\subsection{Does PFWS Always Improve Performance?}
No. Its synchronous paired success difference is not statistically decisive ($p=0.1173$), despite reduced boundary reversals. Across the LIBERO-Plus perturbation catalog, PFWS raises synchronous success from 78.36\% to 80.15\% and asynchronous success from 76.38\% to 78.09\%, leaving an asynchronous deficit of 1.98 percentage points without reuse and 2.06 percentage points with reuse. Because that gap already appears without PFWS, aggregate scores establish a within-controller benefit from trajectory reuse alongside perturbation families where reuse hurts rather than greater absolute perturbation robustness or disentangled memory-and-scheduling errors. Dense flow paths are already low-curvature in both conditions, so PFWS shortens the correction interval rather than learning a new field or proving smaller training-distribution shift. Full perturbation-family results and the matched flow-interval comparison are retained below.

\subsection{How Should Pretrained Comparisons Be Interpreted?}
Public checkpoints retain their released action heads and inference budgets. Latent Bridge~\citep{liu2026latentbridge} is the available R0 model, not an unreleased or retrained R1, while VLA-Rolling~\citep{he2026vlarolling} retains its LeRobot backbone and longer horizon alongside its paired backbone control. Native success has an action budget, whereas paced success limits wall time under distinct hardware scopes. Model-only timing measures resident-input forward latency (60.01, 36.00, 35.54, and 11.91\,ms across ten-step, one-step, synchronous PFWS, and shared-GPU \method), so fewer flow evaluations explain part of the 5.04$\times$ capacity ratio and removing visual encoding from the fast path accounts for the remaining reduction while the world model preserves conditioning rather than accelerating compute. By contrast, single-GPU timed execution includes observation transport, co-located rendering, and executor guards, producing smaller realized gains of $3.28/2.57=1.28$ in replans/s and $16.60/13.02=1.27$ in supplied actions/s. RTC's separate-GPU 93/100 asynchronous evaluation appears in Table~\ref{tab:rtc_isolated_execution} outside the co-located single-GPU ranking, while the teaser radar plot depicts qualitative trade-offs rather than normalized axis scores, isolating a 12.52$\times$ speedup in Table~\ref{tab:final_timing}.

\subsection{Selection, Physical Transfer, and Failure Cases}
The evaluation seed and updater recipe were chosen after earlier endpoint comparisons, and the selected 12k checkpoint follows a training-budget sweep, making reported uncertainty conditional on that selection. Separately reset physical scenes are not exact outcome pairs, and offline timestamp-conditioned adaptation from processed existing trajectories adjusts the updater rather than deploying an unchanged 10\,Hz model. Held-out task labels denote tasks unseen by the updater without certifying absence from backbone pretraining. Successful duration ends at the operator verdict, HTTP dispatch does not verify actuator consumption, and whole-system hardware comparisons co-vary realized dispatch frequency with controller inference speed rather than isolating frequency-matched module effects.

\subsection{Is This a General Dynamics or Future-Video Model?}
No. The predictor estimates current frozen-expert conditioning rather than synthesizing future pixels, predicting counterfactual rewards, or identifying physical contact dynamics. Temporal causality follows the information filtration in Appendix~\ref{sec:causal_proofs}, and the indistinguishability bound in Appendix~\ref{sec:pfws_details} shows why execution history alone cannot resolve exogenous scene changes without new visual anchors. Empirical prediction quality between anchors is therefore evaluated rather than implied by causal measurability, age-stratified curves reflect sampled history probes rather than live deployment age distributions, and RGB-colored KV panels depict shared-basis PCA projections rather than reconstructed camera views.

\subsection{Is Direct KV Prediction the Novelty?}
Direct prediction alone is not the distinction: Latent Bridge already includes a $\pi_{0.5}$ KV bridge~\citep{liu2026latentbridge}. Our proposed combination is an ordered record of executed transitions, prediction anchored at the latest published real memory, and supervision through the unchanged action expert. The prior action chunk $\cache_t$ initializes PFWS; it is not supplied as an unexecuted trajectory to the history encoder. The history interventions and matched objective control isolate these design choices more directly than a comparison of aggregate success across independently trained public checkpoints.

\subsection{Could Extra Training or the Sampling Rule Explain the Gain?}
The updater adds learned parameters and training data; freezing the backbone does not make the entire controller training-free. Public-method comparisons do not equalize training budgets and do not imply sample-efficiency advantages. Within our ablations, the same data, initialization, budget, and validation pairs isolate the action-loss contribution, while inference-time module toggles keep the learned checkpoint fixed. Every retained training row is eligible with stride one; $C$ controls deployed chunk consumption and cache shifting, not dataset subsampling. Validation splits and normalization stay separate from evaluation.

\subsection{Does the Teacher Leak Future Evaluation Information?}
Offline supervised pairs contain a real target encoding at the later recorded state. This target supplies a detached label, like standard teacher supervision~\citep{rusu2016distillation}; it is never available to the deployed updater. At inference, only a completed real prefix and transitions already issued and observed are admissible. The warm training prior comes from a previous fresh-teacher endpoint and therefore differs from recursive deployment, an acknowledged training-distribution approximation rather than hidden online oracle access.

\subsection{Why Not Simply Increase the Command Rate?}
Higher-rate resending can repeat an unchanged target without computing a new plan, so distinct action-node first dispatches are reported separately from plan adoption, model capacity, and response latency. Under uninterrupted steady execution consuming $C$ nodes per plan, realized replanning is bounded by both fast computation capacity and node-consumption rate divided by $C$, alongside startup, gripper barriers, and horizon exhaustion. Multi-plan overlap within one System~2 window is measured directly from monotonic timestamps rather than inferred from latency ratios; in the archived 20-episode $C=1$ simulation probe, 250 of 4,874 completed slow windows contain multiple fast plans first used in that window (5.13\%), demonstrating workload-dependent overlap without claiming multiple replans in every pass or identical ratios at $C=5$.

\subsection{Does Timestamp Conditioning Make \texorpdfstring{10\,Hz}{10 Hz} Data Native High-Rate Data?}
No. The physical controller uses a separately adapted updater while preserving the original expert and action units. Its $K=24$, seven-state/seven-action base vector has 358 entries. The shared training/inference encoder appends the 24 measured transition durations divided by 0.1\,s and their sum on that same scale, giving 383 inputs. These duration features expose irregular sampling; they neither create previously unobserved contact events nor prove rate invariance. Offline processing of existing trajectories is not equivalent to collecting independent 100\,Hz measurements. The physical results establish performance for the evaluated 100\,Hz ceiling and actual measured rates, not for arbitrary 200\,Hz or faster execution.

\subsection{Could Synchronization, Smoothing, or Pair Selection Produce the Motion Result?}
Both controllers share the pose-arrival and gripper-completion executor. During a gripper barrier, arm progression and policy history/cache updates pause; resumption requires fresh observation and real memory. Active-rate statistics remove those waits, whereas successful-trial duration retains them. These are complementary quantities rather than interchangeable notions of speed. The speed-CV and acceleration-RMS statistics use the same unsmoothed 100\,ms EEF samples for all outcomes; display-only filtering does not enter the numbers. The new illustration pairs are selected first by complete phase-path similarity and then initial layout, without ranking the size of our advantage. Requiring complete grasp--place phases restricts visual panels to that subset, while the 138-trial table covers all outcomes.

\subsection{Cohort Isolation, Statistical Evidence, and Uncertainty}
\label{sec:review_evidence}
Table~\ref{tab:cohorts} separates three independent evaluations whose counts are interpreted solely within each cohort rather than pooled into a single estimate: the selected native checkpoint (469/500 vs.\ 456/500), matched module ablations (467/500 full, 460/500 synchronous PFWS, 465/500 independent synchronous PFWS), and matched objective training (464/500 vs.\ 428/500). For discordant paired starts in Table~\ref{tab:cohorts}, where $w$ counts starts solved only by the tested arm and $l$ starts solved only by its reference, the exact two-sided McNemar test conditions on $n_d=w+l$:
\begin{equation}
p=\min\left(1,\;2\sum_{j=0}^{\min(w,l)}\binom{n_d}{j}2^{-n_d}\right).
\end{equation}
Applying this test to the archived discordant counts summarizes pairwise strength within each cohort:
\begin{center}\small
\begin{tabular}{lrrr}\toprule
Comparison & $w/l$ & Gain (pp) & Exact $p$\\\midrule
Memory prediction & 80/14 & 13.20 & $2.12\times10^{-12}$\\
PFWS, full controller & 24/17 & 1.40 & 0.3489\\
Action distillation & 54/18 & 7.20 & $2.57\times10^{-5}$\\
PFWS, synchronous & 26/15 & 2.20 & 0.1173\\\bottomrule
\end{tabular}
\end{center}
Memory prediction and action distillation remain below $0.05$ under a four-test Bonferroni threshold, whereas the PFWS success comparisons remain statistically inconclusive and rest primarily on boundary flip reduction. For aggregate totals, descriptive 95\% Wilson intervals span 91.33--95.60\% (469/500) vs.\ 88.39--93.38\% (456/500) in native simulation, and 77.03--92.98\% (60/69) vs.\ 60.95--81.61\% (50/69) in physical trials; marginal totals do not yield paired test statistics, and physical intervals treat separately reset trials without modeling task or block correlation. Full LIBERO-Plus perturbation families retain localized losses alongside gains, leaving statistical inference conditioned on the evaluated starts and selected checkpoints rather than universal dominance across seeds or operators.

\subsection{Implementation Checks and Unresolved Warnings}
RTC uses a precision-preserving compiled input-gradient backend. Its numerical audit retained a rare decoded-action discrepancy in one of 120 independent contexts; the compiled implementation is therefore evaluated empirically rather than claimed to be numerically identical to the reference. The 500-start native result remains 459/500 (91.80\%). The corrected 100-start asynchronous evaluation reaches 93/100 (93.00\%); its scheduling and hardware allocation are specified in Appendix~\ref{sec:pretrained_protocol}. Rolling's native simulator warnings are audited independently of policy success. Replaying all 1,000 paired-arm initializations and the first three recorded commands localizes both warnings to task~5, initial state~18, on the first command. Simulator time resets from 0.500 to 0.002\,s in both arms. Both original episodes were failures and remain failures in the 500-start denominators; the audit changes no outcome labels.

\subsection{Limitations, Applicability to Single-System VLAs, and Future Directions}
\label{sec:limitations}
Adapting the memory predictor to another VLA requires a compatible cached prefix conditioning interface and training against that policy's frozen action expert, while PFWS as formulated requires a flow velocity head. Other benchmarked backbones remain reference baselines without controller integration, leaving cross-architecture transfer and alternative action heads untested.

In execution, the controller cold-starts at episode onset and clears prior tails with fresh anchors across physical gripper barriers or upon exceeding history capacity, yet does not run a learned confidence test against sudden viewpoint or scene shifts. Future extensions therefore include innovation- or entropy-triggered visual refreshes for exogenous motion in Appendix~\ref{sec:pfws_details}, adaptive flow interpolation $\rho_t$ between warm reuse and cold resets, and high-frequency visuo-tactile sensing during dexterous contact.
\FloatBarrier
\section{Supplementary Experimental Results}
\label{sec:supp_results}
\subsection{Per-Task Native LIBERO-10 Evaluation}
\begin{table}[!htbp]\centering\small
\caption{\textbf{Native successes by task.} Each task uses 50 official starts at seed 7 and $H=10,C=5$. Bold marks the higher success count, including ties.}
\label{tab:supp_final}
\setlength{\tabcolsep}{3pt}\input{tables/final_by_task.tex}\end{table}
Table~\ref{tab:supp_final} localizes the native gain: both-moka-pot and book placement improve, while several tasks approach ceiling. This heterogeneity complements the aggregate competence test in the main text.

Synchronous one-step PFWS uses the same 500 official task--initial-state identities, seed 7, $H=10,C=5$, restart time 0.5 and 520-action horizon. This reference uses eager Torch 2.8/CUDA 12.8; the selected controller uses compiled Torch 2.6/CUDA 12.4 on A100. Table~\ref{tab:controls} fixes the backend and allocated GPU within each task to isolate memory prediction from synchronous PFWS.

\begin{table}[!htbp]\centering\small
\caption{\textbf{Matched evaluation cohorts.} Each comparison uses 500 starts per arm. Wins/losses count starts solved only by the full arm or only by its reference. Native, module, and objective comparisons are separate runs.}
\label{tab:cohorts}
\setlength{\tabcolsep}{7pt}
\begin{tabular}{lrrr}\toprule
Comparison & \method{} successes $\uparrow$ & Wins $\uparrow$ & Losses $\downarrow$\\\midrule
\rowcolor{tablerow}Native checkpoint run & 469 & --- & ---\\
w/ vs.\ w/o latent WM & 467 & 80 & 14\\
w/ vs.\ w/o PFWS & 467 & 24 & 17\\
w/ vs.\ w/o action loss & 464 & 54 & 18\\
Frozen policy: PFWS vs.\ cold & 465 & 26 & 15\\\bottomrule
\end{tabular}\end{table}


\FloatBarrier
\subsection{Training Budget and Module Results}
\begin{table}[!htbp]\centering\small
\caption{\textbf{Validation error across training budgets.} All checkpoints share 320 held-out pairs. Weighted action MSE weights the gripper by 0.1, unweighted MSE treats all seven axes equally, and bold or underline marks the top two errors.}
\label{tab:validation}
\input{tables/final_validation_table.tex}
\end{table}

\begin{figure}[!htbp]\centering
\begin{subfigure}[t]{0.485\textwidth}
\centering
\includegraphics[width=\linewidth]{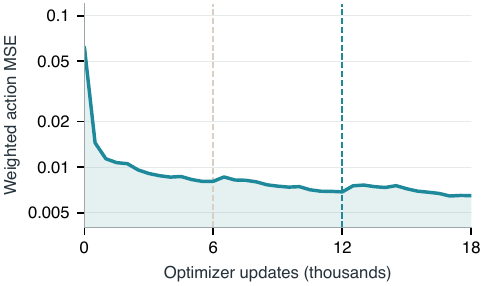}
\caption{Weighted action error on fixed validation pairs.}
\label{fig:supp_training_a}
\end{subfigure}\hfill
\begin{subfigure}[t]{0.485\textwidth}
\centering
\includegraphics[width=\linewidth]{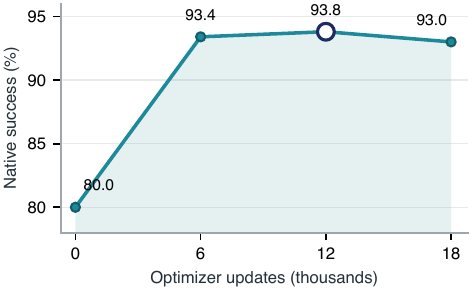}
\caption{Native success on 500 fixed initial states.}
\label{fig:supp_training_b}
\end{subfigure}
\caption{\textbf{Training loss and control performance.} (a) Weighted action error across three cycles. (b) Native success at matched update budgets, with open markers highlighting 12k.}
\label{fig:supp_training}\label{fig:training_success}\end{figure}

Figure~\ref{fig:supp_training} and Table~\ref{tab:validation} show why optimization progress and control quality require separate checks. Weighted loss continues to decrease after the selected checkpoint, whereas gripper error increases and native success declines. The lower gripper weight permits this disagreement. Table~\ref{tab:cohorts} separates this native checkpoint run from matched controls.

\begin{table}[!htbp]\centering\small
\caption{\textbf{Module ablations by task.} Cells count successes out of 50 paired starts. All arms share the assigned device, compiled runtime, seed 7, and $H=10,C=5$ within each task.}
\label{tab:controls}
\setlength{\tabcolsep}{2pt}\input{tables/final_module_tasks.tex}\end{table}

Table~\ref{tab:controls} resolves the module comparison by task; Table~\ref{tab:cohorts} gives its discordant pairs. Removing memory prediction causes a broader competence loss than removing PFWS, complementing the distinct boundary-continuity effect in Table~\ref{tab:ablation_main}. Module flip rates pool boundary transitions across episodes; the independent synchronous assay below weights episodes equally.


\FloatBarrier
\subsection{Action-Relevant Prediction Assay}
Table~\ref{tab:supp_age} reports the final checkpoint's validation errors by anchor age, measuring stale and updated KV against fresh target KV on the same windows and flow latents. Action MSE gives equal weight to all seven output dimensions, whereas training error weights the gripper by 0.1, making both metrics complementary.
\begin{table}[!htbp]\centering\small
\caption{\textbf{Memory and action error by anchor age.} Normalized KV error and unweighted action MSE use fresh-memory targets. Bold marks the lower error in each stale/updated pair; the final bin contains ages 40--45.}
\label{tab:supp_age}
\setlength{\tabcolsep}{3pt}
\input{tables/final_age_table.tex}\end{table}

The input interventions use fixed weights, pair seed 100042, and noise seed 200042 on held-out demonstrations and expert trajectories. Replacing history with age or endpoint states tests the value of execution evidence; row shuffling tests its ordering. These are inference interventions, not retrained reduced-input models.

The objective control starts from the same zero-update state and uses the same training pool and update budget, changing only the action-loss coefficient. The main ablation reports representation error, action error, and native success together. Their reversed ranking under KV-only training supports supervising latent prediction via expert actions.

\FloatBarrier
\subsection{Trajectory Correction and Boundary Smoothness}
Appendix~\ref{sec:pfws_details} specifies the alignment and one-step correction. The independent synchronous test uses fresh visual memory, removing the learned updater from the comparison. We measure command changes at chunk boundaries and within chunks, using an equal weight for each episode. Table~\ref{tab:smoothness} further isolates starts solved by both policies to test success-set confounding.

\begin{table}[!htbp]\centering\small
\caption{\textbf{PFWS continuity under synchronous fresh vision.} Command-change metrics are averaged equally across episodes. Boundary excess is the boundary-minus-interior flip rate. Paired bootstrap intervals report changes in excess for all 500 starts and the 439 starts solved by both arms.}
\label{tab:smoothness}
\setlength{\tabcolsep}{7pt}
\begin{tabular}{lrr}\toprule
Metric & Cold & PFWS\\\midrule
\multicolumn{3}{l}{\textit{All 500 paired starts}}\\
Translation boundary $L_2$ & 0.08566 & \best{0.08221}\\
Rotation boundary $L_2$ & 0.01775 & \best{0.01742}\\
Boundary flips (\%) & 2.57 & \best{1.36}\\
Interior flips (\%) & 1.69 & \best{1.39}\\
Boundary excess (pp; $\to0$) & +0.88 & \best{\textminus0.03}\\
\multicolumn{3}{l}{Excess change: $-0.91$ pp; 95\% CI $[-1.23,-0.62]$}\\\midrule
\multicolumn{3}{l}{\textit{439 jointly solved starts}}\\
Translation boundary $L_2$ & 0.08392 & \best{0.08186}\\
Boundary flips (\%) & 2.20 & \best{1.34}\\
Interior flips (\%) & 1.67 & \best{1.40}\\
\multicolumn{3}{l}{Excess change: $-0.58$ pp; 95\% CI $[-0.86,-0.32]$}\\\bottomrule
\end{tabular}\end{table}

The concentration of reversals at handoff decreases in both cohorts. The all-start result therefore persists when conditioning on shared task completion. This localizes the continuity benefit without treating every motion statistic as a proxy for task success. The paired success outcomes appear in Table~\ref{tab:cohorts}; their McNemar test gives $p=0.1173$.

\FloatBarrier
\subsubsection{Frozen-Expert Flow Geometry}
\label{sec:flow_geometry}
To examine the initializer used by PFWS, we visualize actual model ODE trajectories and velocity directions, following the trajectory-straightness perspective of rectified flow~\citep{liu2022rectified}. The frozen $\pi_{0.5}$ expert and current real visual prefix are identical for both conditions. Native LIBERO-10 tasks 0, 4, and 8 supply contexts at executed steps 20 and 40 under a ten-step reference policy. At each of these six contexts, eight matched Gaussian seeds give 48 trajectories per condition. Each prior chunk has executed five commands before shift-and-repeat builds $\cache_t$.

We densely integrate cold starts from 1 to 0 and PFWS starts from $\rho=0.5$ to 0, using Euler step size 0.025. The cold suffix from 0.5 to 0 provides a matched-interval reference. These dense paths diagnose the field rather than change the deployed one-step method. All metrics use the 70 active normalized action coordinates, excluding padding. We measure path length divided by endpoint chord, initial-to-endpoint RMS distance, and the RMS discrepancy between a single Euler step and the corresponding dense endpoint. Figure~\ref{fig:flow_geometry} shows the preselected task-0, step-20 context; the numerical summaries below average all six contexts and eight seeds per context.

PFWS reduces mean initial-to-endpoint RMS distance from 1.115 to 0.550 and one-step endpoint error from 0.03655 to 0.02937 (19.7\%). The cold matched suffix has distance 0.550 and error 0.02970. Path/chord ratios are 1.00182 for full cold, 1.00302 for its suffix, and 1.00301 for PFWS, showing that PFWS shortens the integration interval along nearly straight frozen streamlines without reshaping the underlying flow field. This diagnostic uses a shared GPU for trajectory visualization.


\FloatBarrier
\subsection{LIBERO-Plus Robustness Benchmark}
Table~\ref{tab:plus_full} complements the main LIBERO-Plus~\citep{fei2025liberoplus} plot with absolute counts. The frozen-policy comparison tests PFWS without learned memory, while the \method toggle holds the updater fixed. Both support a positive aggregate contribution, with different family-level effects. Viewpoint and layout changes can reduce \method's benefit, so the evidence supports useful trajectory reuse rather than uniform robustness to every visual shift.

All variants remain in each arm's denominator, including 17 instructions beyond the updater's active-slot capacity. Every paired variant shares its initial state and physical device; Table~\ref{tab:hardware} records the runtimes. These measurements evaluate success, not model speed.

\begin{table}[!htbp]
\centering\small
\setlength{\tabcolsep}{3pt}
\caption{\textbf{PFWS effects across all LIBERO-Plus perturbations.} Cells report success percentage and count at $H=10,C=5$, seed 7. Top: synchronous $\pi_{0.5}$. Bottom: \method with a fixed world model. Bold marks the better cold/PFWS result at equal NFE within each category.}
\label{tab:plus_full}
\input{tables/final_plus_table.tex}
\end{table}


\subsection{Model Timing and Deadline Protocols}
\begin{table}[!htbp]\centering\small
\caption{\textbf{Model-only latency and action throughput.} Mean/P95 use 100 synchronized calls after 15 warmups, batch one, BF16. Warm inference includes the updater and PFWS at anchor age 3; the first cold plan includes visual encoding. Throughput is $1000C/\bar t$. Bold/underline rank entries within each $C$.}
\label{tab:final_timing}
\setlength{\tabcolsep}{3pt}\input{tables/final_timing_table.tex}\end{table}
Table~\ref{tab:final_timing} reports model-only timings. For model intervals $t_i$ in milliseconds, replanning capacity is $1000/\bar t$ and action-generation throughput is $C\,1000/\bar t$, where $\bar t$ is the mean synchronized latency. We measure the final checkpoint at both $C=1$ and $C=5$, retaining its actual PFWS shift, renoising, updater and expert computation. The isolated condition has no background visual worker; the concurrent condition runs the visual backbone continuously on another CUDA stream on the same card. Simulator stepping, rendering, action execution, CPU history construction, host transforms and random-noise generation are excluded from both model-only conditions.

Table~\ref{tab:deadline_detail} gives the complete deadline grid, which remeasures five controllers on the same ten starts at each of four settings (20/50\,Hz crossed with $C=1/5$): $\pi_{0.5}$ 10-step hold, 1-step hold, 10-step tail, 1-step tail, and \method hold. Hold issues zero arm displacement and preserves the last gripper command. Old-tail dispatch takes the unused command for the current tick from the preceding chunk, holding only after that tail is exhausted. All issued commands, including holds and tail commands, are recorded as actual feedback.

The scheduling audit reports pooled P50/P95 feedback-to-plan latency, $t_{\mathrm{ready}}-t_{\mathrm{observation}}$. This excludes the first cold plan and does not include simulator work preceding feedback availability. Table~\ref{tab:deadline_detail} retains scheduler-target adherence: a request is ready if it finishes by its own scheduled target. A dispatched action can meet a command deadline while originating in an older chunk. We therefore report ready requests, hold ticks, committed-prefix ticks and old-tail ticks separately. The all-start capped goal time is $\min(T_i,26)$ for a solved start and 26 seconds otherwise. Joint-success completion times, when available, use identical initial states. Policy/renderer GPU binding is checked in each run; clock overruns remain in the results.

Table~\ref{tab:clock_audit} reports the original-timestamp audit at the main operating point. Finishing within one command period and meeting the next absolute deadline differ because feedback arrives partway through that period. Catch-up dispatch can still hold even when a candidate plan is already ready. This audit separates plan readiness from command dispatch and hold. All traces, including long-tail requests and clock overruns, remain in the audit.

Continuing a chunk supplies previously predicted commands while a replacement is pending. The incoming plan is aligned to its actual decision-to-adoption lag. This control tests command supply, without implementing RTC's guided generation~\citep{black2025rtc}.

\begin{table}[!htb]\centering\footnotesize
\caption{\textbf{Fixed-clock response and dispatch.} The five-controller grid contains 200 episodes; backbone controls add 40 episodes on the same ten starts. Each episode has a 26-second budget. Overrun ticks remain in the denominators, and capped completion time assigns 26 seconds to unsolved starts.}
\label{tab:deadline_detail}
\setlength{\tabcolsep}{3.5pt}
\input{tables/final_deadline_detail.tex}
\end{table}


\FloatBarrier
\subsection{Additional Backbone Protocols}
\label{sec:additional_backbones}
GR00T N1.7 retains its official LIBERO-10 weights, 16-action horizon, and four-step head. OFT retains its released front/wrist cameras, state input, center crop, eight-action horizon, and parallel L1 head. Table~\ref{tab:deadline_detail} includes their continued-chunk dispatch controls; Table~\ref{tab:hardware} records the hardware difference.

These references test how released policies interact with hold and continued-chunk dispatch. Continuous command supply can improve without a new inference result, reinforcing the response/supply distinction in the main text.

\FloatBarrier
\needspace{18\baselineskip}
\subsection{Fixed-Clock Timestamp Audit}
\label{sec:clock_audit}
\begin{table}[H]\centering\small
\caption{\textbf{Feedback response versus scheduled dispatch.} Counts are recomputed from all 50 original traces at 20\,Hz and $C=5$. Request-level readiness and tick-level holds have distinct denominators, shown explicitly.}
\label{tab:clock_audit}
\setlength{\tabcolsep}{7pt}
\begin{tabular}{lr}\toprule
Diagnostic & Value\\\midrule
\rowcolor{tablerow}\multicolumn{2}{l}{\textit{\method}}\\
Responses within 50\,ms & 528/529\\
Responses meeting absolute target & 289/529\\
Within 50\,ms but after target & 239\\
Median remaining target budget (ms) & 28.72\\
Held ticks / eligible ticks & 270/2,944\\
Held: no completed candidate at dispatch & 245\\
Held: candidate complete at dispatch & 25\\
Held ticks at least one period late & 31\\
Longest response (ms) / resulting holds & 294.36 / 6\\\midrule
\multicolumn{2}{l}{\textit{One-step baseline with hold}}\\
Responses / range (ms) & 487 / 52.41--59.16\\
Mean pre-worker delay (ms) & 2.84\\
Mean visual-prefix selection (ms) & 47.70\\
Mean remaining computation (ms) & 5.59\\
Held ticks / eligible ticks & 487/2,946\\\bottomrule
\end{tabular}\end{table}
The response decomposition locates the one-step baseline's primary cost in visual processing, whereas the dispatch decomposition separately exposes deadline-budget loss and catch-up holds. Together, latency and hold columns isolate distinct regimes of asynchronous control.

\FloatBarrier
\subsection{Real-Robot Training and Evaluation}
\label{sec:real_robot_supp}
\paragraph{Frozen expert and timestamp-conditioned updater.}
The FR3 policy uses front/wrist images and seven-dimensional end-effector/gripper state. Its original action expert, action magnitude, and normalization remain fixed. Updater pretraining uses 205 demonstrations and 36 validation trajectories across screw sorting, water pouring, marker sorting, cup stacking, and test-tube tidying. The existing filtered 10\,Hz dataset contains 35,198 training and 6,201 validation rows across 6k pretraining updates.

For variable-rate feedback, we augment the existing filtered trajectories with timestamped histories while retaining every original action target. Actual collection timestamps determine interval duration. Continuous states are interpolated only within gaps of at most 150\,ms without gripper transitions; commands and RGB observations use causal holds. The extra history states are synthetic augmentation, rather than newly acquired high-rate measurements. Training samples use approximately 10/20/30\,ms history intervals, preserve command changes, and cap history at 24 nodes and 250\,ms. The conditioner appends the 24 measured intervals and elapsed anchor time, scaled by 100\,ms. Runtime supplies the same features from actual state samples and dispatched commands.

We adapt only the updater for 4k additional updates on two L40S GPUs with global batch 32, AdamW, learning rate $3\!\times\!10^{-5}$, 200 warmup updates, and cosine decay. The new time-input weights start at zero. The objective combines KV regression and action distillation with weights 1 and 2, respectively, and gripper weight 0.1. Fixed validation on 600 pairs selects adaptation step 2,500, giving 8,500 cumulative updater updates; the total optimization budget is 10,000. No new robot rollouts enter training. Cup stacking is updater-trained; fruit placement and vegetable sorting are held out from updater training, without assuming they are unseen by the supplied backbone.

\paragraph{Shared physical protocol.}
Both controllers use the same FR3/Robotiq hardware, two D405 cameras, RTX 5090, compiled inference, ten-action horizon, and two consumed actions per plan. Feedback is read after the preceding action has been dispatched, with a 100\,Hz sending ceiling. The baseline uses ten denoising evaluations with dual prediction and PFWS disabled; \method uses one evaluation, the timestamp-conditioned updater, and PFWS restart $\rho=0.5$. A slow pass uses the newest image pair after both camera sequences advance, up to the wrist camera's 30\,Hz limit, without an added slow-stream delay.

Before a gripper transition, the shared executor waits for the associated arm pose, then confirms fresh native gripper completion. Both model streams pause during this interval. After completion, \method builds a real anchor from new observations, clears the old action tail, and resumes from a cold plan. Both controllers share workspace, force, freshness, and first-post-grasp action guards. Trials are collected in controller blocks with manually reset objects and recorded initial images; they are not paired by identical physical object poses or random noise.

\paragraph{Trial accounting.}
The final evaluation contains 23 operator-adjudicated trials per controller for each of three tasks. Success requires all task goals: stacking the brown, green, and blue cups; placing the tennis ball on the green cup and the orange on the brown cup; or placing the carrot on the large plate and the eggplant on the small plate. All 138 trials contribute to success rates and available timing/motion summaries, including failed and interrupted runs. Baseline and \method record 50 and 60 successes out of 69. Fifteen baseline runs and five \method runs contain an execution error; operator verdicts determine the outcome, including one successful \method trial with a terminal controller warning. The main table's time column averages trial start to the operator's success verdict over successful trials only, including synchronization waits. Its denominators are 17/19, 16/22, and 17/19 for baseline/\method on stacking, placement, and sorting, respectively; pooled means use 50/60 successful trials. This operational completion measure includes operator response time and is distinct from autonomously detected goal attainment.

\begin{table}[!htbp]\centering\footnotesize
\caption{\textbf{Measured physical update rates and response latency on the Franka FR3.} Entries report medians of 23 per-episode statistics across each controller and task. Active rates exclude pose/gripper synchronization waits, whereas wall-clock rates retain them; P50 and P95 are first computed within each episode, then summarized separately by their cross-episode medians.}
\label{tab:fr3_final_timing}
\setlength{\tabcolsep}{3pt}\input{tables/fr3_final_timing.tex}
\end{table}

\paragraph{Rate and response definitions.}
A new action is the first dispatch of a distinct policy action node. Repeated pose holds maintain the interface heartbeat but add no new action or plan. A new plan is counted at its first dispatched action. Active time removes the shared pose/gripper wait while retaining post-gripper real-anchor computation; Table~\ref{tab:fr3_final_timing} also reports complete wall-clock rates. The endpoint is command dispatch, not independent confirmation of physical pose attainment or low-level controller consumption. Response starts at the measured state sample and ends at the first dispatch from its resulting plan. All measurements use same-host monotonic timestamps. Across the 69 trials per controller, median feedback-to-plan-ready latency is 64.08\,ms for the baseline and 11.39\,ms for \method; both send repeated targets at approximately 99\,Hz.

\begin{table}[!htbp]\centering\small
\caption{\textbf{Multi-plan dispatches per System~2 window.} A qualifying plan completes and issues its first action between consecutive System~2 starts across 69 \method trials.}
\label{tab:fr3_final_windows}
\setlength{\tabcolsep}{3pt}\input{tables/fr3_final_windows.tex}
\end{table}

\paragraph{Slow-window analysis.}
For each pair of consecutive System~2 start timestamps, we count distinct completed fast plans and their first dispatches. Table~\ref{tab:fr3_final_windows} requires both events inside the same window. Repeated transmissions, unfinished final windows, and windows overlapping pose/gripper pauses are excluded. The resulting 35.00\% fraction measures actual asynchronous plan use rather than a ratio of isolated model timings.

\paragraph{Motion descriptors.}
We use measured EEF positions and segment trajectories at pose/gripper waits or state gaps above 150\,ms. Within each active segment, linear resampling onto a common 100\,ms grid precedes forward differences. Speed CV is the population standard deviation of speed divided by its mean; acceleration RMS is the square root of mean squared acceleration magnitude. No smoothing is applied, and derivatives never cross synchronization gaps. Each episode contributes one descriptor, with medians reported across all 23 trials in its task/controller group, capturing physical motion at 100\,ms resolution.

\paragraph{Selected physical illustration.}
The two selected stacking trials are both successful, with exactly two closing and two opening transitions. They have separately reset object layouts, rather than identical initial states. Figure~\ref{fig:fr3_selected}'s cup-stacking illustration uses original frames from an independent recording camera at corresponding milestones. Markers track the visible rigid gripper housing in image coordinates, not a calibrated TCP projection; trailing lines connect tracked positions from each event-defined phase start to the displayed frame, without filling tracking gaps. The four phases are the initial approach, first transfer, second approach, and second transfer; after the initial approach, each phase starts at the preceding gripper completion. The display retains complete available phase tracks, including high lifts that extended above the previous crop. This image annotation is illustrative. All motion descriptors below are instead computed from the measured EEF log using the common protocol above.

\begin{table}[!htbp]\centering\small
\caption{\textbf{Metrics for the selected cup-stacking pair.} Prompt: \textit{``stack green cup on brown cup then blue cup on green cup.''} Motion descriptors evaluate the complete active trajectory at 100\,ms resolution, while successful-trial duration includes synchronization waits through the operator verdict.}
\label{tab:fr3_selected_pair}
\setlength{\tabcolsep}{3pt}\input{tables/fr3_selected_pair_table.tex}
\end{table}

\begin{figure}[!htbp]\centering
{\fontfamily{phv}\selectfont\small
\textcolor[HTML]{856B59}{\rule[.5ex]{16pt}{1.3pt}}\; $\pi_{0.5}$\qquad
\textcolor[HTML]{1E899B}{\rule[.5ex]{16pt}{1.5pt}}\; \method}
\par\smallskip
\begin{subfigure}[t]{.485\textwidth}\centering
\includegraphics[width=\linewidth]{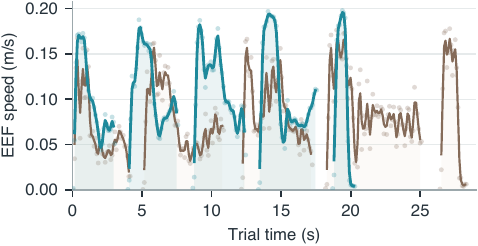}
\caption{Measured EEF speed.}\label{fig:fr3_speed}
\end{subfigure}\hfill
\begin{subfigure}[t]{.485\textwidth}\centering
\includegraphics[width=\linewidth]{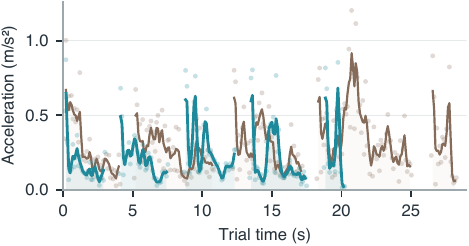}
\caption{Measured EEF acceleration.}\label{fig:fr3_accel}
\end{subfigure}

\caption{\textbf{Measured end-effector speed and acceleration for the selected cup-stacking pair.} Translucent markers show unsmoothed 100\,ms samples and solid curves apply identical three-point triangular display smoothing within active segments, preserving synchronization gaps. Unsmoothed acceleration RMS and speed CV drop by 21.3\% and 5.6\%, respectively.}
\label{fig:fr3_selected_motion}
\end{figure}

\begin{figure}[!htbp]\centering
{\fontfamily{phv}\selectfont\small
\textcolor[HTML]{856B59}{\rule[.5ex]{16pt}{1.3pt}}\; $\pi_{0.5}$\qquad
\textcolor[HTML]{1E899B}{\rule[.5ex]{16pt}{1.5pt}}\; \method}
\par\smallskip
\begin{subfigure}[t]{.325\textwidth}\centering
\includegraphics[width=\linewidth]{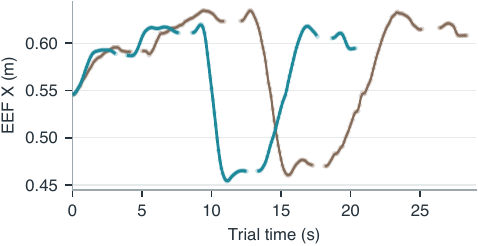}
\caption{Measured X coordinate.}\label{fig:fr3_eef_x}
\end{subfigure}\hfill
\begin{subfigure}[t]{.325\textwidth}\centering
\includegraphics[width=\linewidth]{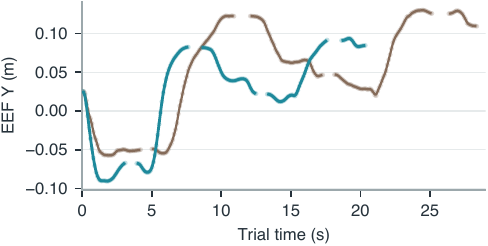}
\caption{Measured Y coordinate.}\label{fig:fr3_eef_y}
\end{subfigure}\hfill
\begin{subfigure}[t]{.325\textwidth}\centering
\includegraphics[width=\linewidth]{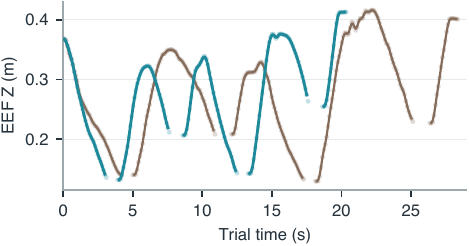}
\caption{Measured Z coordinate.}\label{fig:fr3_eef_z}
\end{subfigure}

\caption{\textbf{Measured end-effector Cartesian coordinates for the selected cup-stacking pair.} Translucent markers preserve unsmoothed 100\,ms samples while curves apply identical three-point display weights $(0.25,0.5,0.25)$ within active segments, leaving synchronization and missing-state gaps unbroken.}
\label{fig:fr3_eef_coordinates}
\end{figure}

For this illustration, speed CV decreases by 5.6\% and acceleration RMS by 21.3\%. Feedback-to-first-dispatch P50/P95 decrease from 72.16/73.84\,ms to 13.50/23.26\,ms, while active replanning increases from 10.20 to 21.59\,Hz. These descriptors agree with the full-cohort motion/response trends; the two selected trials do not supply a separate population estimate.


\FloatBarrier
\subsubsection{Motion- and Layout-Matched Physical Sequences}
\label{sec:matched_physical}
To inspect complete trajectories under matched kinematics and initial scene layouts, we first align motion phase paths across the 98 physical trials containing four complete close--open--close--open transitions and sufficient end-effector samples. Within each task and layout reference, each 3D phase trajectory is resampled at 40 equidistant arc-length points to rank pairwise RMS geometric distance, retaining the closest 10\% of candidate pairs for visual layout screening without conditioning selection on success, latency, or smoothness. Five disjoint pairs spanning cup stacking, fruit placement, and vegetable sorting meet this geometry-and-layout criterion, excluding one vegetable candidate with mismatched carrot pose.

\begin{figure}[H]\centering
\captionsetup[subfigure]{font={scriptsize},skip=2pt}
\makebox[\textwidth][l]{\footnotesize\textbf{$\pi_{0.5}$}}\par\vspace{2pt}
\begin{subfigure}{.16\textwidth}\centering
\includegraphics[width=\linewidth]{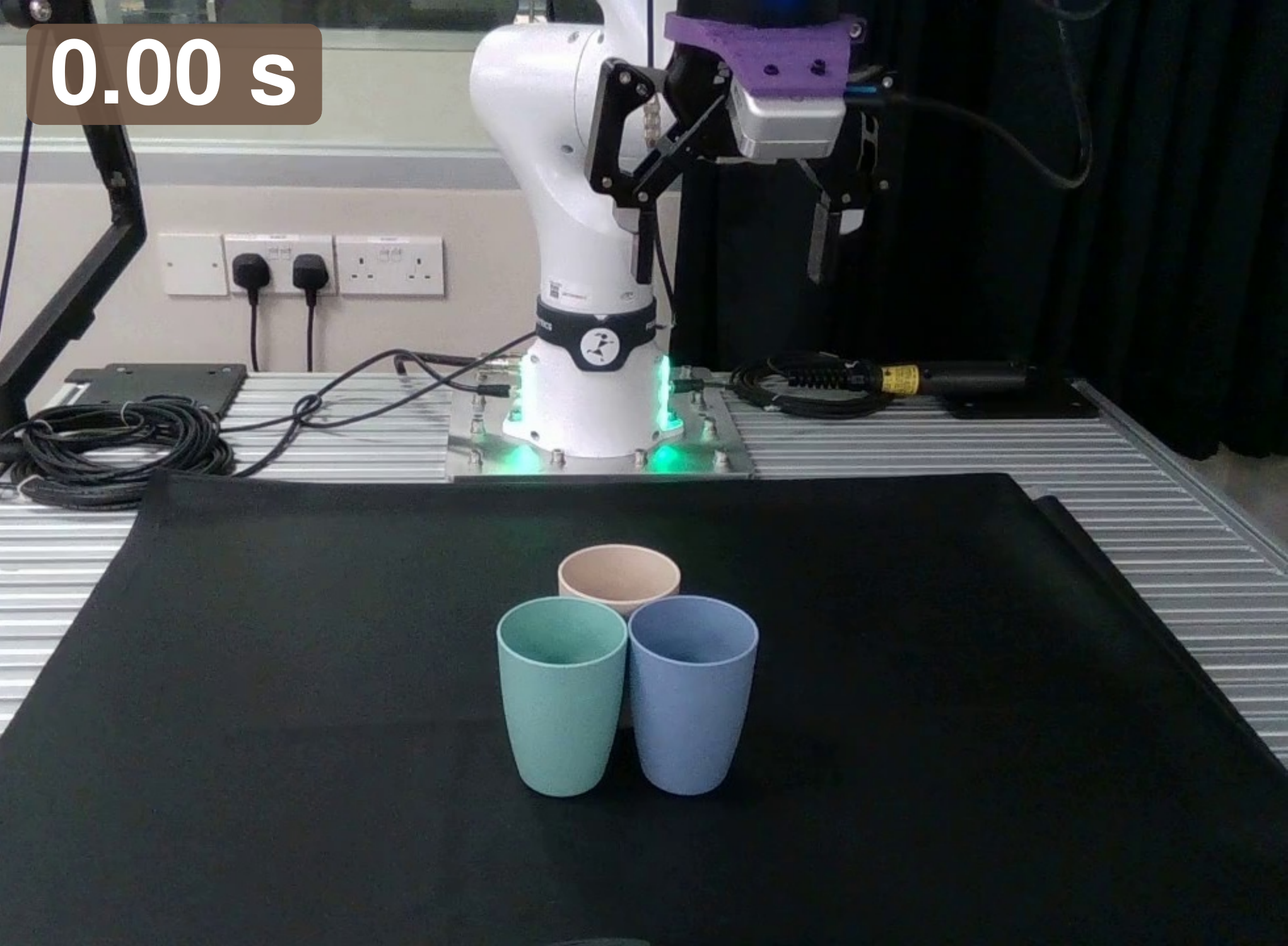}
\caption{Initial.}\end{subfigure}\hfill
\begin{subfigure}{.16\textwidth}\centering
\includegraphics[width=\linewidth]{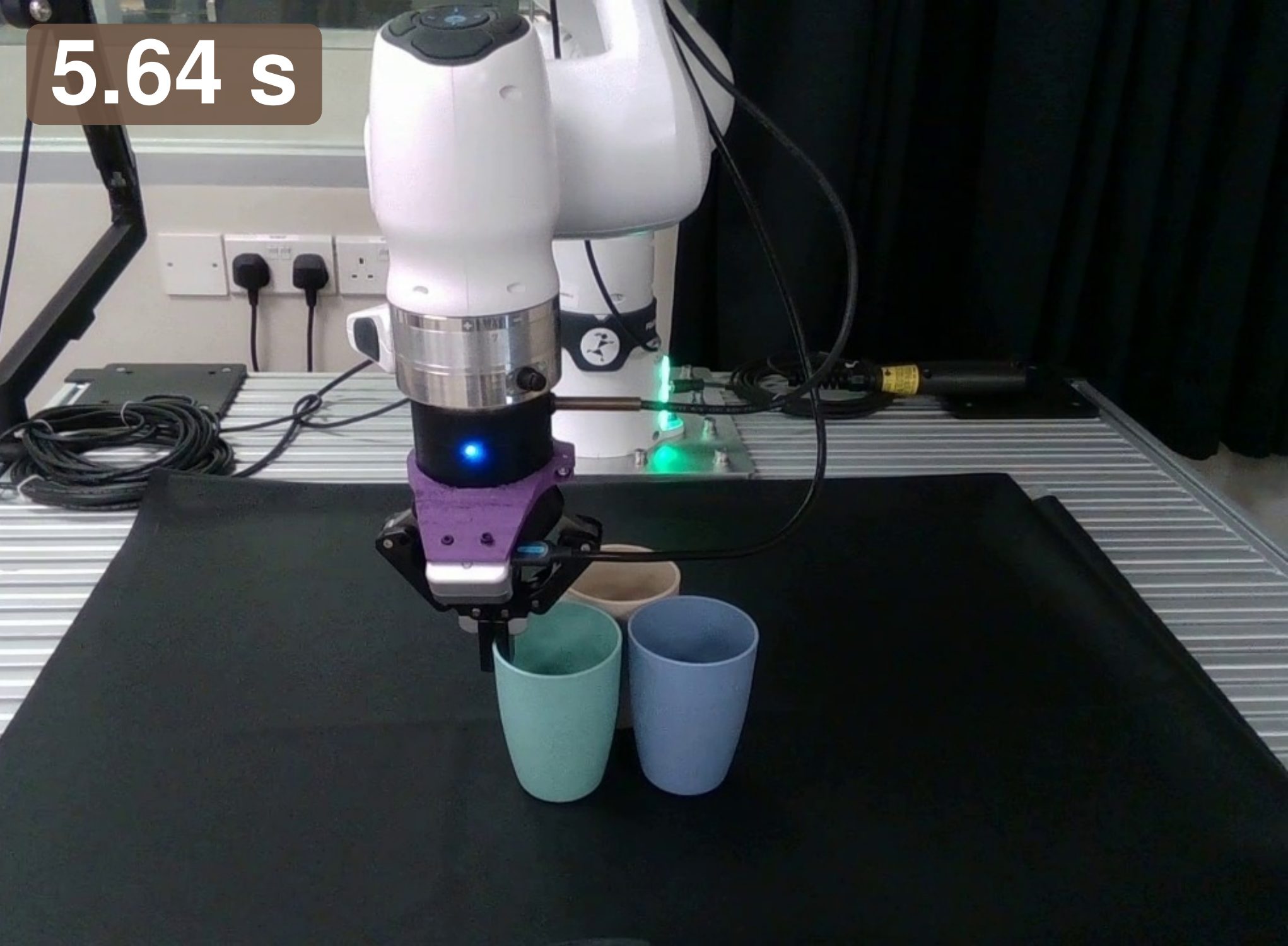}
\caption{Grasp 1.}\end{subfigure}\hfill
\begin{subfigure}{.16\textwidth}\centering
\includegraphics[width=\linewidth]{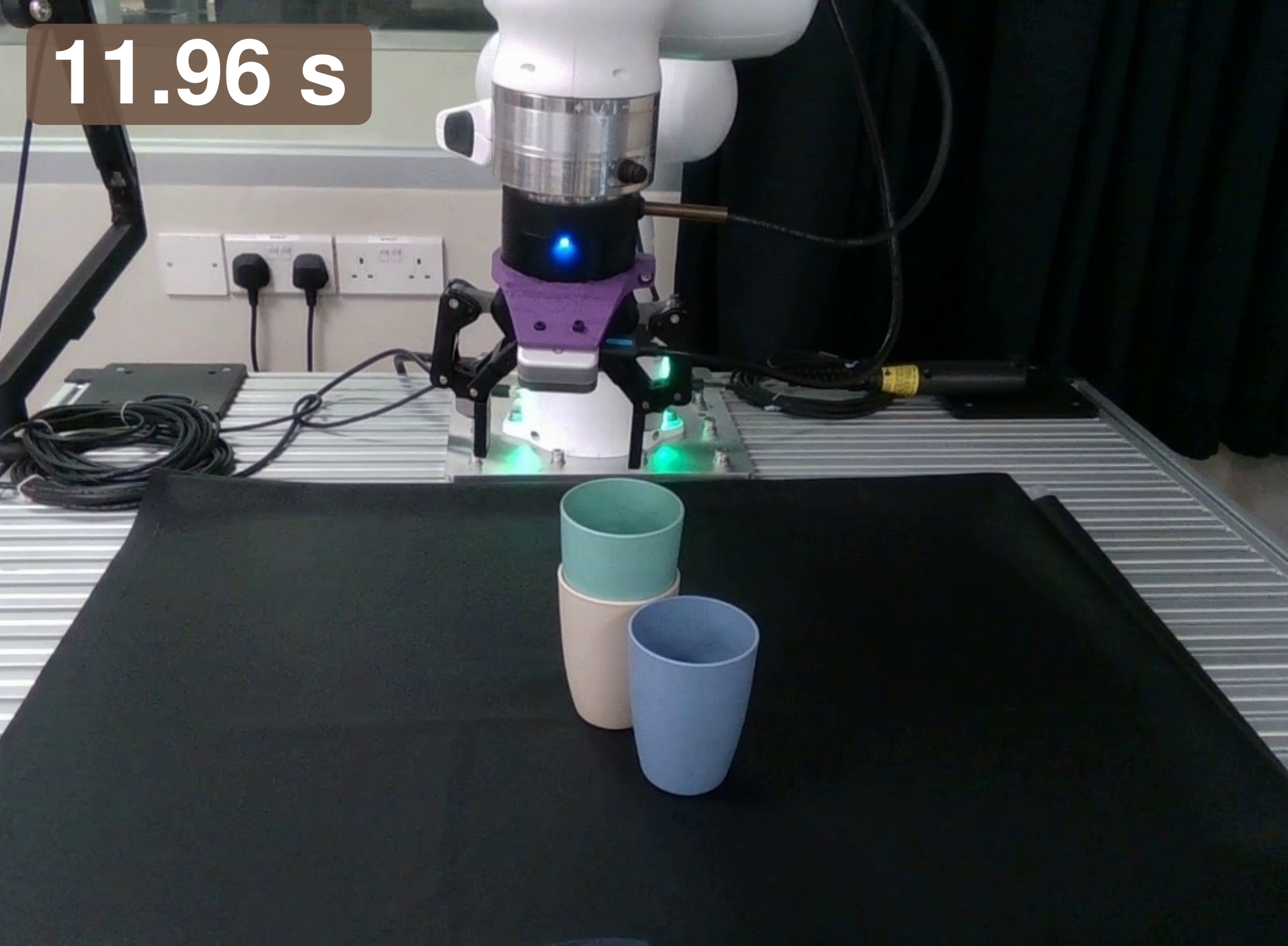}
\caption{Release 1.}\end{subfigure}\hfill
\begin{subfigure}{.16\textwidth}\centering
\includegraphics[width=\linewidth]{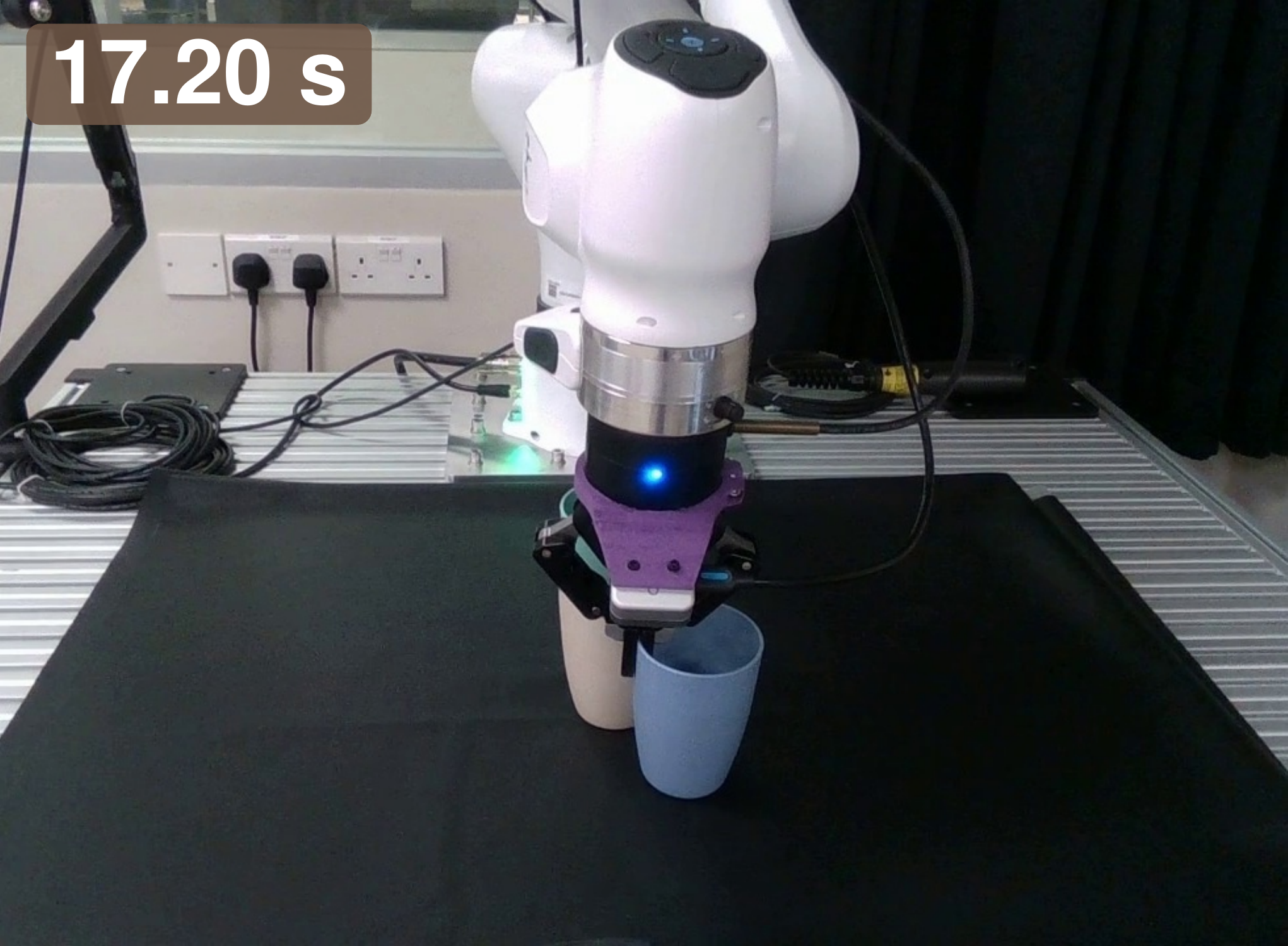}
\caption{Grasp 2.}\end{subfigure}\hfill
\begin{subfigure}{.16\textwidth}\centering
\includegraphics[width=\linewidth]{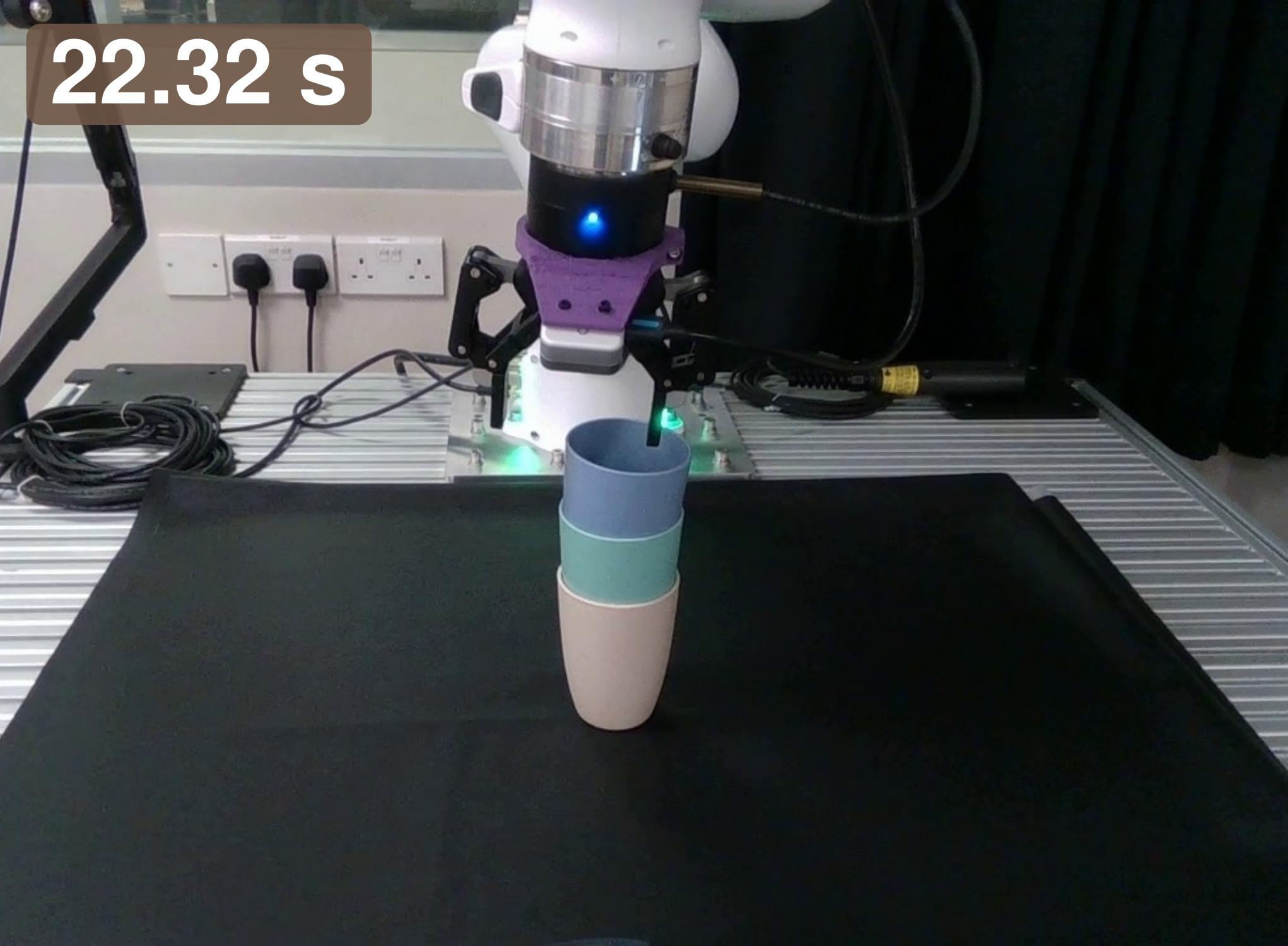}
\caption{Release 2.}\end{subfigure}\hfill
\begin{subfigure}{.16\textwidth}\centering
\includegraphics[width=\linewidth]{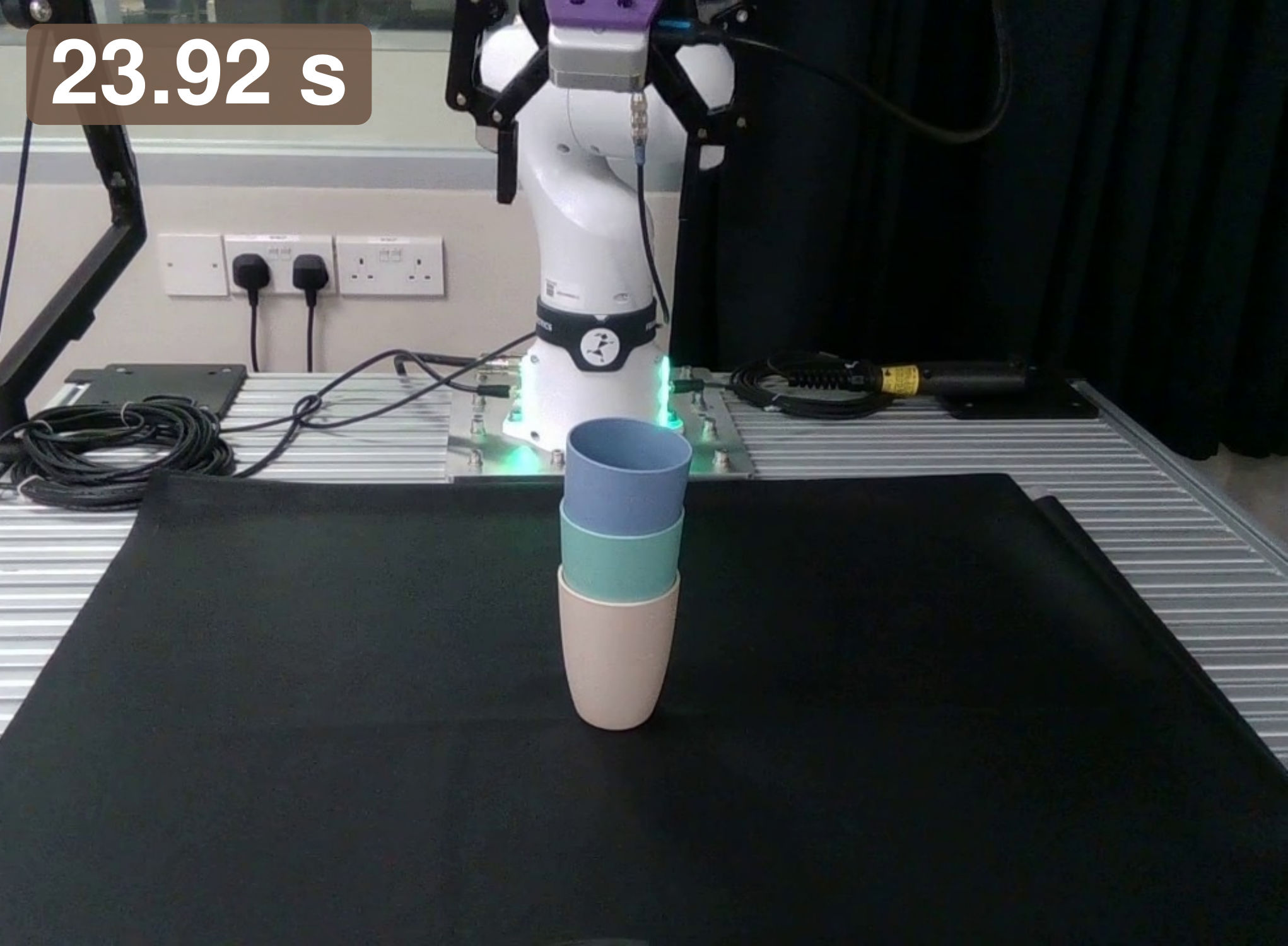}
\caption{Final.}\end{subfigure}\par\vspace{3pt}
\makebox[\textwidth][l]{\footnotesize\textbf{\method}}\par\vspace{2pt}
\begin{subfigure}{.16\textwidth}\centering
\includegraphics[width=\linewidth]{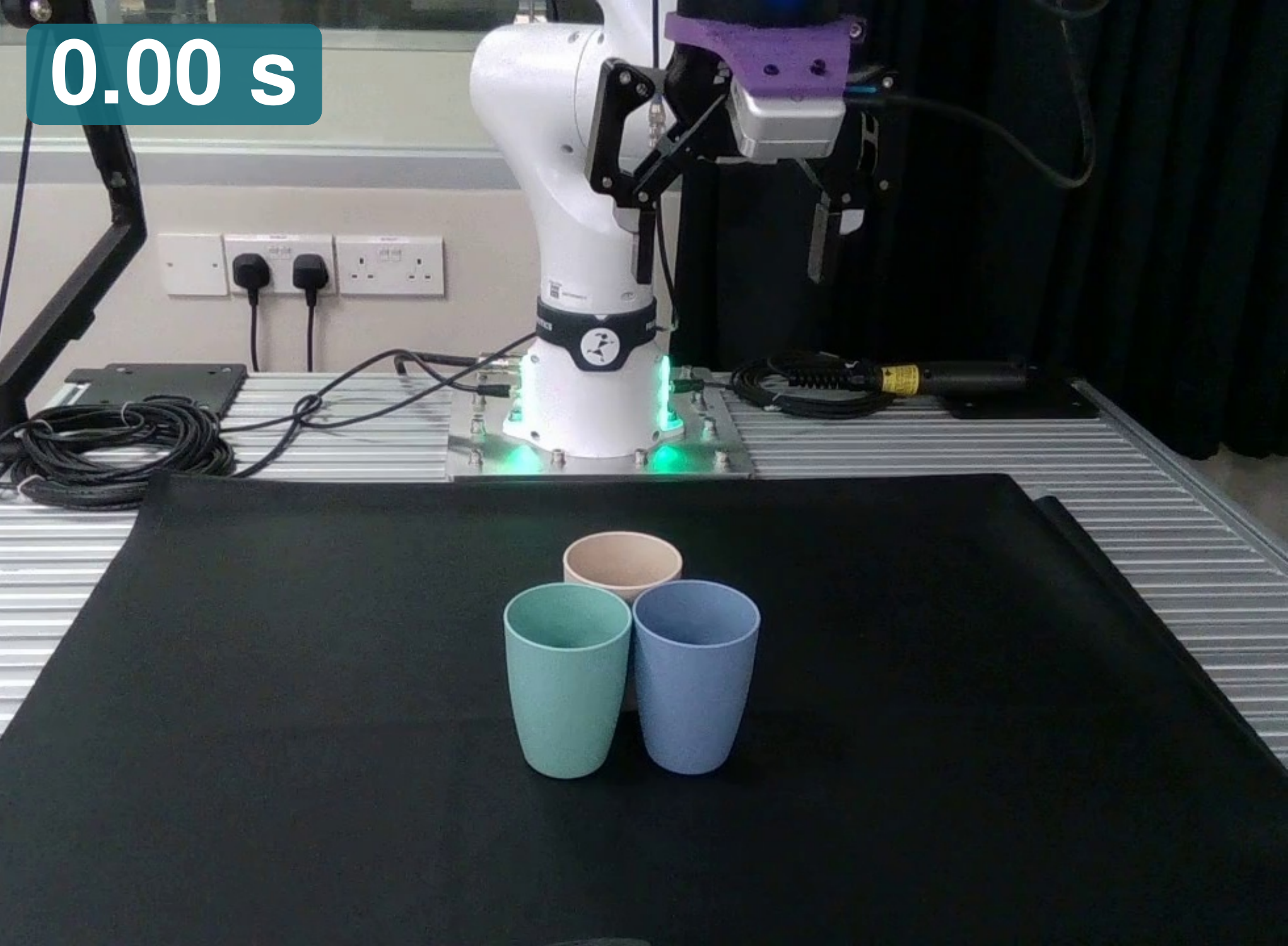}
\caption{Initial.}\end{subfigure}\hfill
\begin{subfigure}{.16\textwidth}\centering
\includegraphics[width=\linewidth]{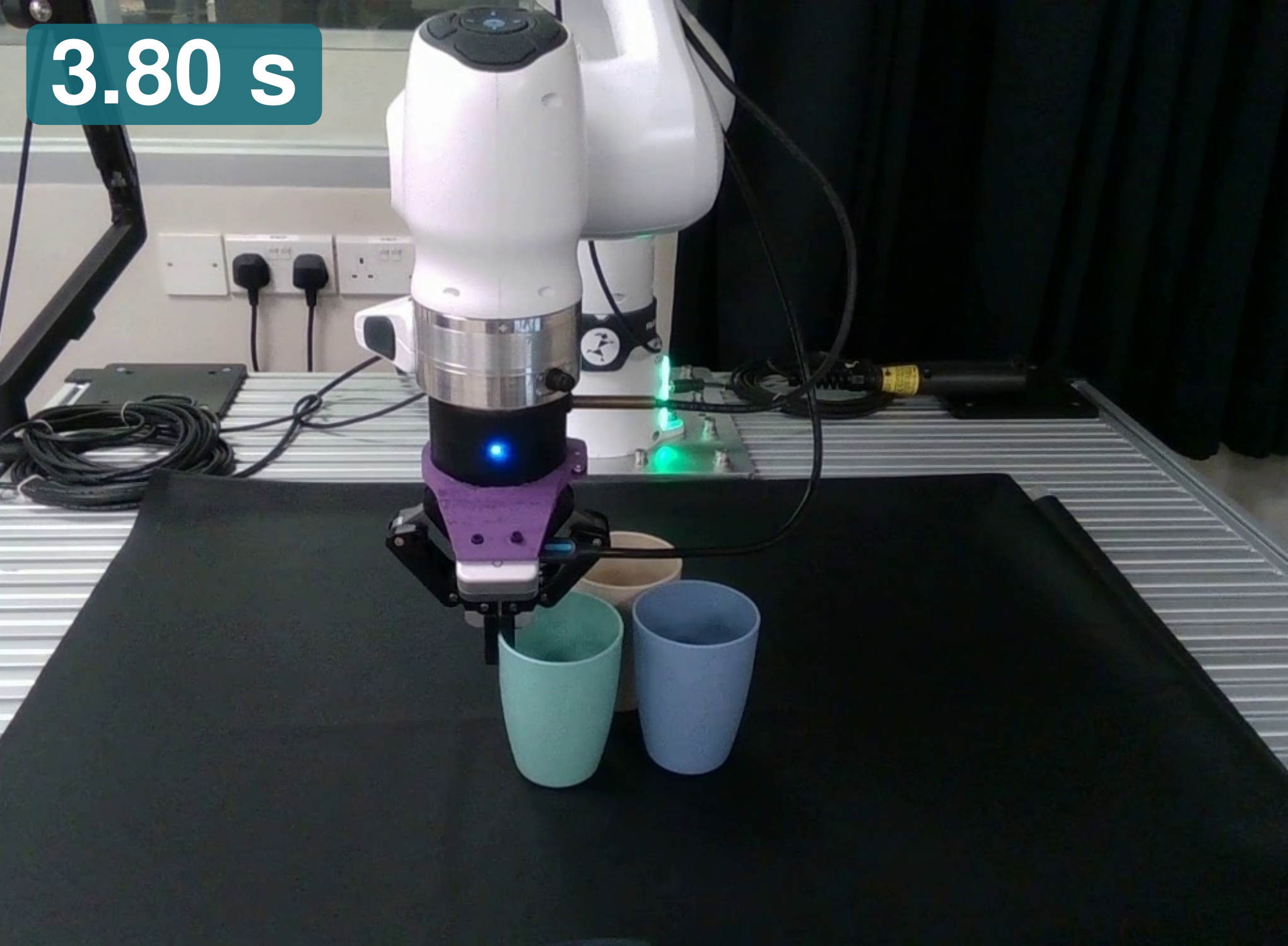}
\caption{Grasp 1.}\end{subfigure}\hfill
\begin{subfigure}{.16\textwidth}\centering
\includegraphics[width=\linewidth]{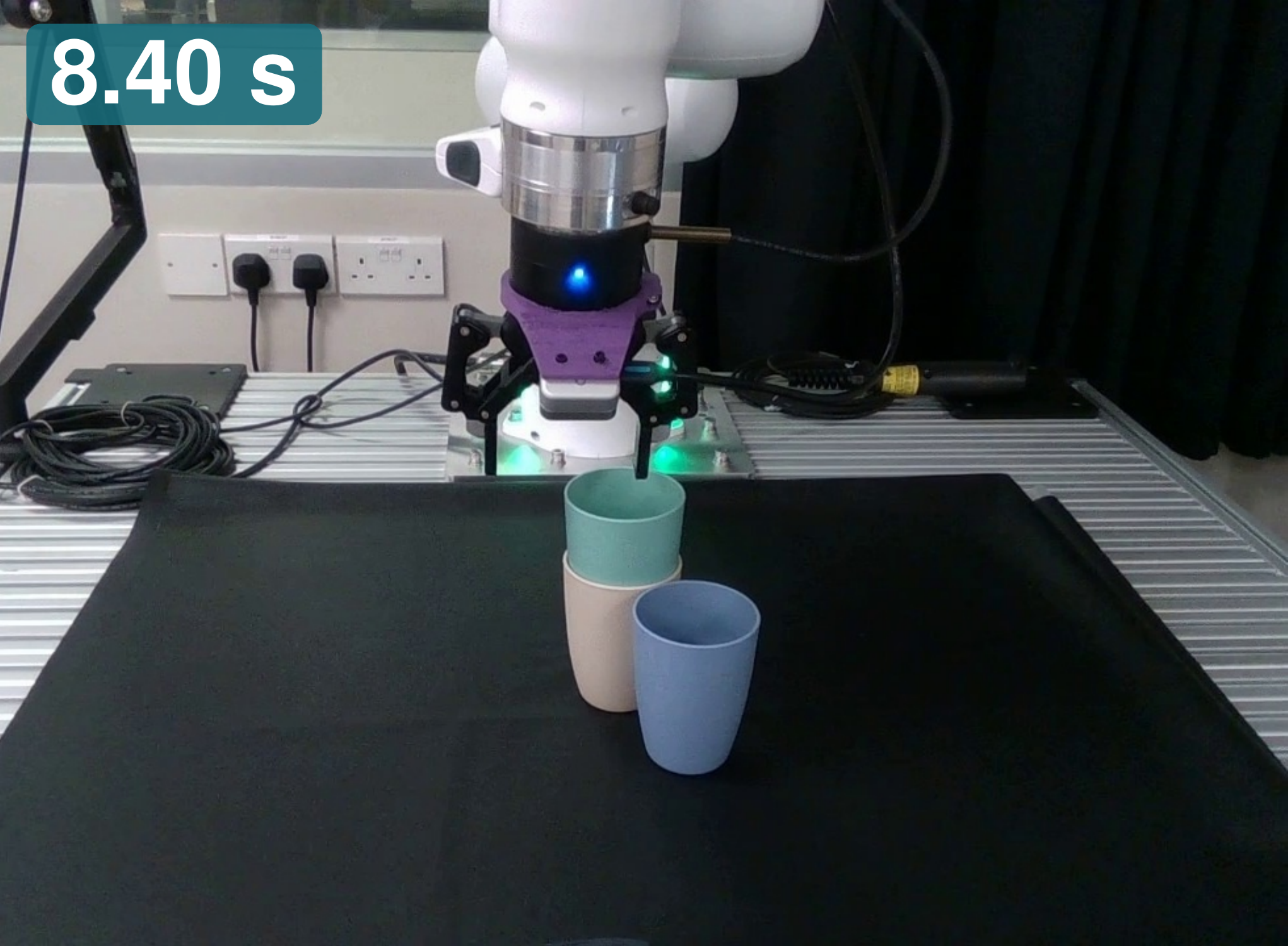}
\caption{Release 1.}\end{subfigure}\hfill
\begin{subfigure}{.16\textwidth}\centering
\includegraphics[width=\linewidth]{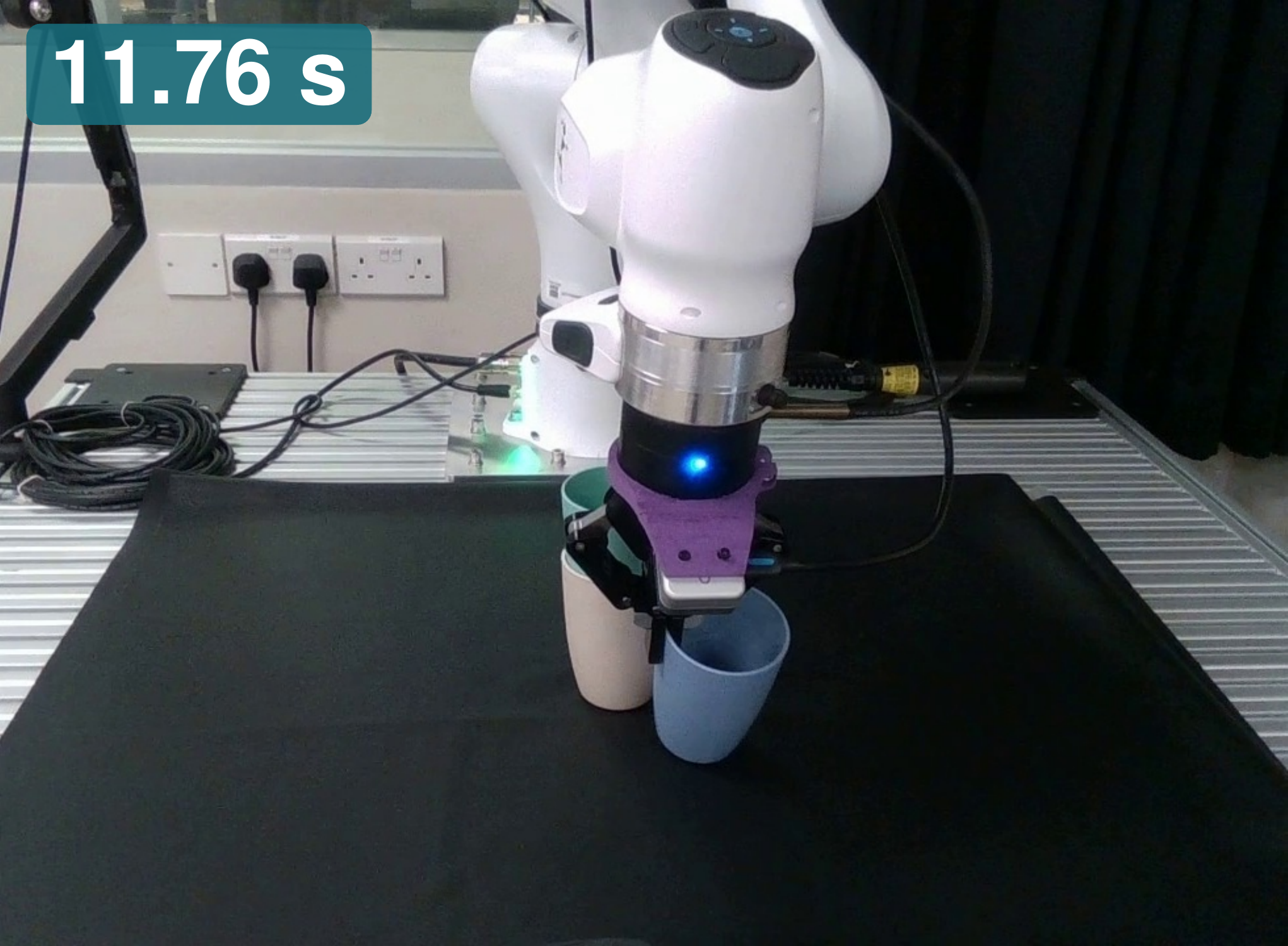}
\caption{Grasp 2.}\end{subfigure}\hfill
\begin{subfigure}{.16\textwidth}\centering
\includegraphics[width=\linewidth]{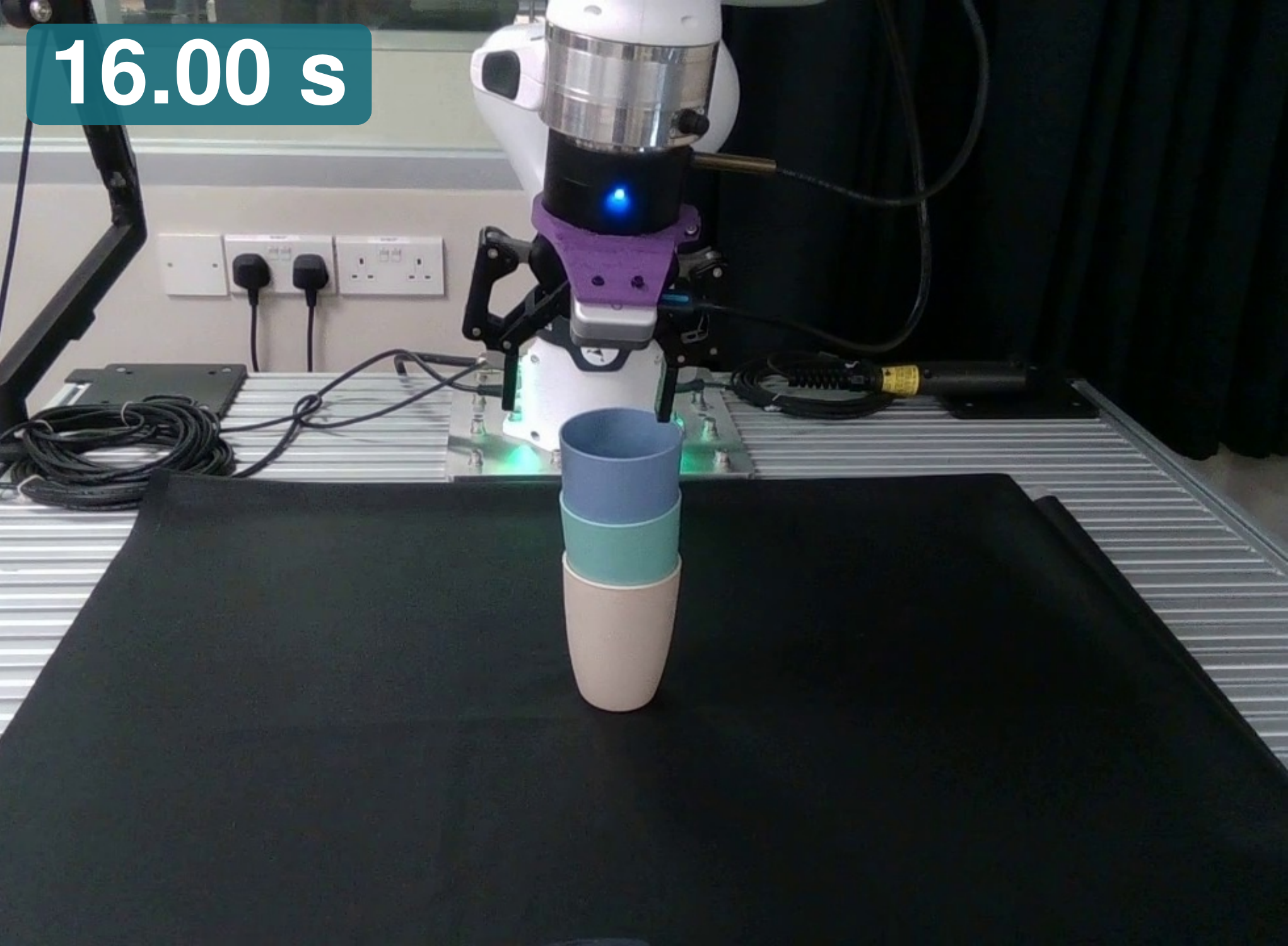}
\caption{Release 2.}\end{subfigure}\hfill
\begin{subfigure}{.16\textwidth}\centering
\includegraphics[width=\linewidth]{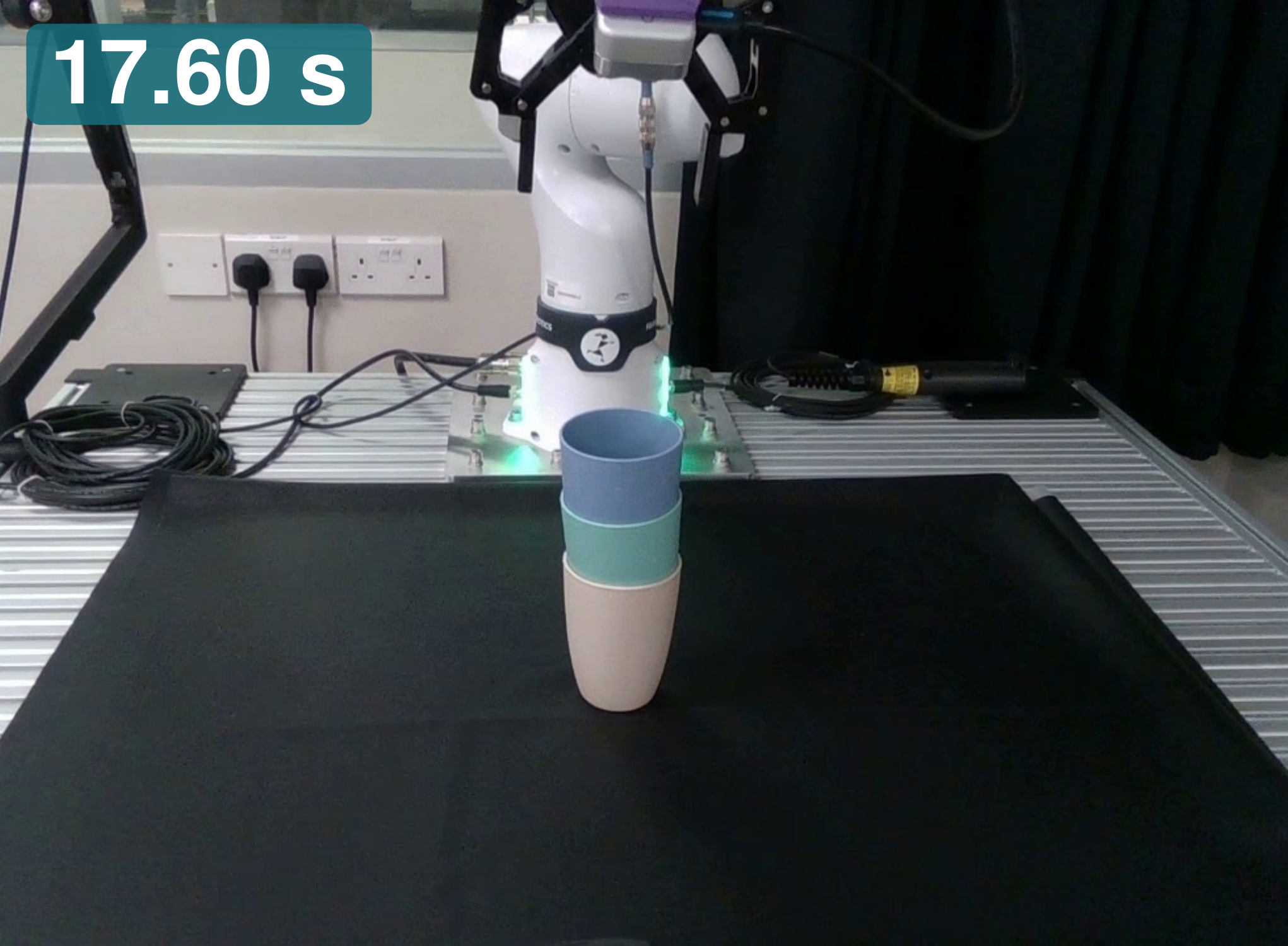}
\caption{Final.}\end{subfigure}\par\smallskip
\caption{\textbf{Cup stacking: matched sequence 1.} Prompt: \textit{``stack green cup on brown cup then blue cup on green cup.''} Rows compare $\pi_{0.5}$ above and \method below across six event-defined stages under shared framing, with top-left badges indicating elapsed trial time.}
\label{fig:motion_pair_1}
\end{figure}

\begin{figure}[H]\centering
\captionsetup[subfigure]{font={scriptsize},skip=2pt}
\makebox[\textwidth][l]{\footnotesize\textbf{$\pi_{0.5}$}}\par\vspace{2pt}
\begin{subfigure}{.16\textwidth}\centering
\includegraphics[width=\linewidth]{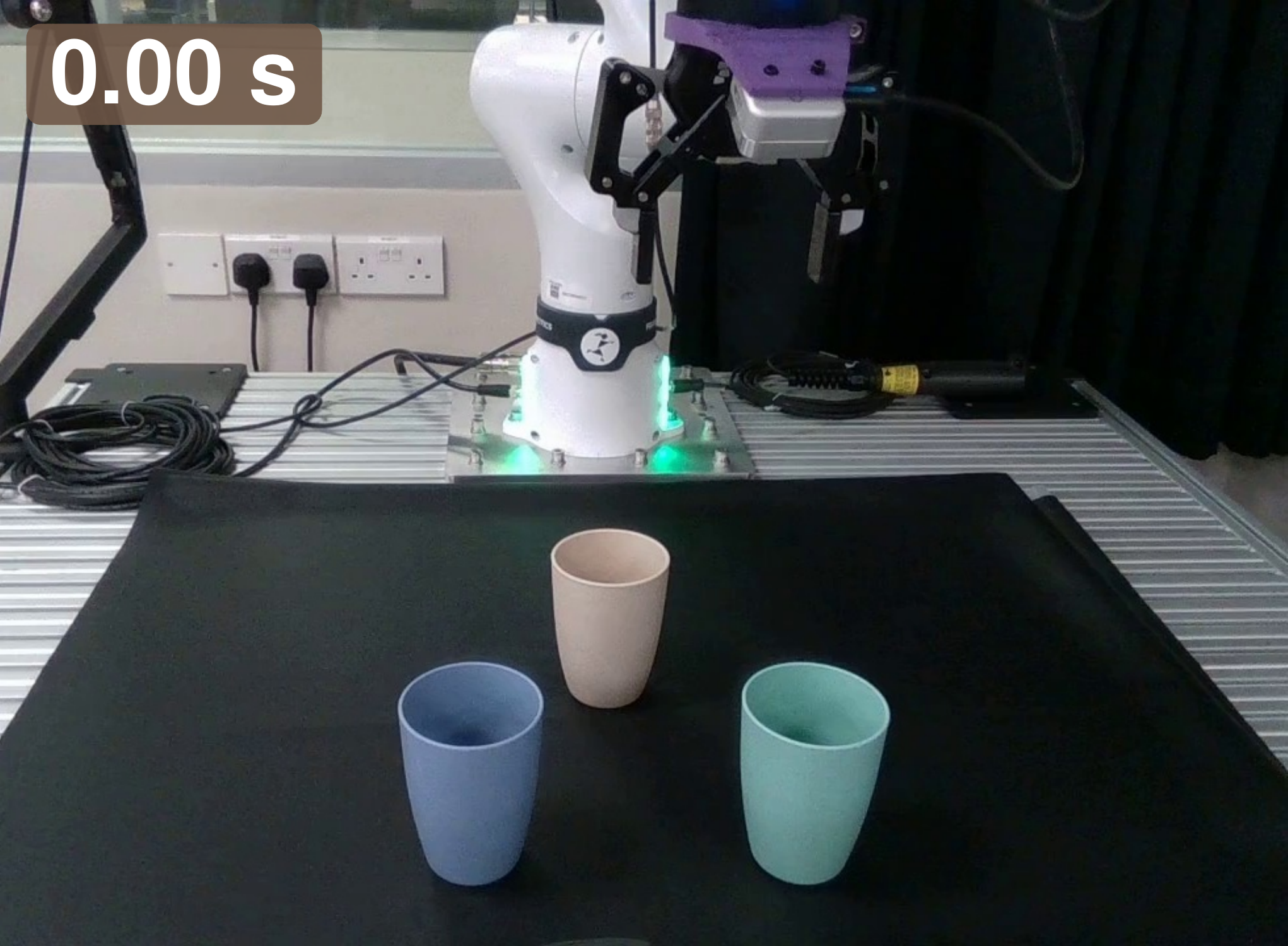}
\caption{Initial.}\end{subfigure}\hfill
\begin{subfigure}{.16\textwidth}\centering
\includegraphics[width=\linewidth]{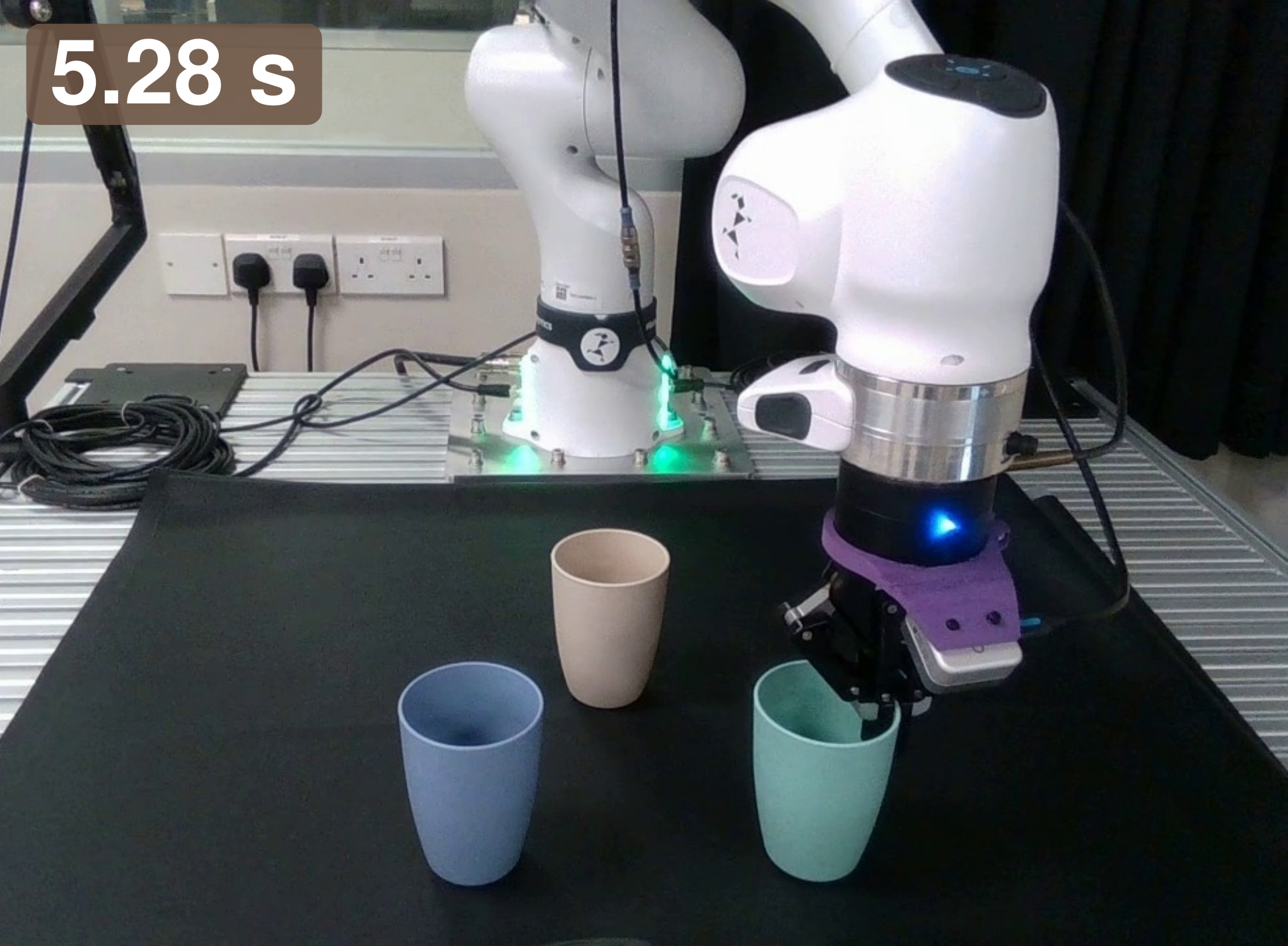}
\caption{Grasp 1.}\end{subfigure}\hfill
\begin{subfigure}{.16\textwidth}\centering
\includegraphics[width=\linewidth]{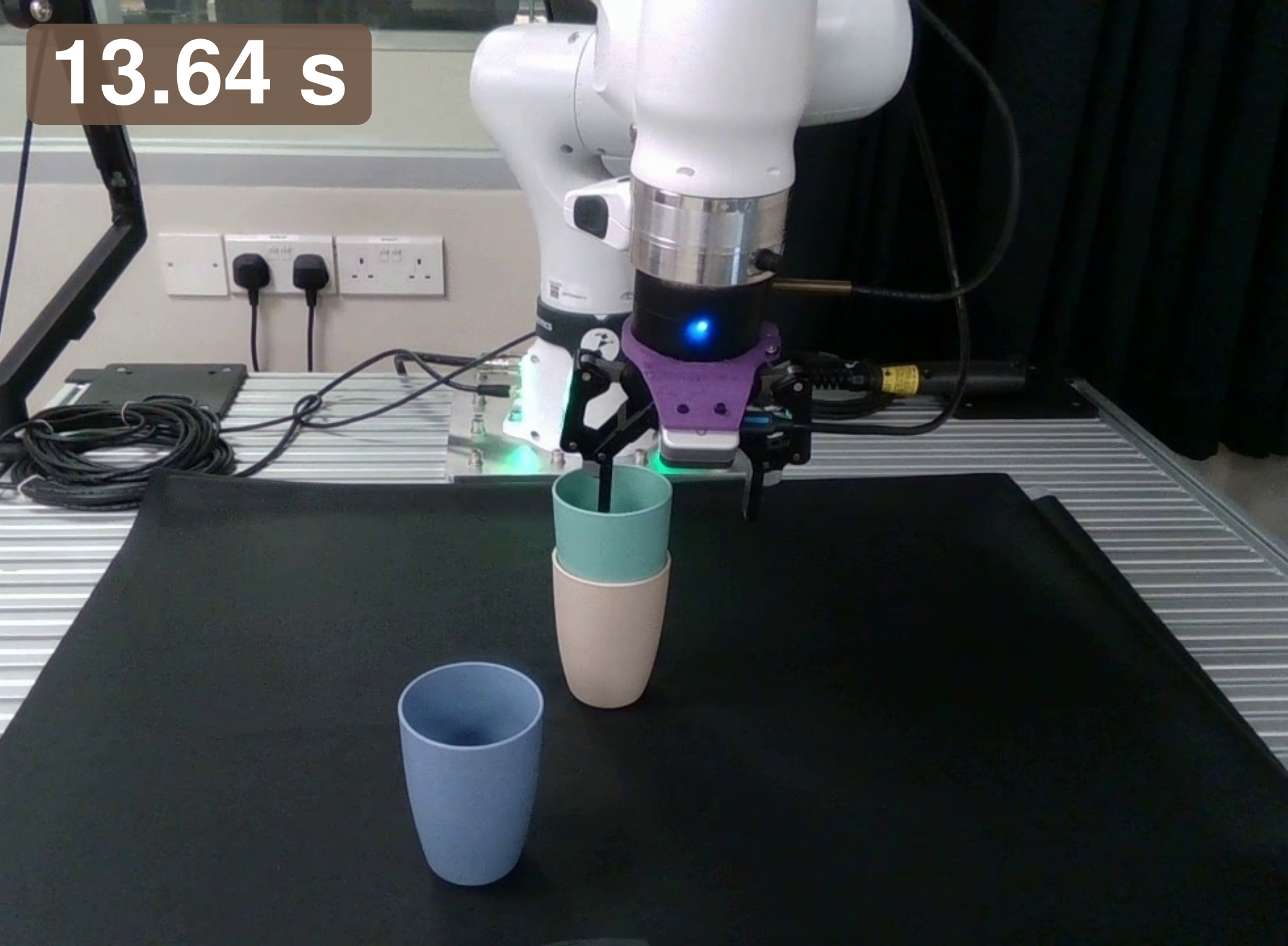}
\caption{Release 1.}\end{subfigure}\hfill
\begin{subfigure}{.16\textwidth}\centering
\includegraphics[width=\linewidth]{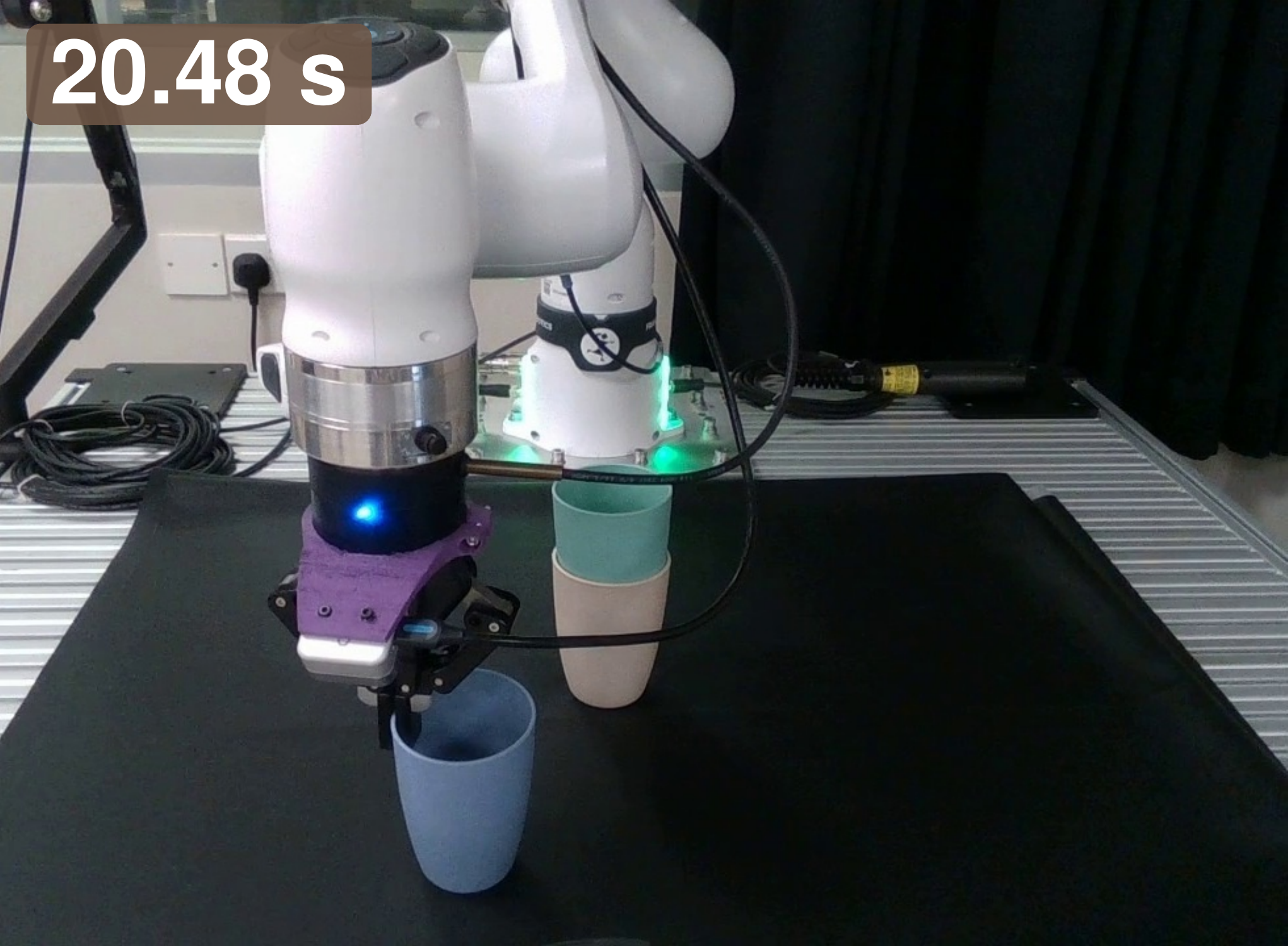}
\caption{Grasp 2.}\end{subfigure}\hfill
\begin{subfigure}{.16\textwidth}\centering
\includegraphics[width=\linewidth]{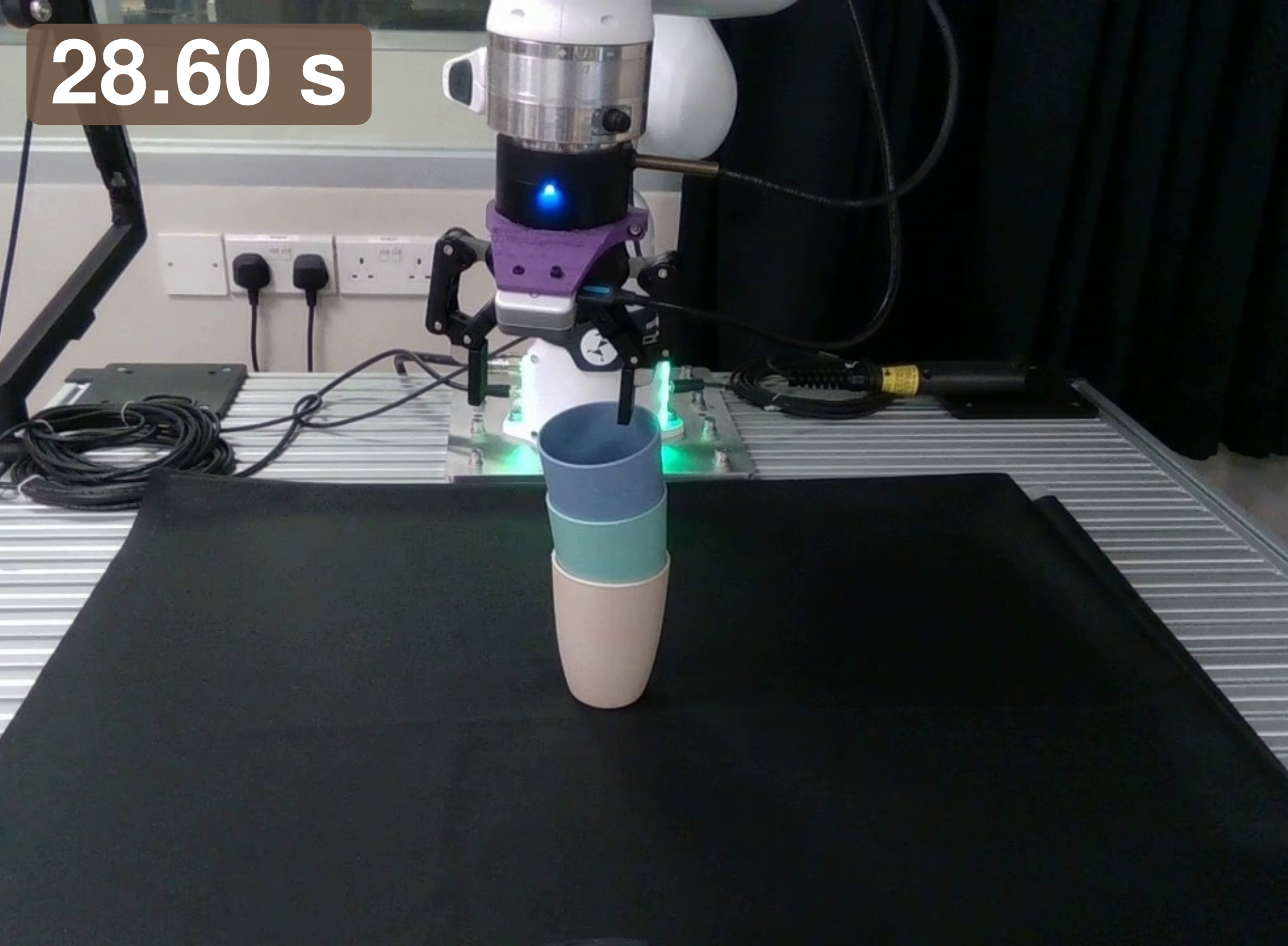}
\caption{Release 2.}\end{subfigure}\hfill
\begin{subfigure}{.16\textwidth}\centering
\includegraphics[width=\linewidth]{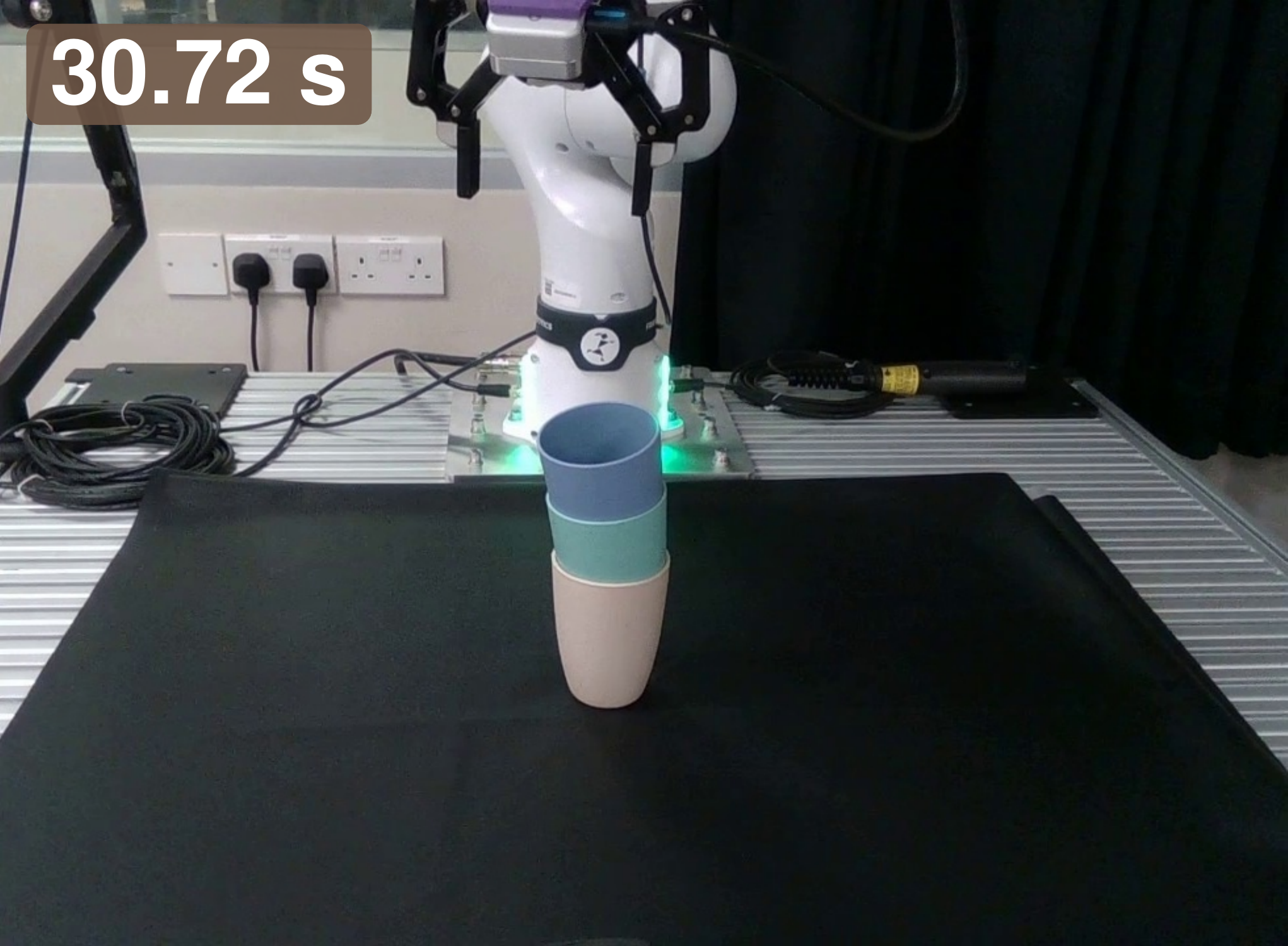}
\caption{Final.}\end{subfigure}\par\vspace{3pt}
\makebox[\textwidth][l]{\footnotesize\textbf{\method}}\par\vspace{2pt}
\begin{subfigure}{.16\textwidth}\centering
\includegraphics[width=\linewidth]{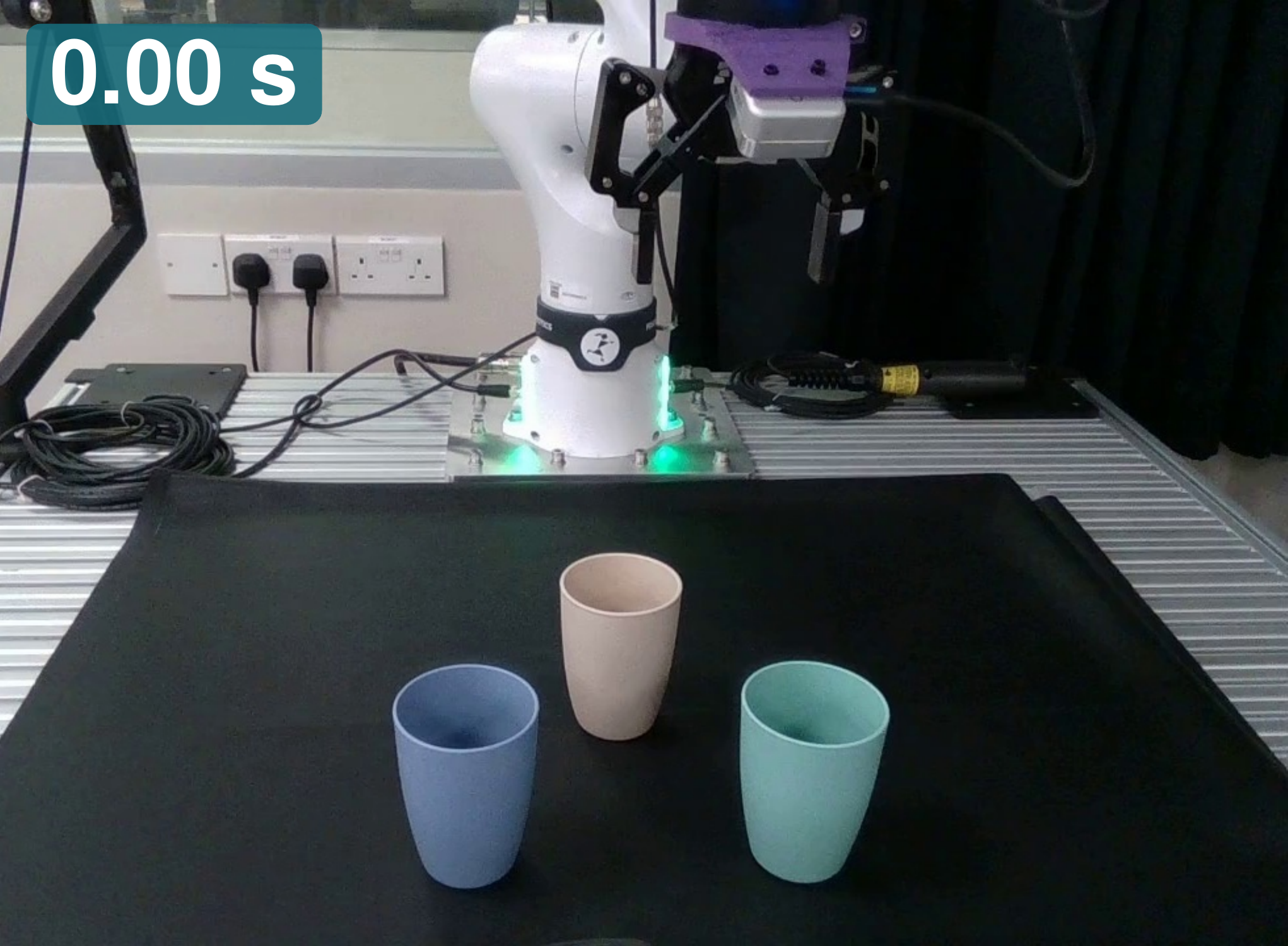}
\caption{Initial.}\end{subfigure}\hfill
\begin{subfigure}{.16\textwidth}\centering
\includegraphics[width=\linewidth]{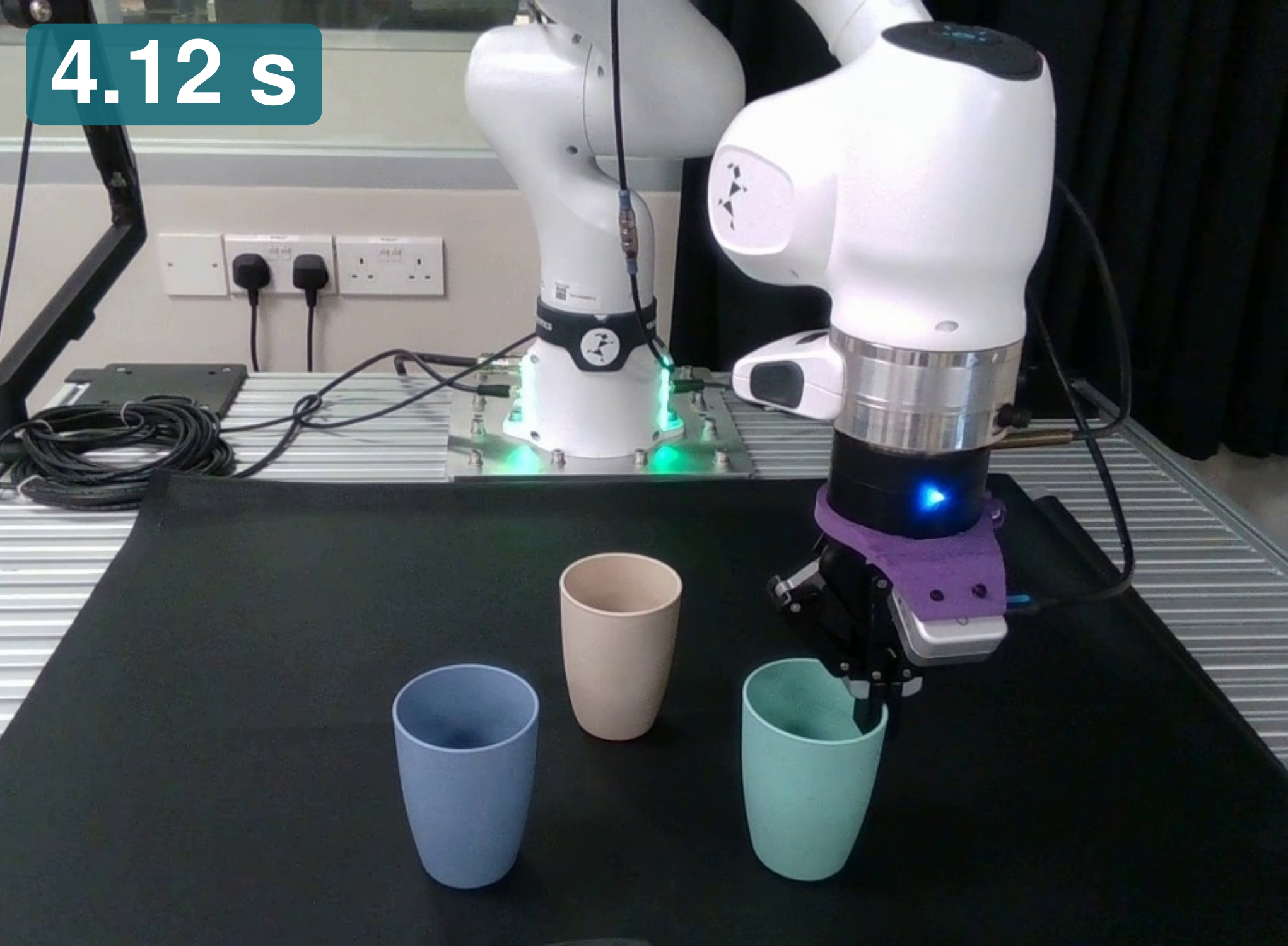}
\caption{Grasp 1.}\end{subfigure}\hfill
\begin{subfigure}{.16\textwidth}\centering
\includegraphics[width=\linewidth]{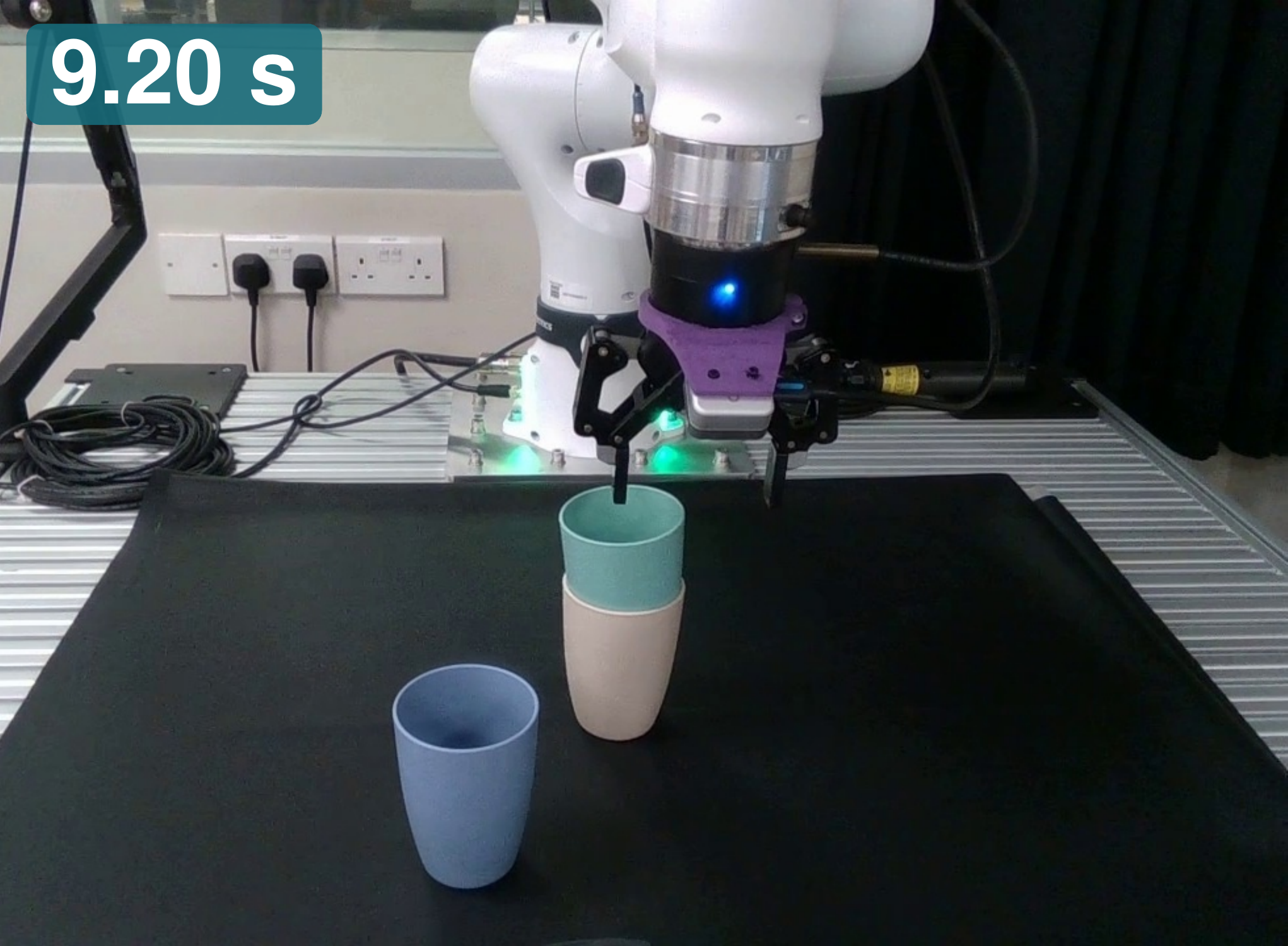}
\caption{Release 1.}\end{subfigure}\hfill
\begin{subfigure}{.16\textwidth}\centering
\includegraphics[width=\linewidth]{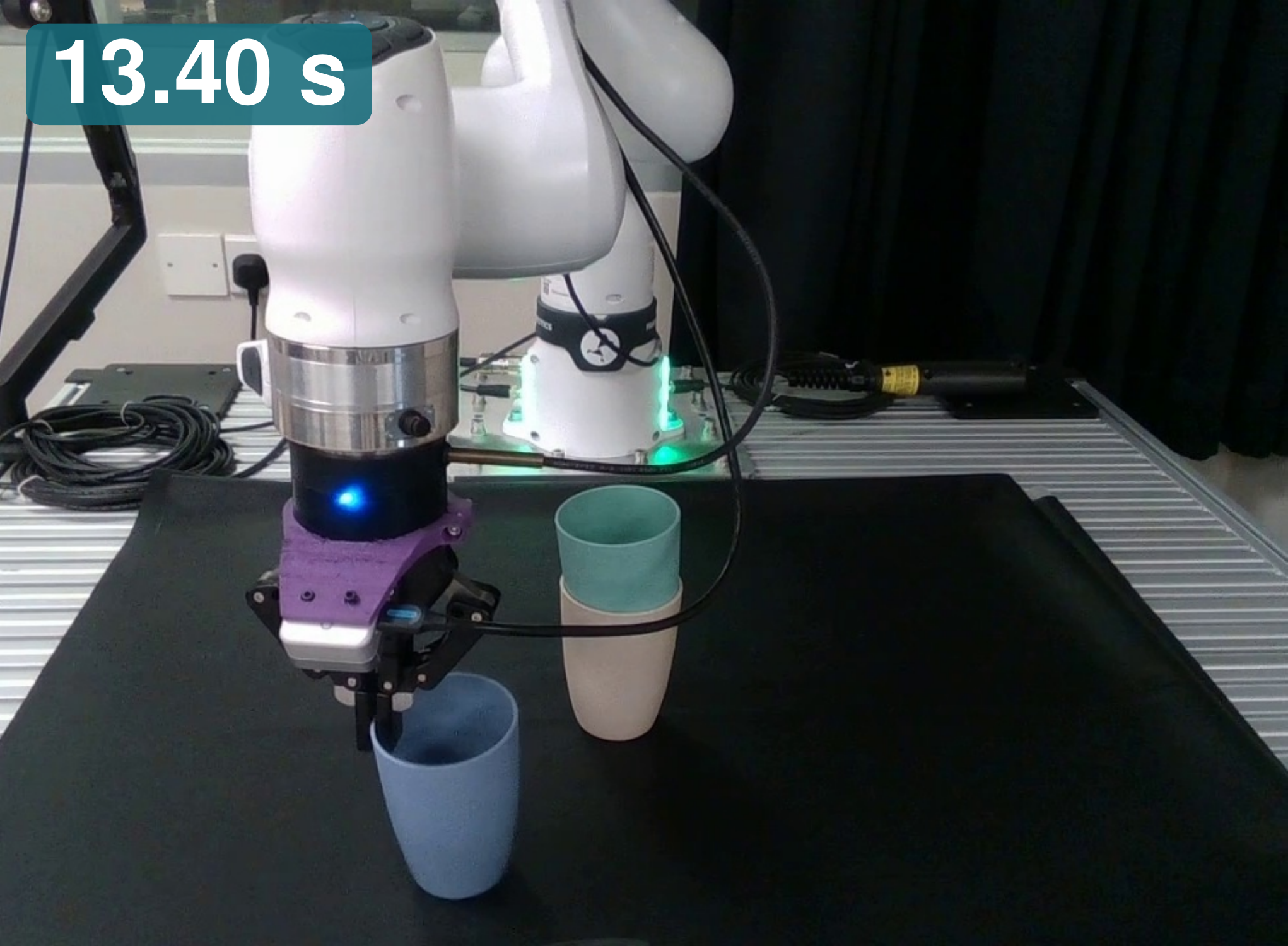}
\caption{Grasp 2.}\end{subfigure}\hfill
\begin{subfigure}{.16\textwidth}\centering
\includegraphics[width=\linewidth]{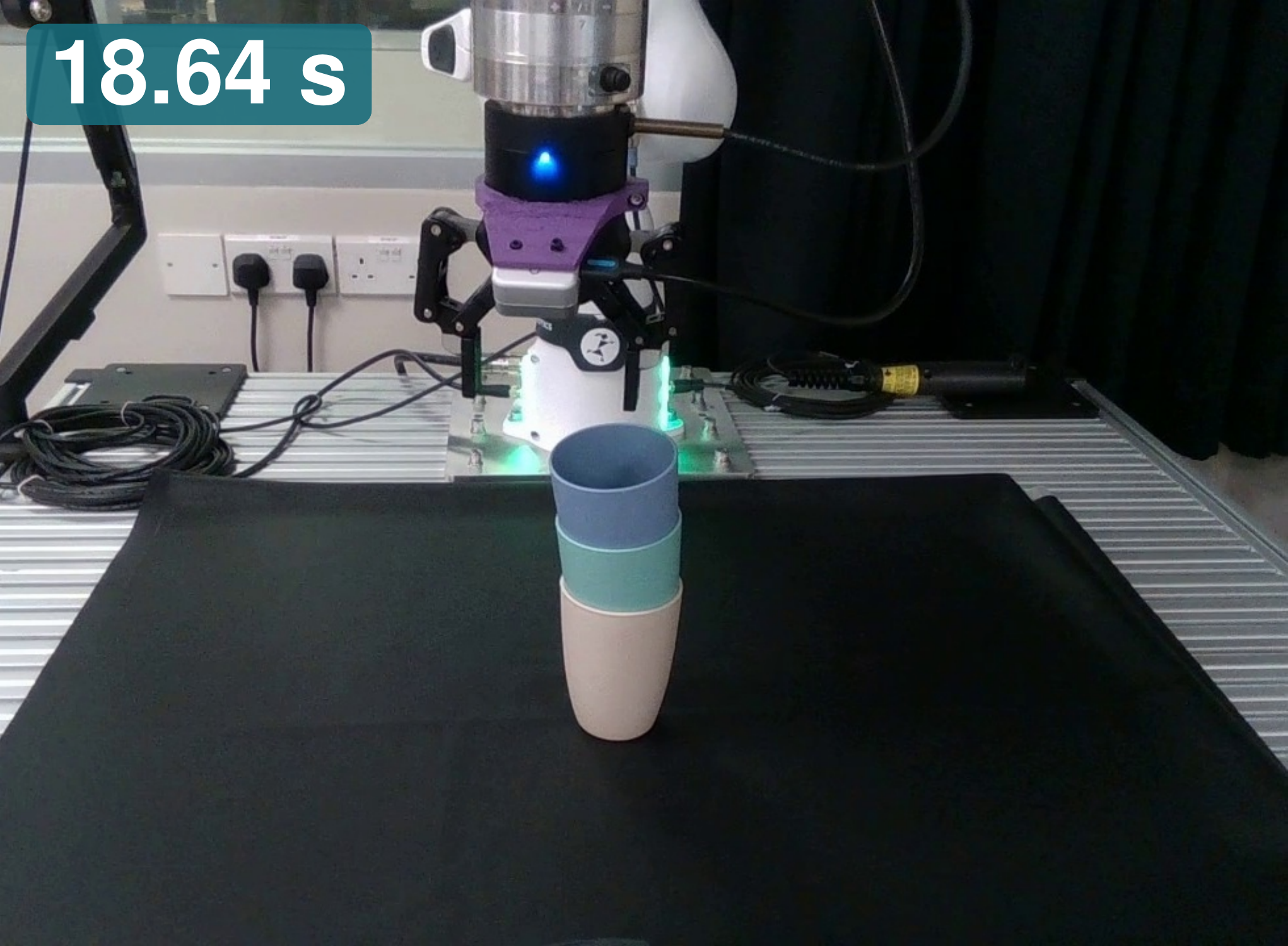}
\caption{Release 2.}\end{subfigure}\hfill
\begin{subfigure}{.16\textwidth}\centering
\includegraphics[width=\linewidth]{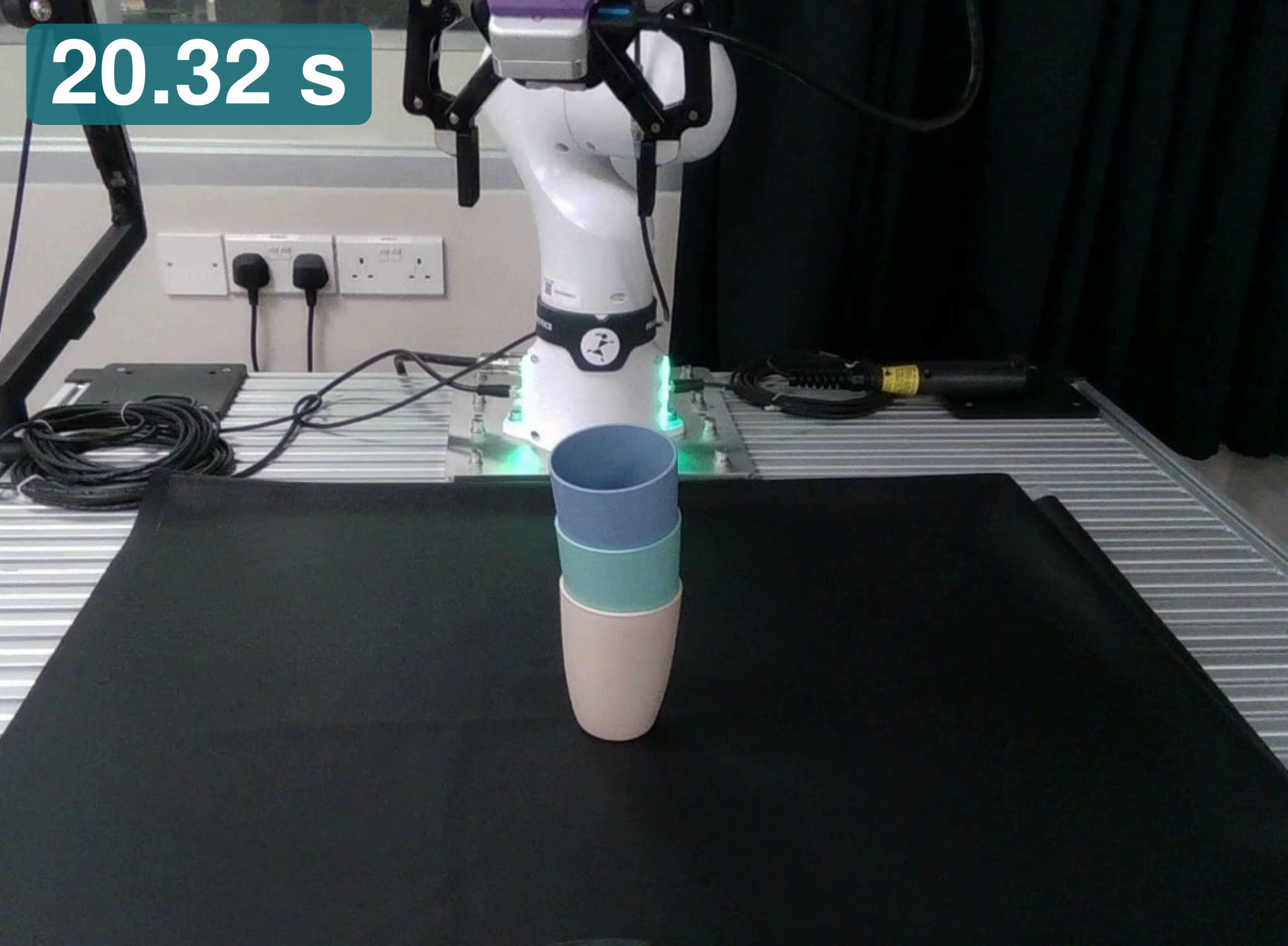}
\caption{Final.}\end{subfigure}\par\smallskip
\caption{\textbf{Cup stacking: matched sequence 2.} Prompt: \textit{``stack green cup on brown cup then blue cup on green cup.''} Rows compare $\pi_{0.5}$ above and \method below across six event-defined stages under shared framing, with top-left badges indicating elapsed trial time.}
\label{fig:motion_pair_2}
\end{figure}

\begin{figure}[H]\centering
\captionsetup[subfigure]{font={scriptsize},skip=2pt}
\makebox[\textwidth][l]{\footnotesize\textbf{$\pi_{0.5}$}}\par\vspace{2pt}
\begin{subfigure}{.16\textwidth}\centering
\includegraphics[width=\linewidth]{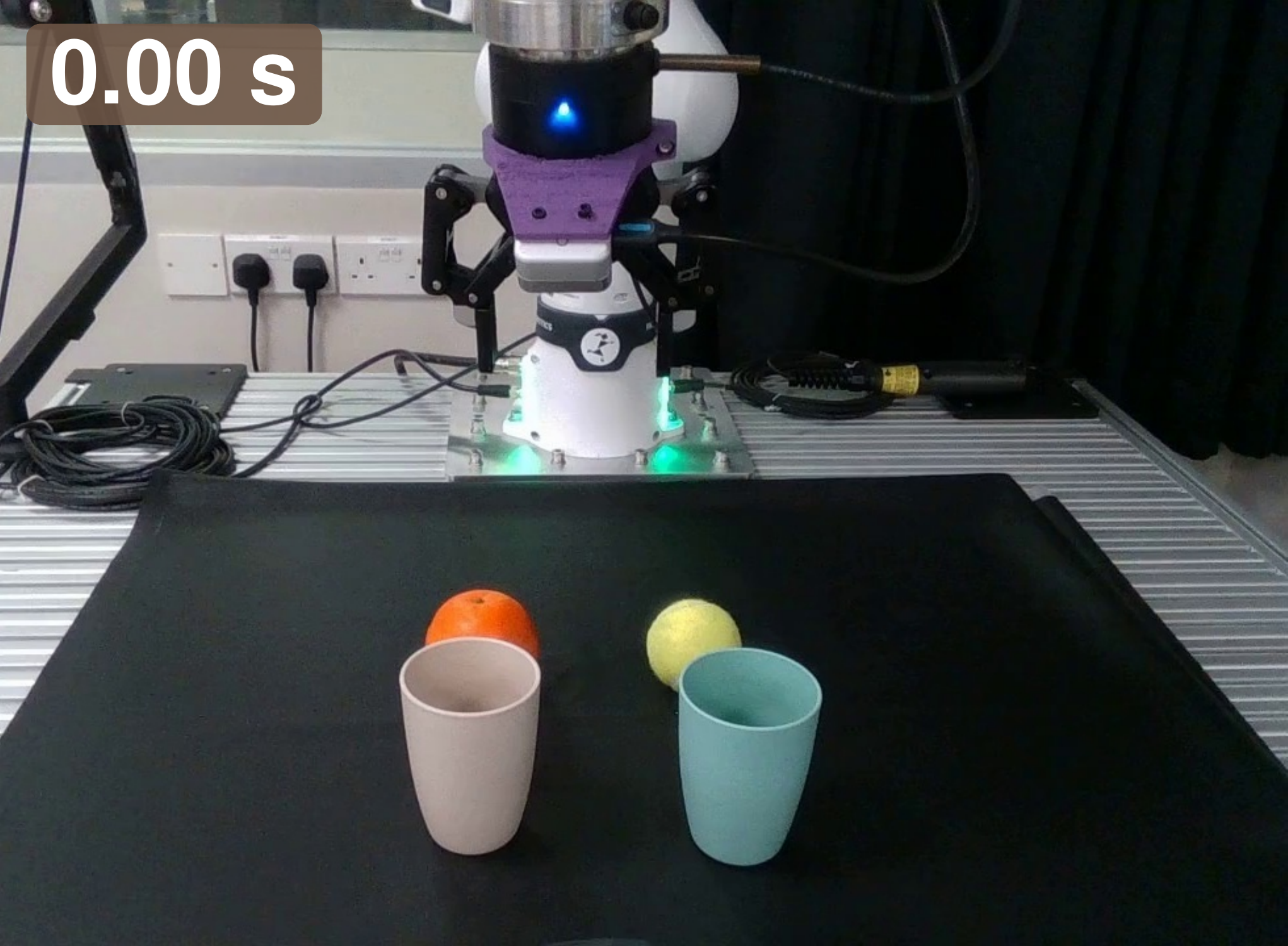}
\caption{Initial.}\end{subfigure}\hfill
\begin{subfigure}{.16\textwidth}\centering
\includegraphics[width=\linewidth]{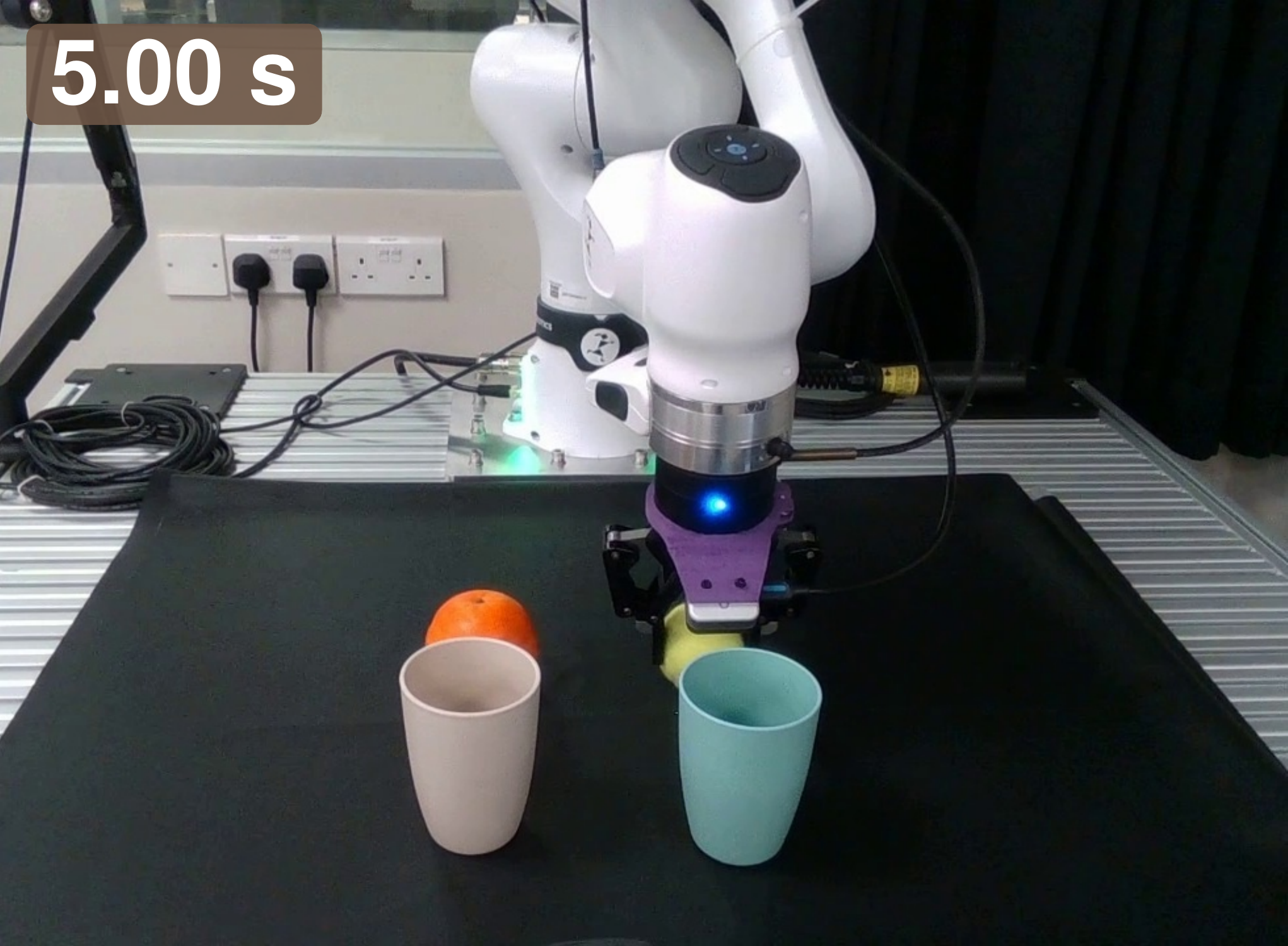}
\caption{Grasp 1.}\end{subfigure}\hfill
\begin{subfigure}{.16\textwidth}\centering
\includegraphics[width=\linewidth]{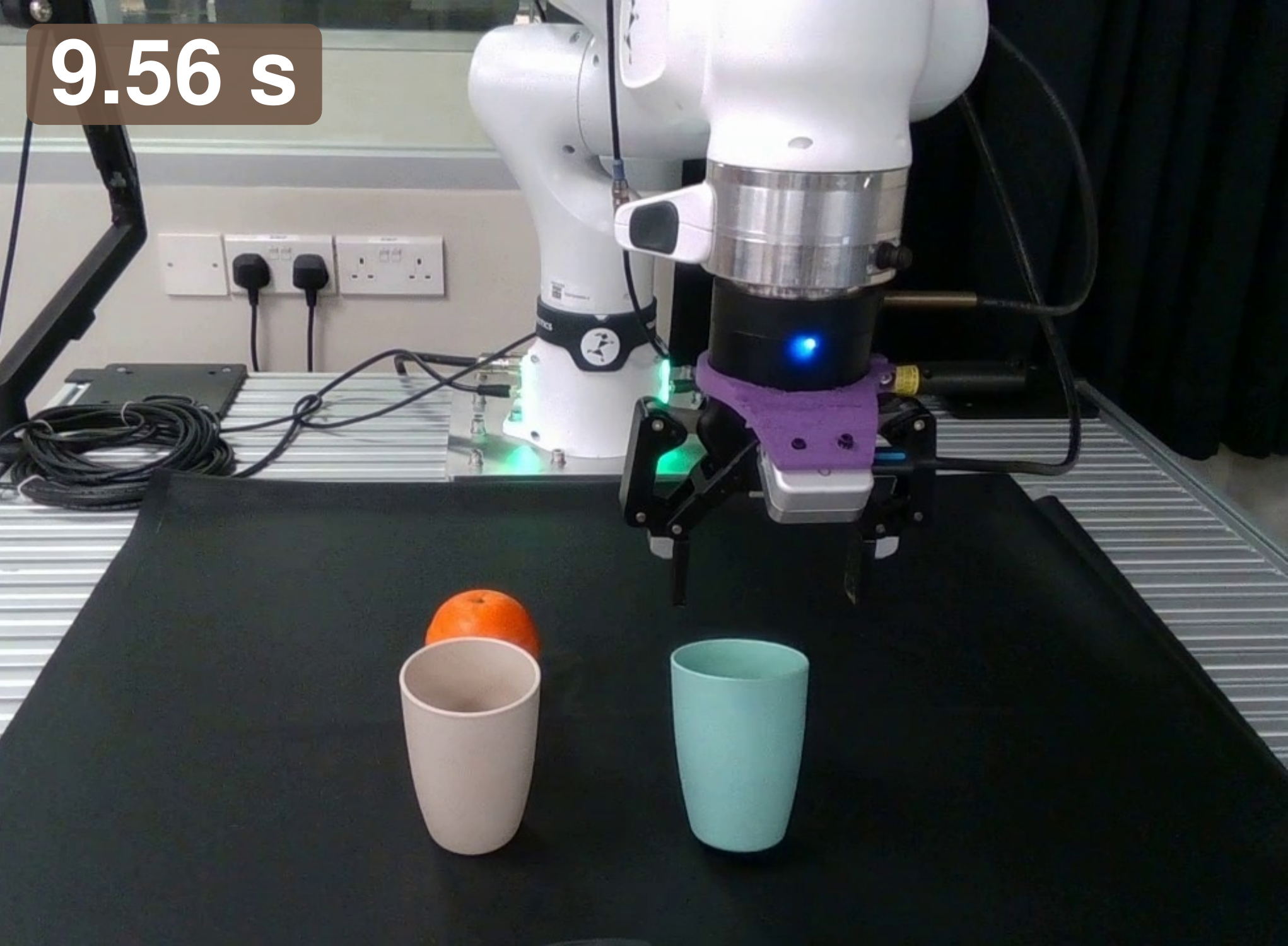}
\caption{Release 1.}\end{subfigure}\hfill
\begin{subfigure}{.16\textwidth}\centering
\includegraphics[width=\linewidth]{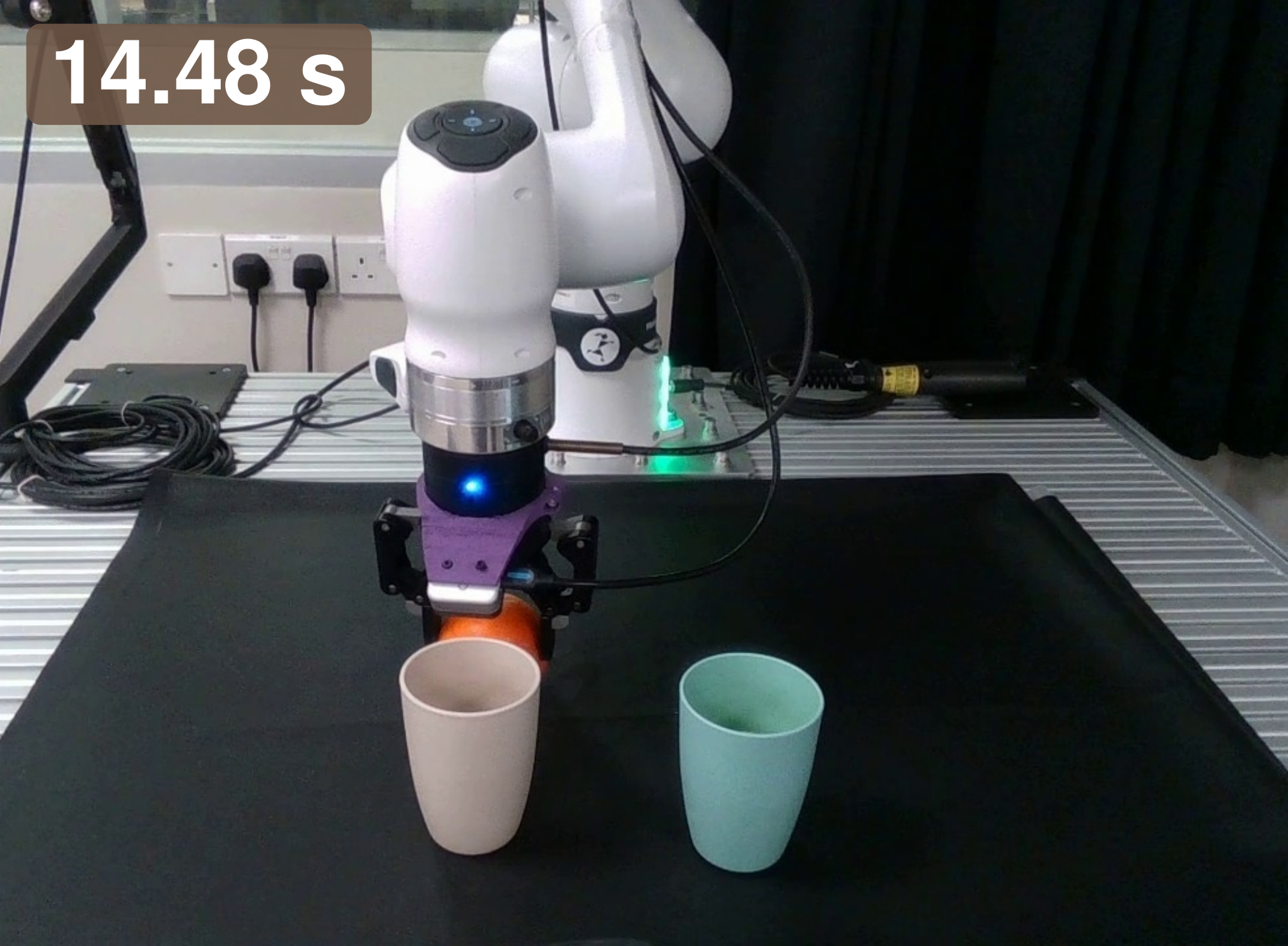}
\caption{Grasp 2.}\end{subfigure}\hfill
\begin{subfigure}{.16\textwidth}\centering
\includegraphics[width=\linewidth]{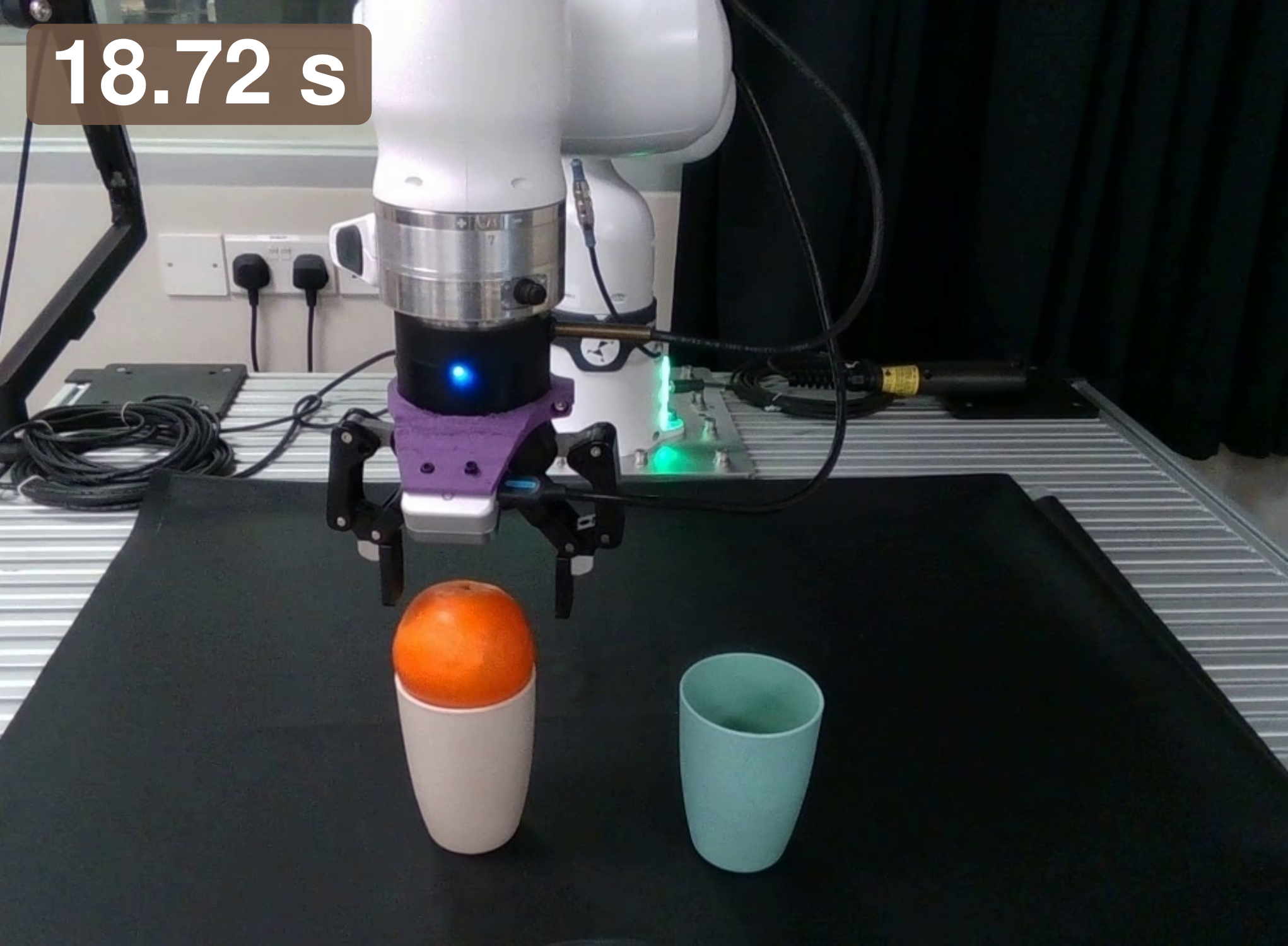}
\caption{Release 2.}\end{subfigure}\hfill
\begin{subfigure}{.16\textwidth}\centering
\includegraphics[width=\linewidth]{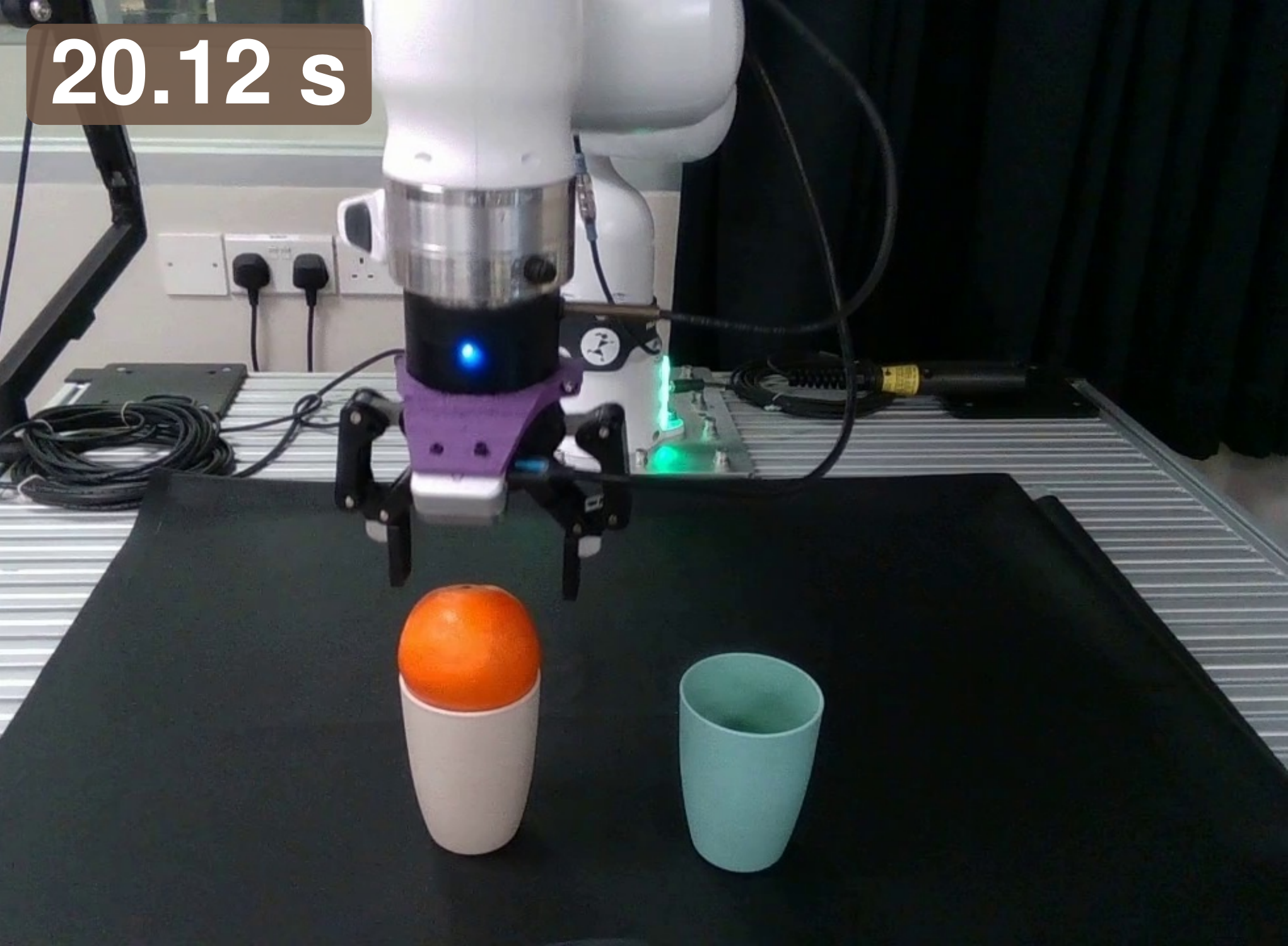}
\caption{Final.}\end{subfigure}\par\vspace{3pt}
\makebox[\textwidth][l]{\footnotesize\textbf{\method}}\par\vspace{2pt}
\begin{subfigure}{.16\textwidth}\centering
\includegraphics[width=\linewidth]{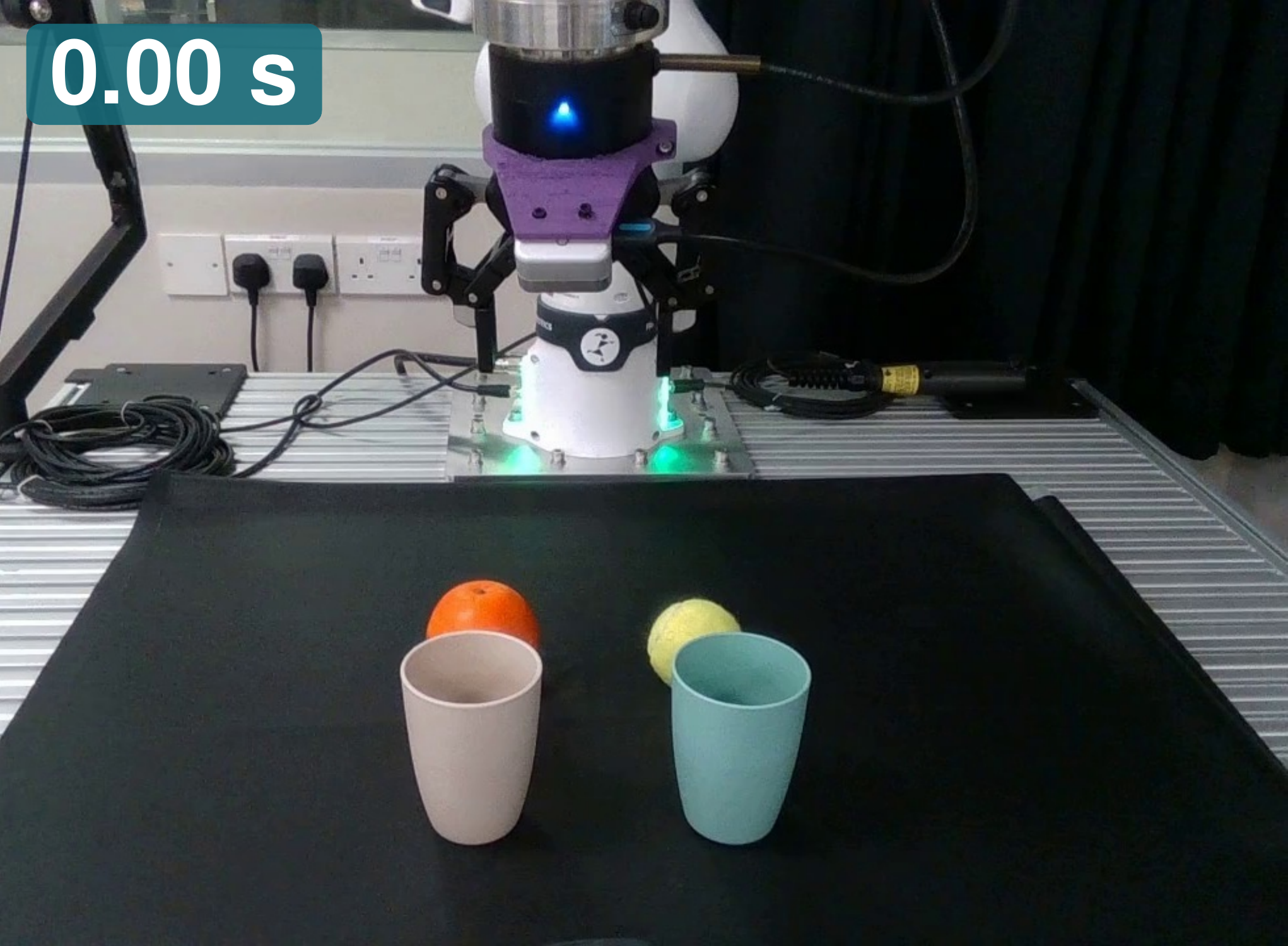}
\caption{Initial.}\end{subfigure}\hfill
\begin{subfigure}{.16\textwidth}\centering
\includegraphics[width=\linewidth]{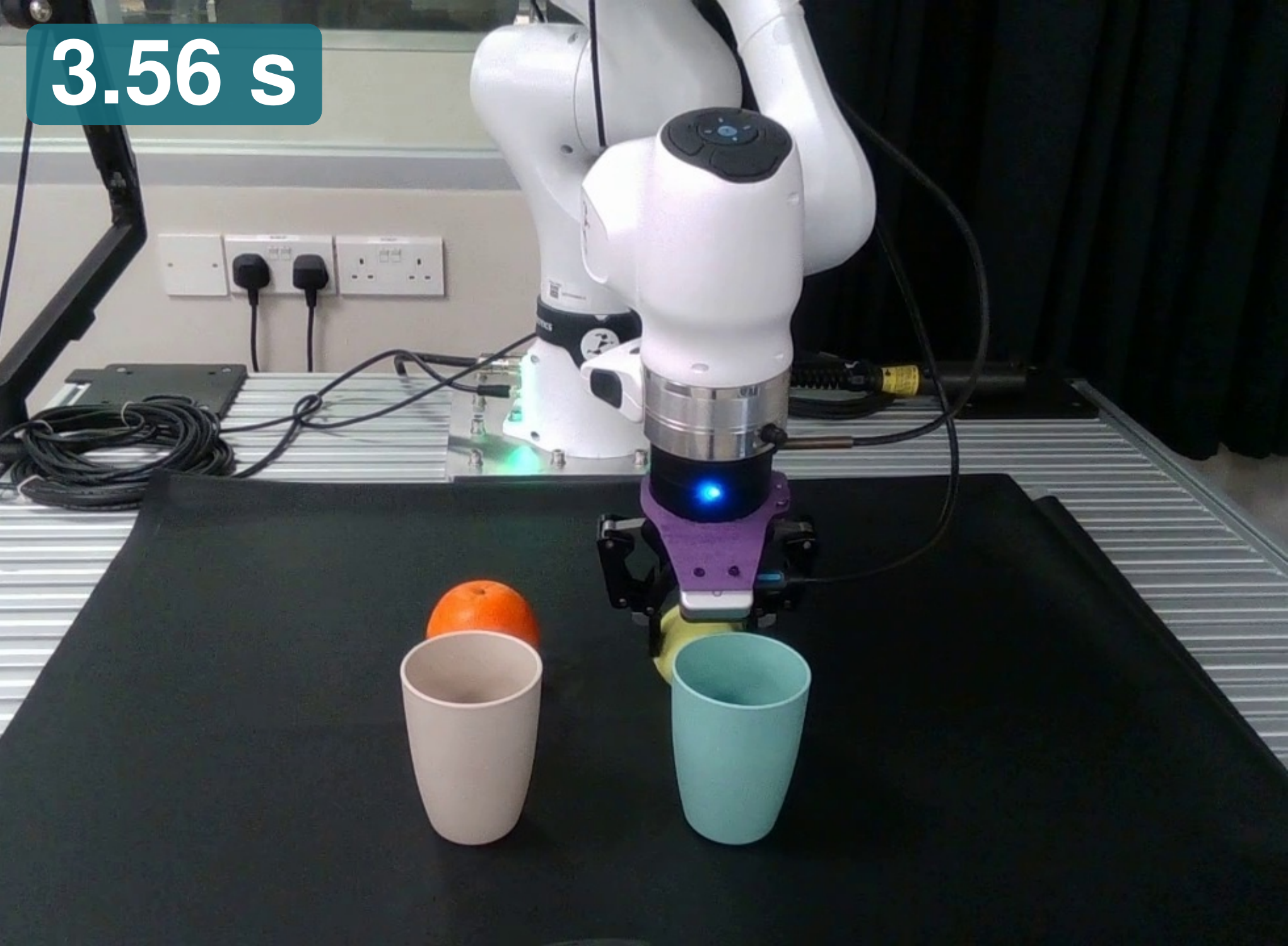}
\caption{Grasp 1.}\end{subfigure}\hfill
\begin{subfigure}{.16\textwidth}\centering
\includegraphics[width=\linewidth]{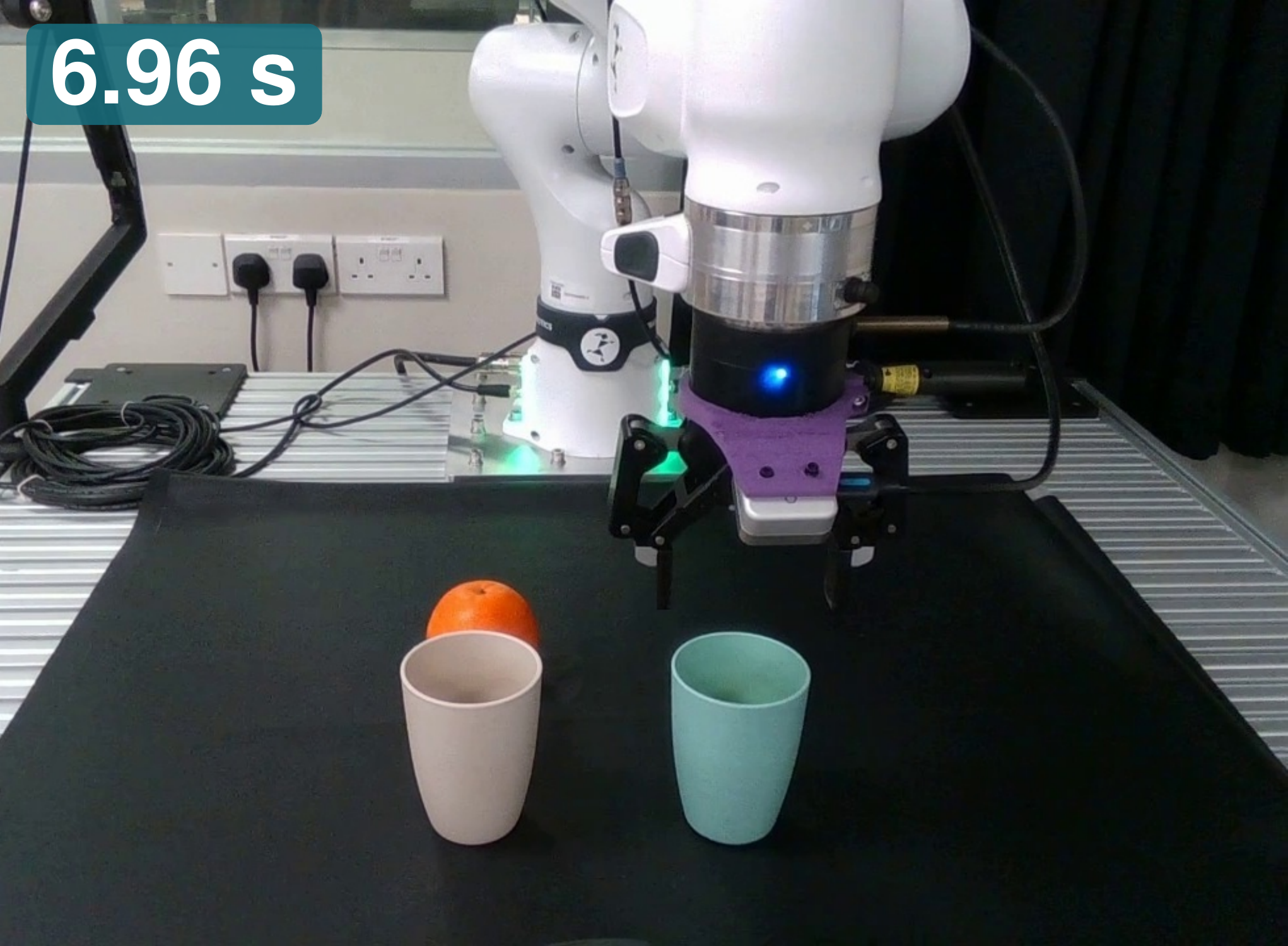}
\caption{Release 1.}\end{subfigure}\hfill
\begin{subfigure}{.16\textwidth}\centering
\includegraphics[width=\linewidth]{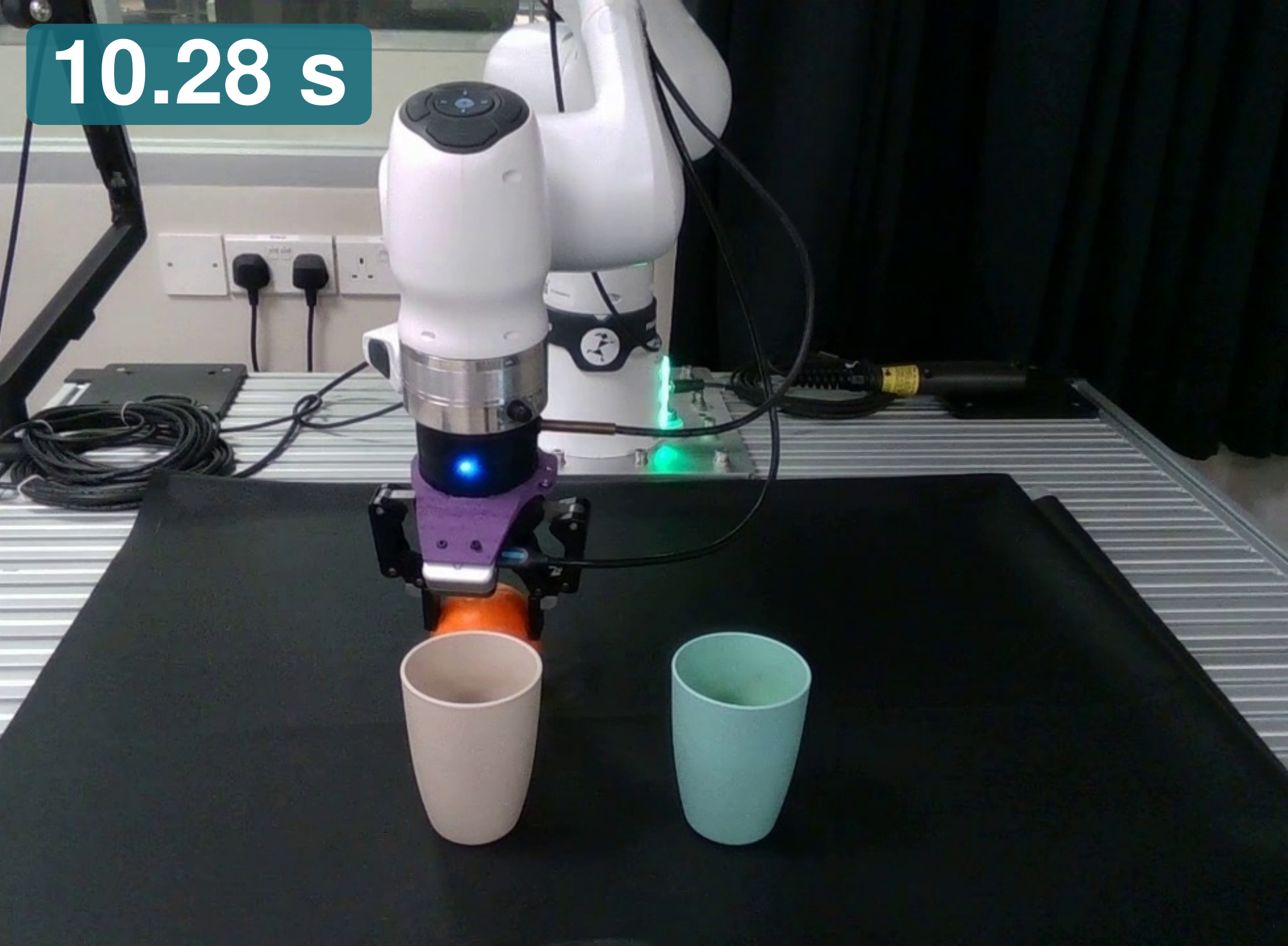}
\caption{Grasp 2.}\end{subfigure}\hfill
\begin{subfigure}{.16\textwidth}\centering
\includegraphics[width=\linewidth]{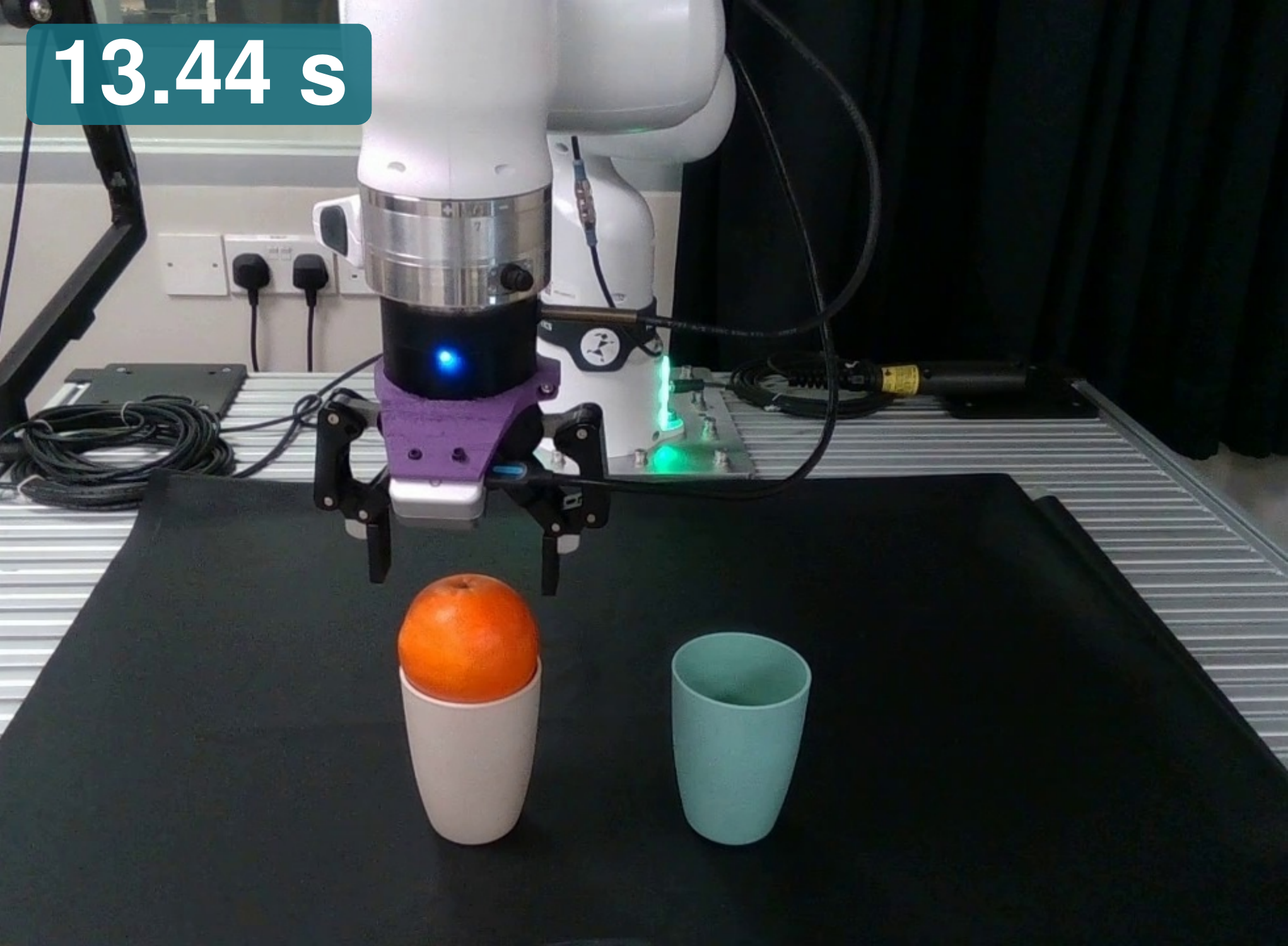}
\caption{Release 2.}\end{subfigure}\hfill
\begin{subfigure}{.16\textwidth}\centering
\includegraphics[width=\linewidth]{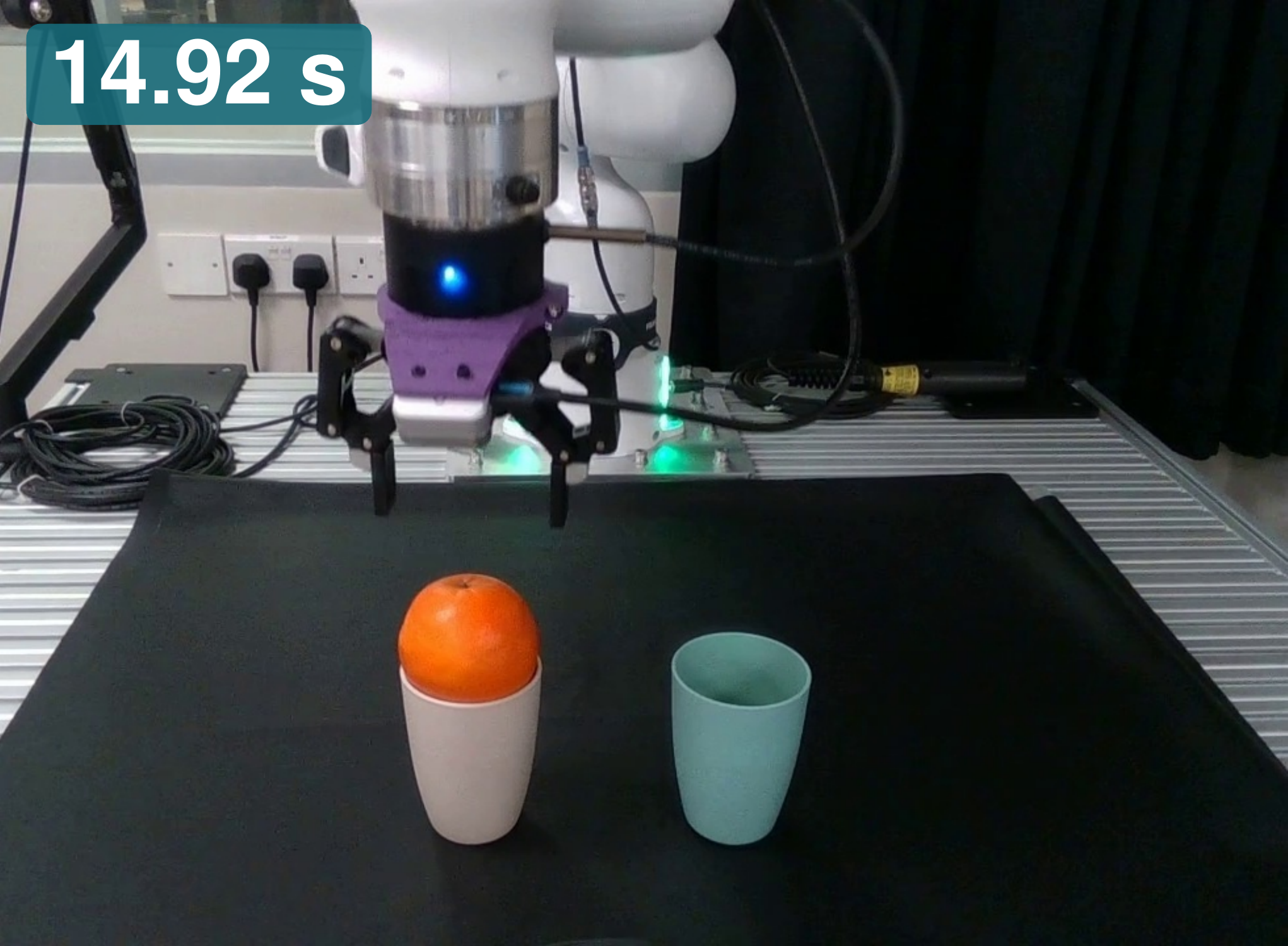}
\caption{Final.}\end{subfigure}\par\smallskip
\caption{\textbf{Sequential fruit placement: matched sequence 3.} Prompt: \textit{``pick tennis ball on green cup then orange on brown cup.''} Rows compare $\pi_{0.5}$ above and \method below across six event-defined stages under shared framing, with top-left badges indicating elapsed trial time.}
\label{fig:motion_pair_3}
\end{figure}

\begin{figure}[H]\centering
\captionsetup[subfigure]{font={scriptsize},skip=2pt}
\makebox[\textwidth][l]{\footnotesize\textbf{$\pi_{0.5}$}}\par\vspace{2pt}
\begin{subfigure}{.16\textwidth}\centering
\includegraphics[width=\linewidth]{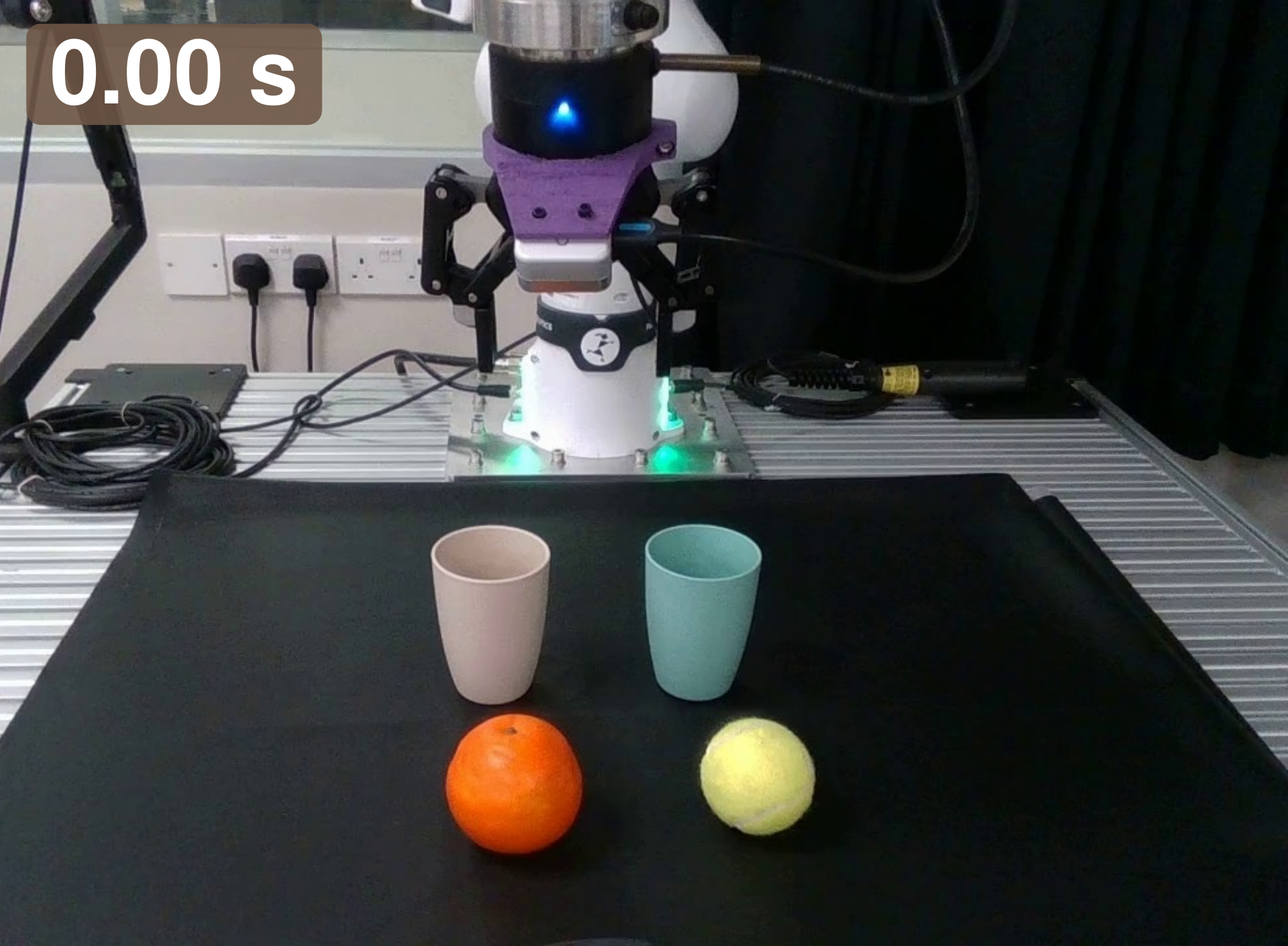}
\caption{Initial.}\end{subfigure}\hfill
\begin{subfigure}{.16\textwidth}\centering
\includegraphics[width=\linewidth]{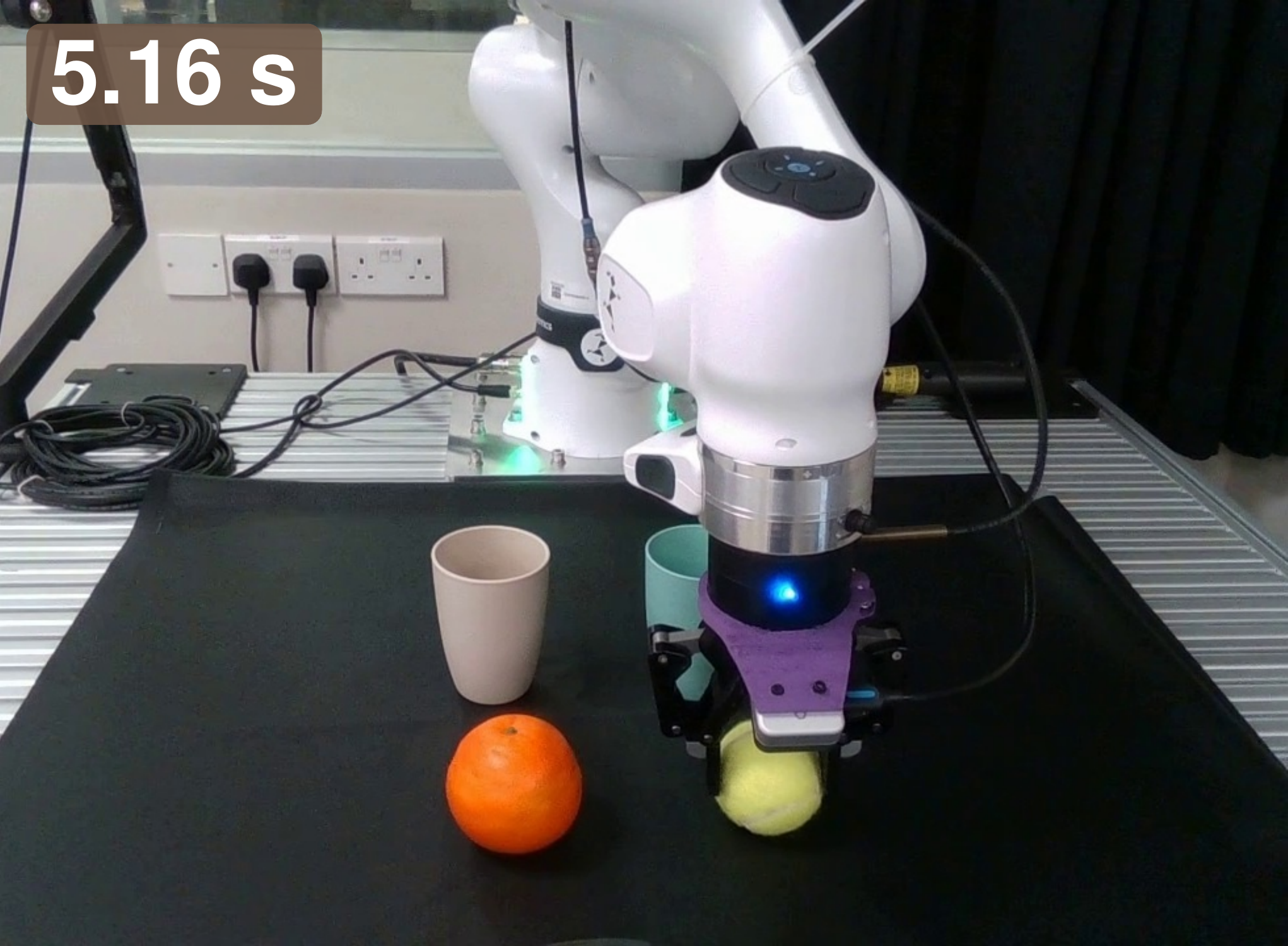}
\caption{Grasp 1.}\end{subfigure}\hfill
\begin{subfigure}{.16\textwidth}\centering
\includegraphics[width=\linewidth]{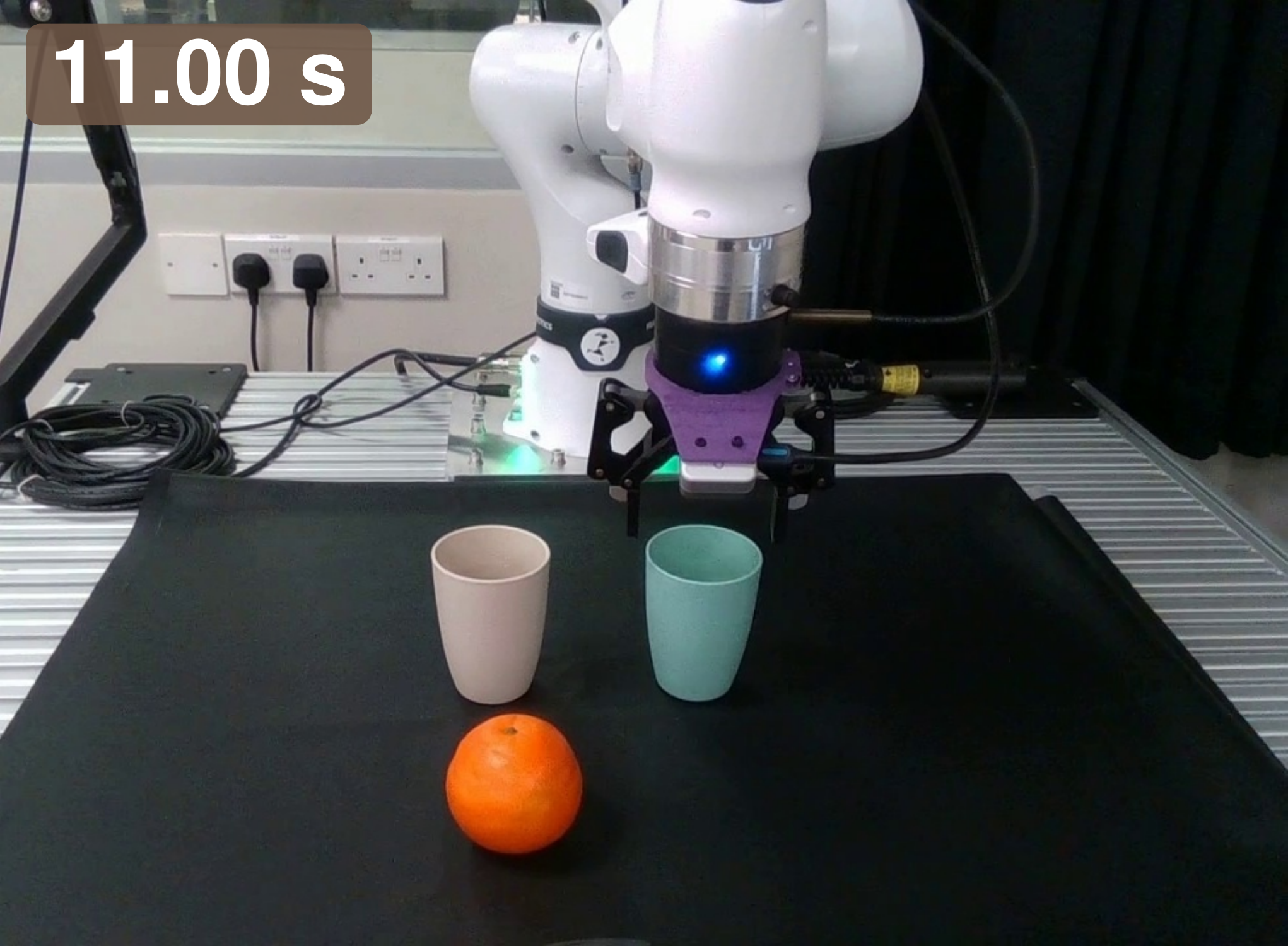}
\caption{Release 1.}\end{subfigure}\hfill
\begin{subfigure}{.16\textwidth}\centering
\includegraphics[width=\linewidth]{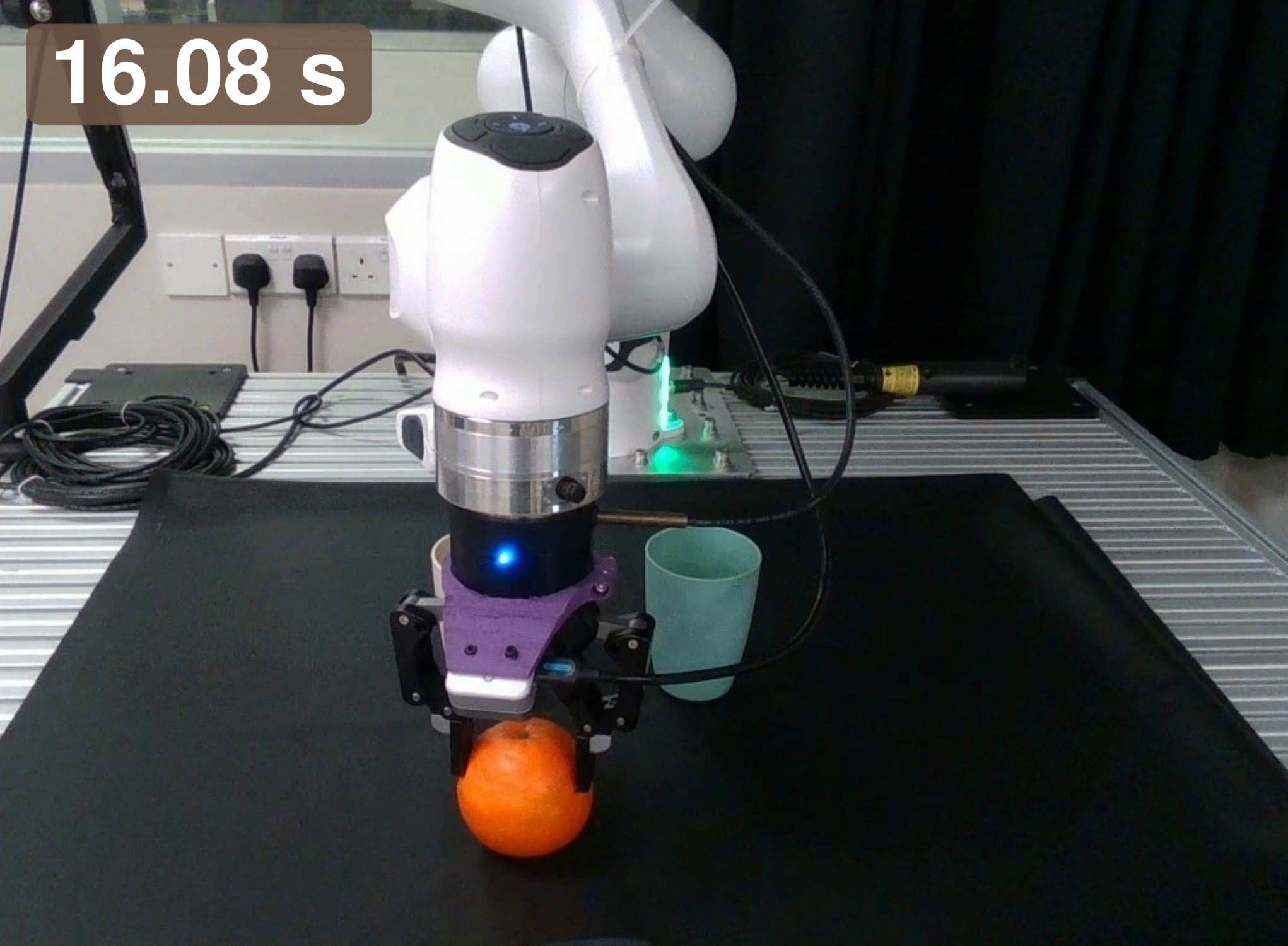}
\caption{Grasp 2.}\end{subfigure}\hfill
\begin{subfigure}{.16\textwidth}\centering
\includegraphics[width=\linewidth]{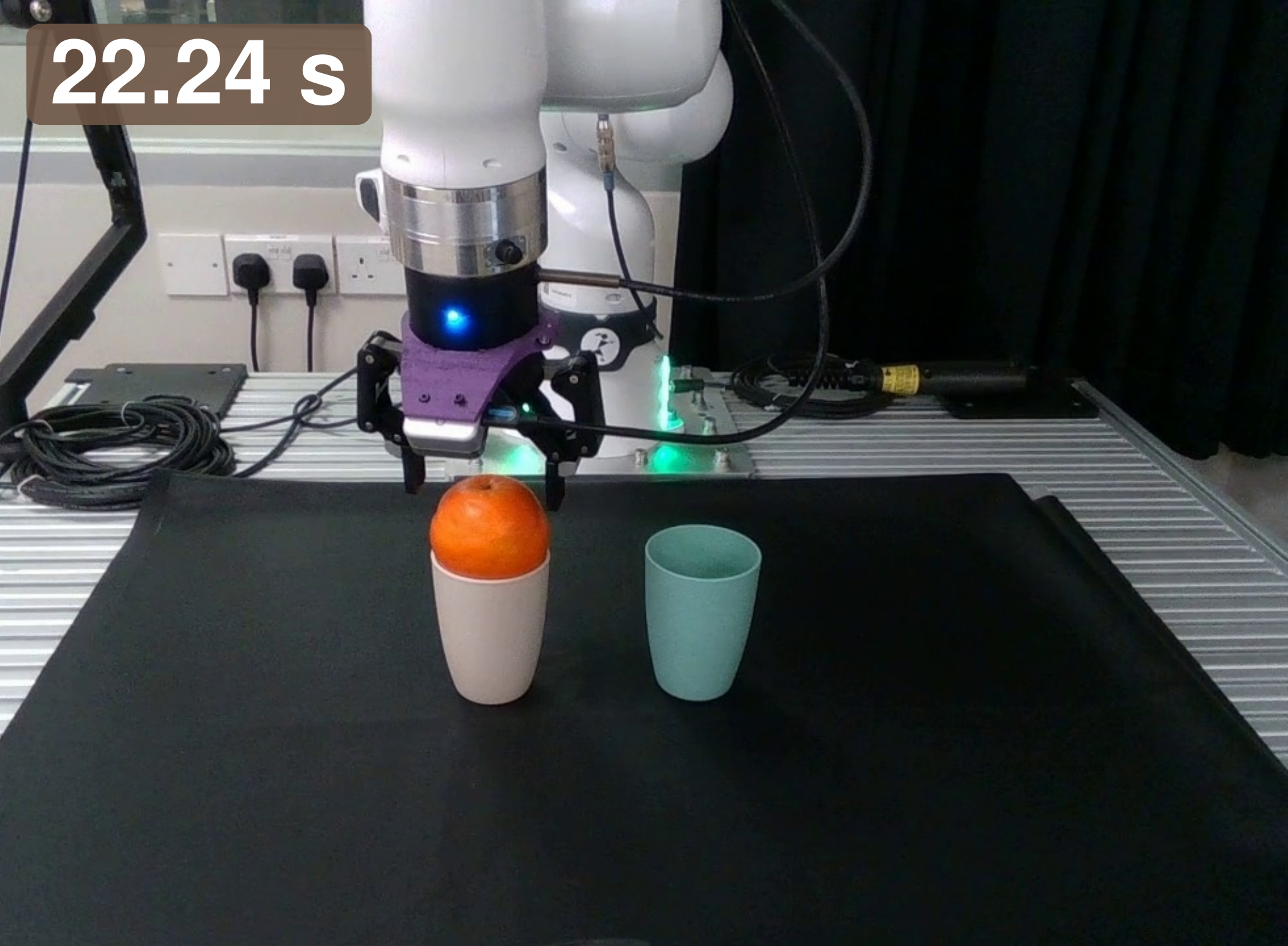}
\caption{Release 2.}\end{subfigure}\hfill
\begin{subfigure}{.16\textwidth}\centering
\includegraphics[width=\linewidth]{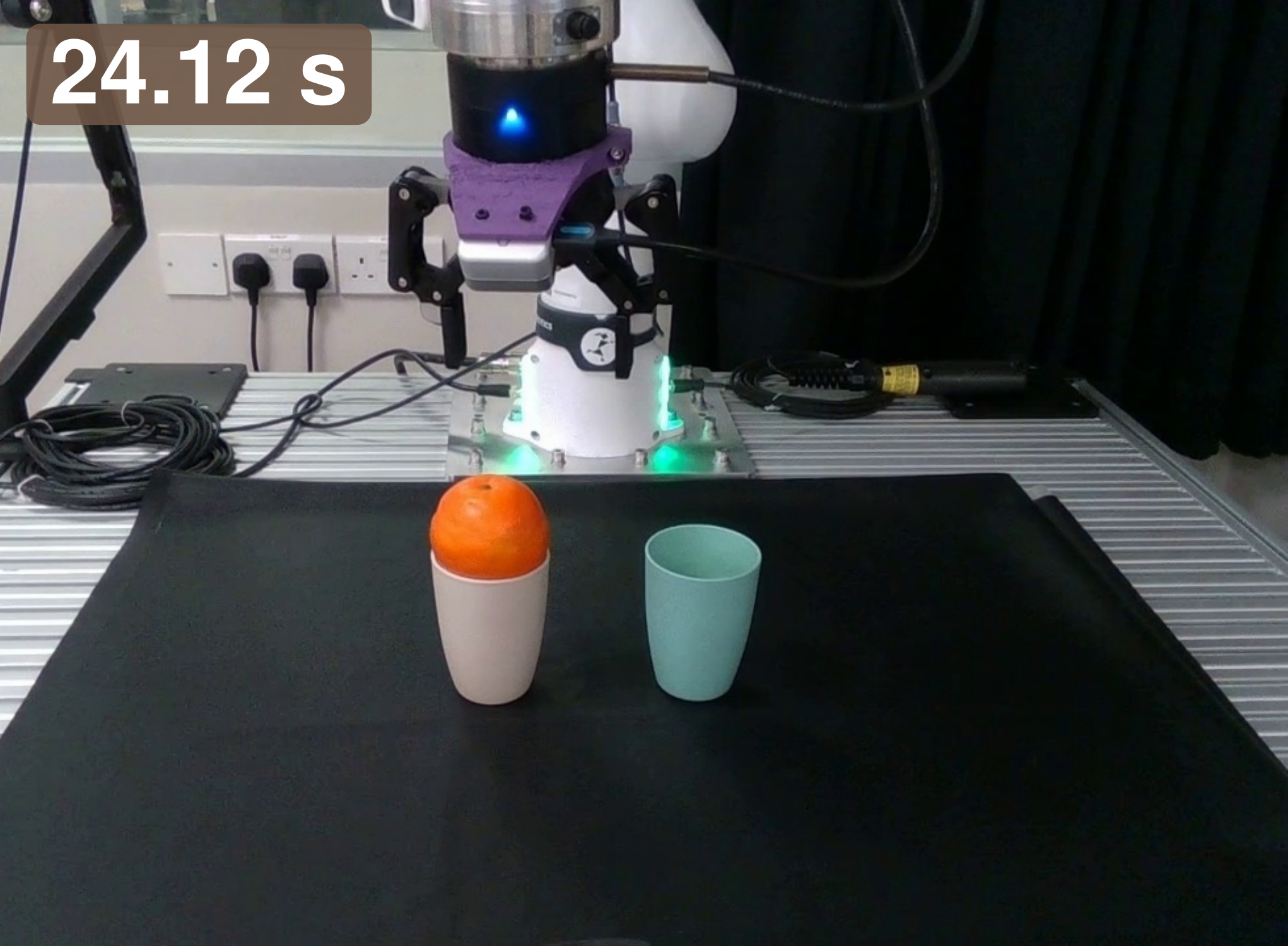}
\caption{Final.}\end{subfigure}\par\vspace{3pt}
\makebox[\textwidth][l]{\footnotesize\textbf{\method}}\par\vspace{2pt}
\begin{subfigure}{.16\textwidth}\centering
\includegraphics[width=\linewidth]{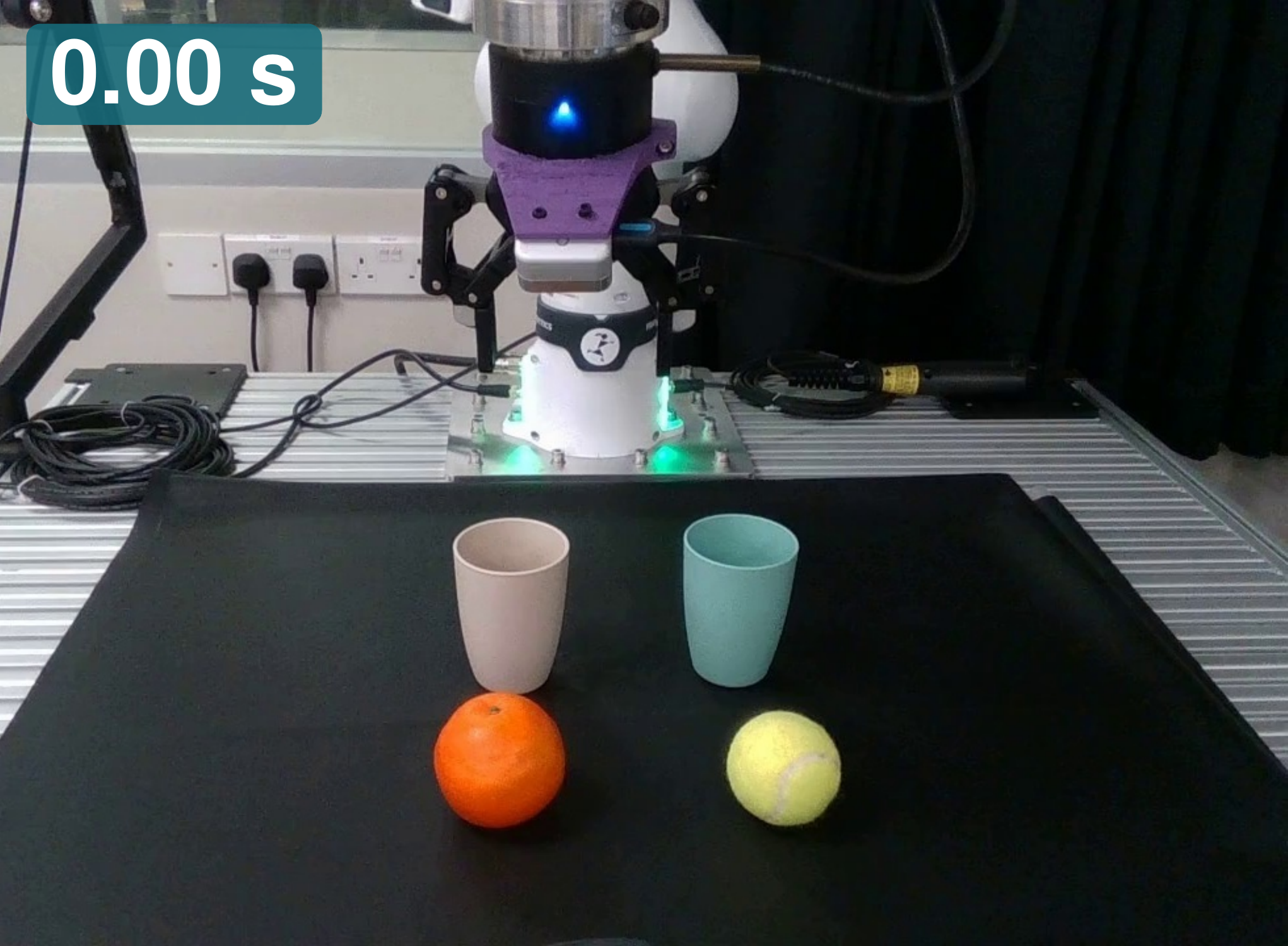}
\caption{Initial.}\end{subfigure}\hfill
\begin{subfigure}{.16\textwidth}\centering
\includegraphics[width=\linewidth]{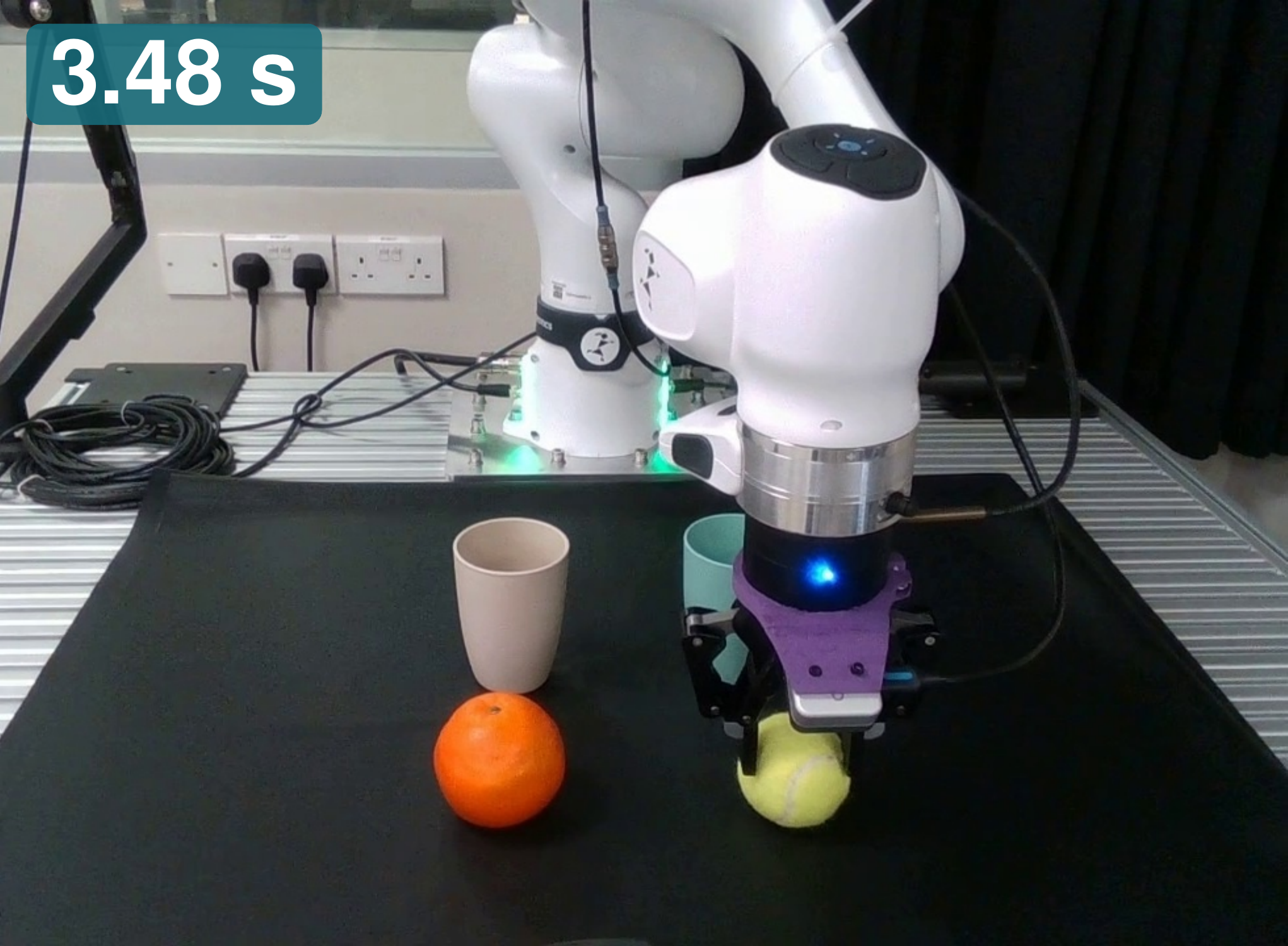}
\caption{Grasp 1.}\end{subfigure}\hfill
\begin{subfigure}{.16\textwidth}\centering
\includegraphics[width=\linewidth]{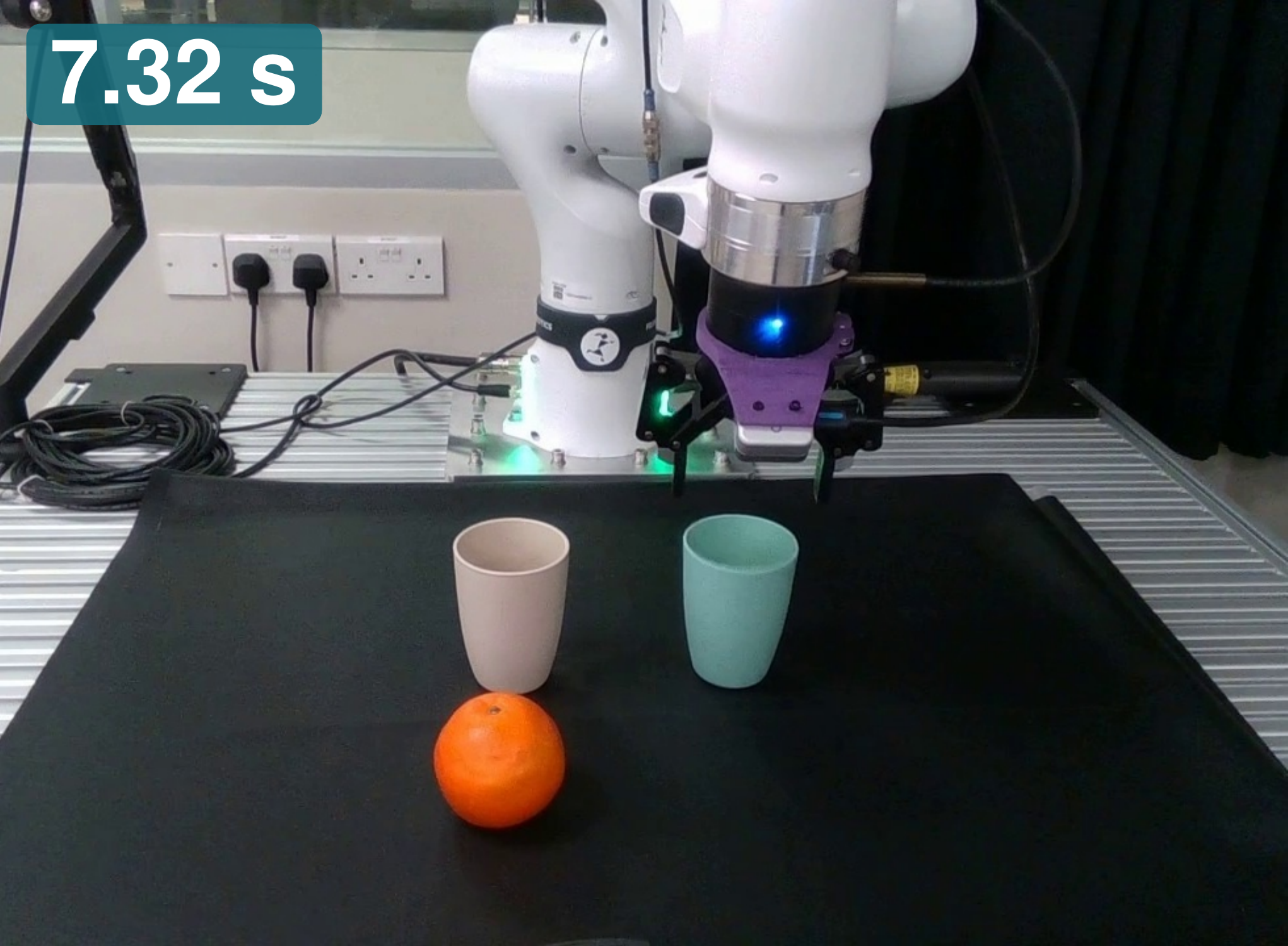}
\caption{Release 1.}\end{subfigure}\hfill
\begin{subfigure}{.16\textwidth}\centering
\includegraphics[width=\linewidth]{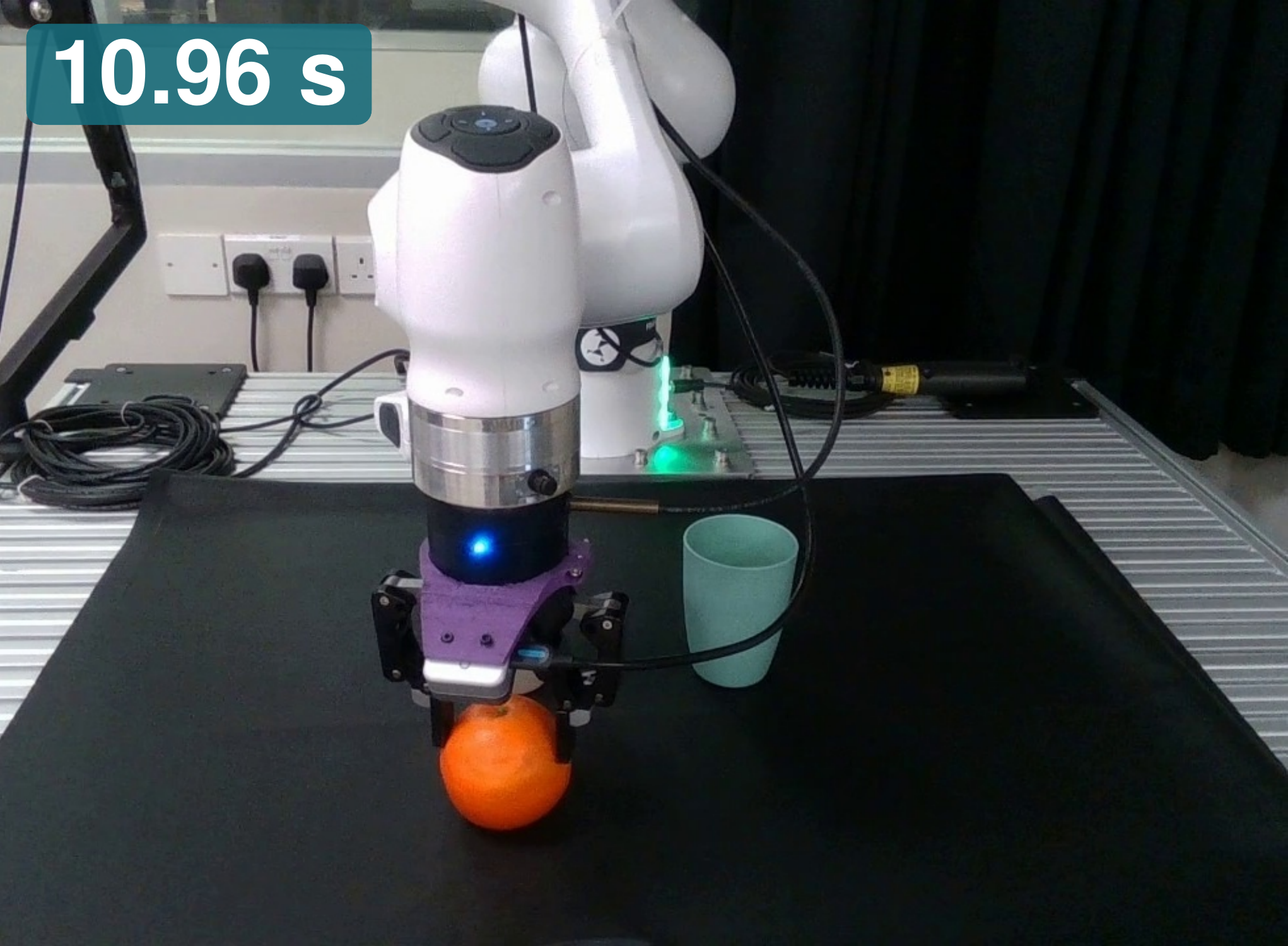}
\caption{Grasp 2.}\end{subfigure}\hfill
\begin{subfigure}{.16\textwidth}\centering
\includegraphics[width=\linewidth]{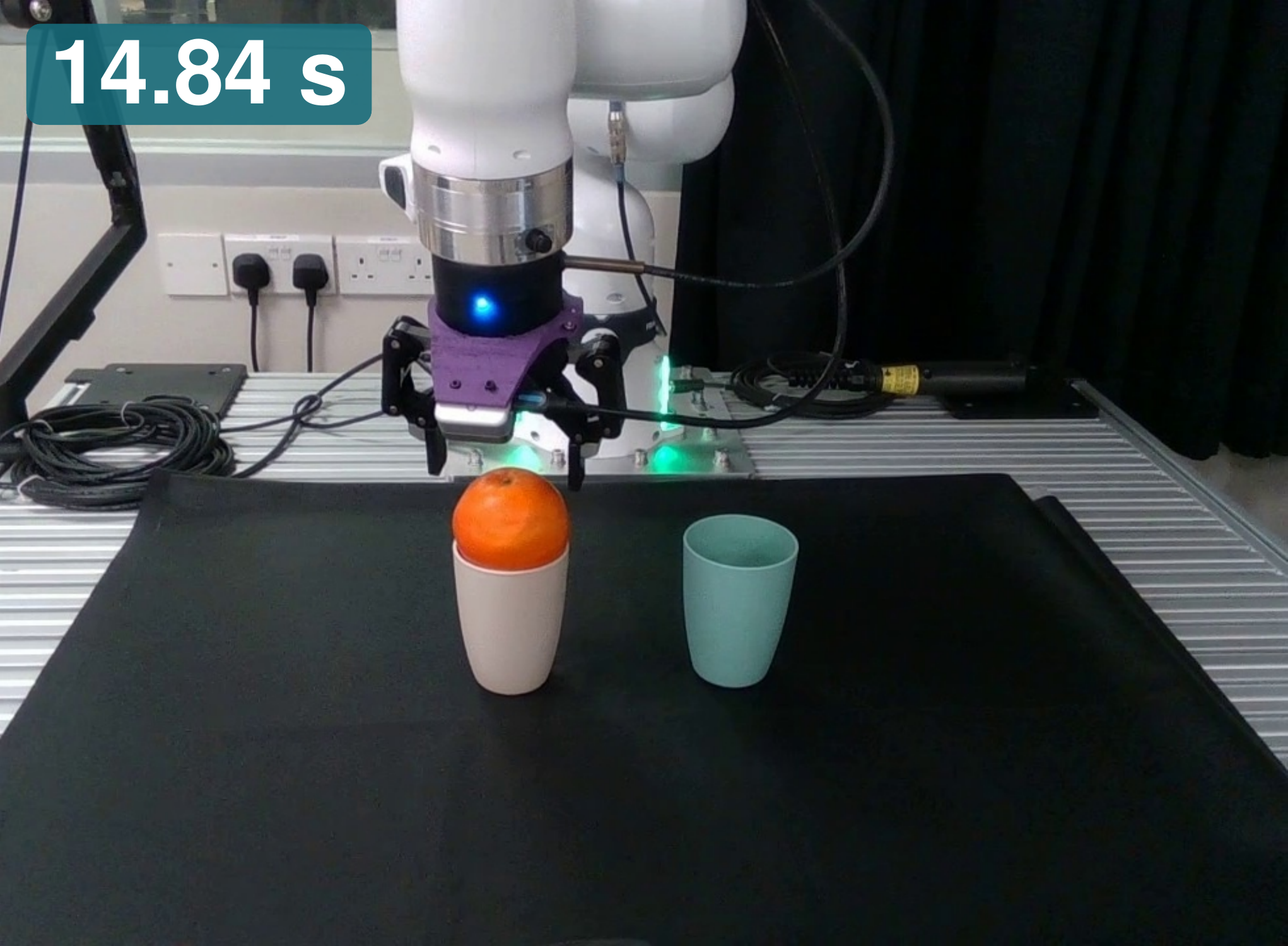}
\caption{Release 2.}\end{subfigure}\hfill
\begin{subfigure}{.16\textwidth}\centering
\includegraphics[width=\linewidth]{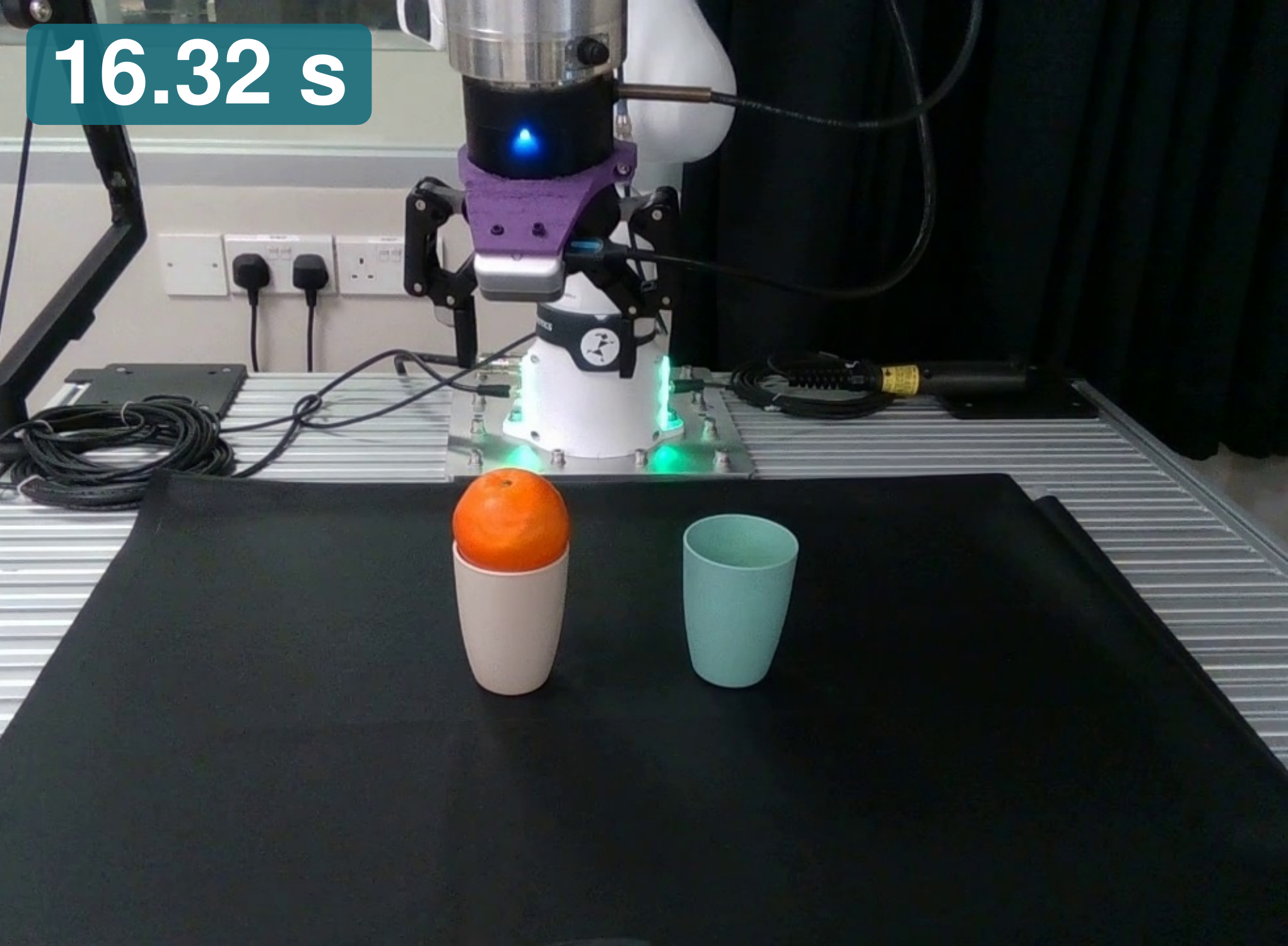}
\caption{Final.}\end{subfigure}\par\smallskip
\caption{\textbf{Sequential fruit placement: matched sequence 4.} Prompt: \textit{``pick tennis ball on green cup then orange on brown cup.''} Rows compare $\pi_{0.5}$ above and \method below across six event-defined stages under shared framing, with top-left badges indicating elapsed trial time.}
\label{fig:motion_pair_4}
\end{figure}

\begin{figure}[H]\centering
\captionsetup[subfigure]{font={scriptsize},skip=2pt}
\makebox[\textwidth][l]{\footnotesize\textbf{$\pi_{0.5}$}}\par\vspace{2pt}
\begin{subfigure}{.16\textwidth}\centering
\includegraphics[width=\linewidth]{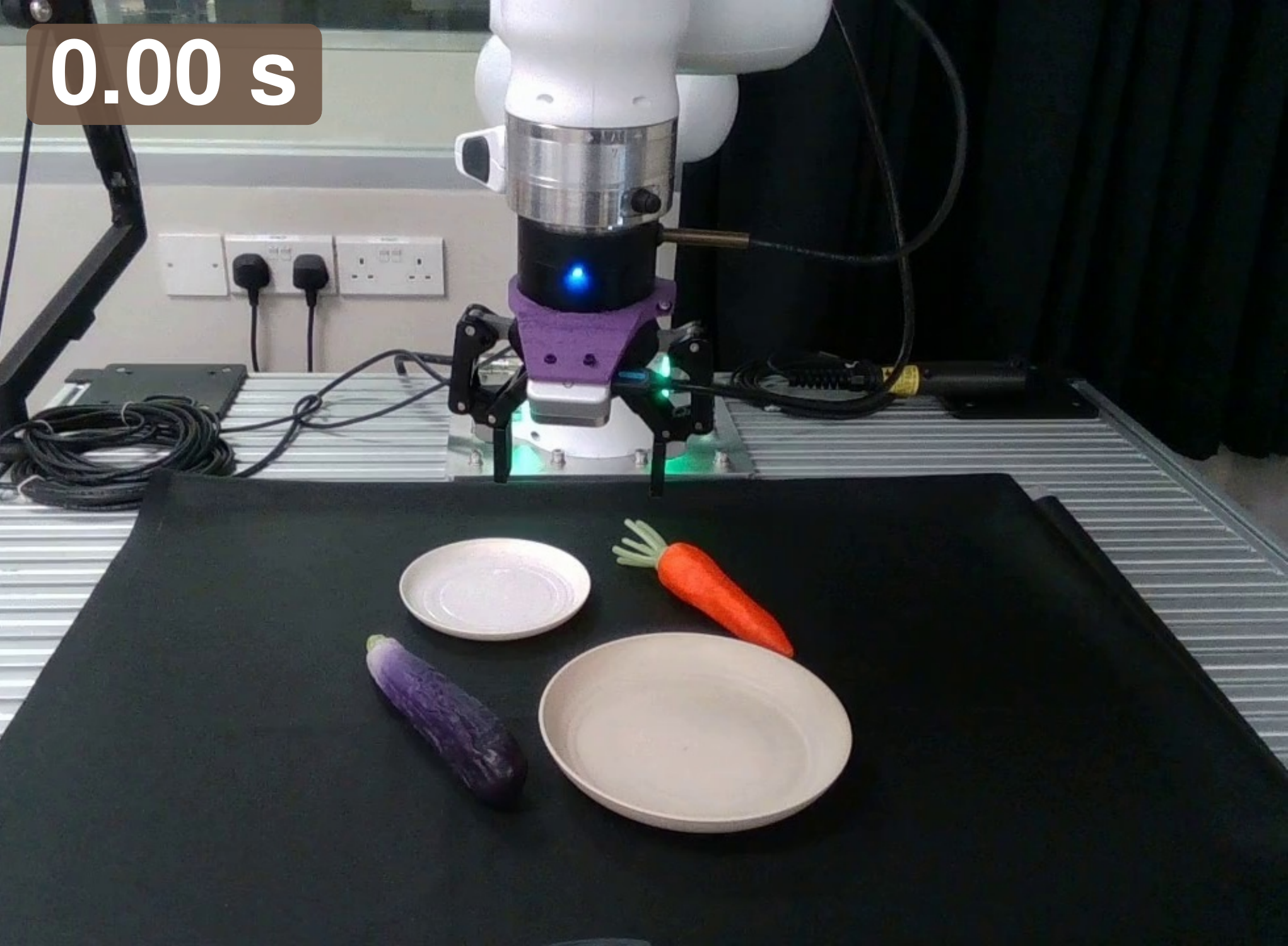}
\caption{Initial.}\end{subfigure}\hfill
\begin{subfigure}{.16\textwidth}\centering
\includegraphics[width=\linewidth]{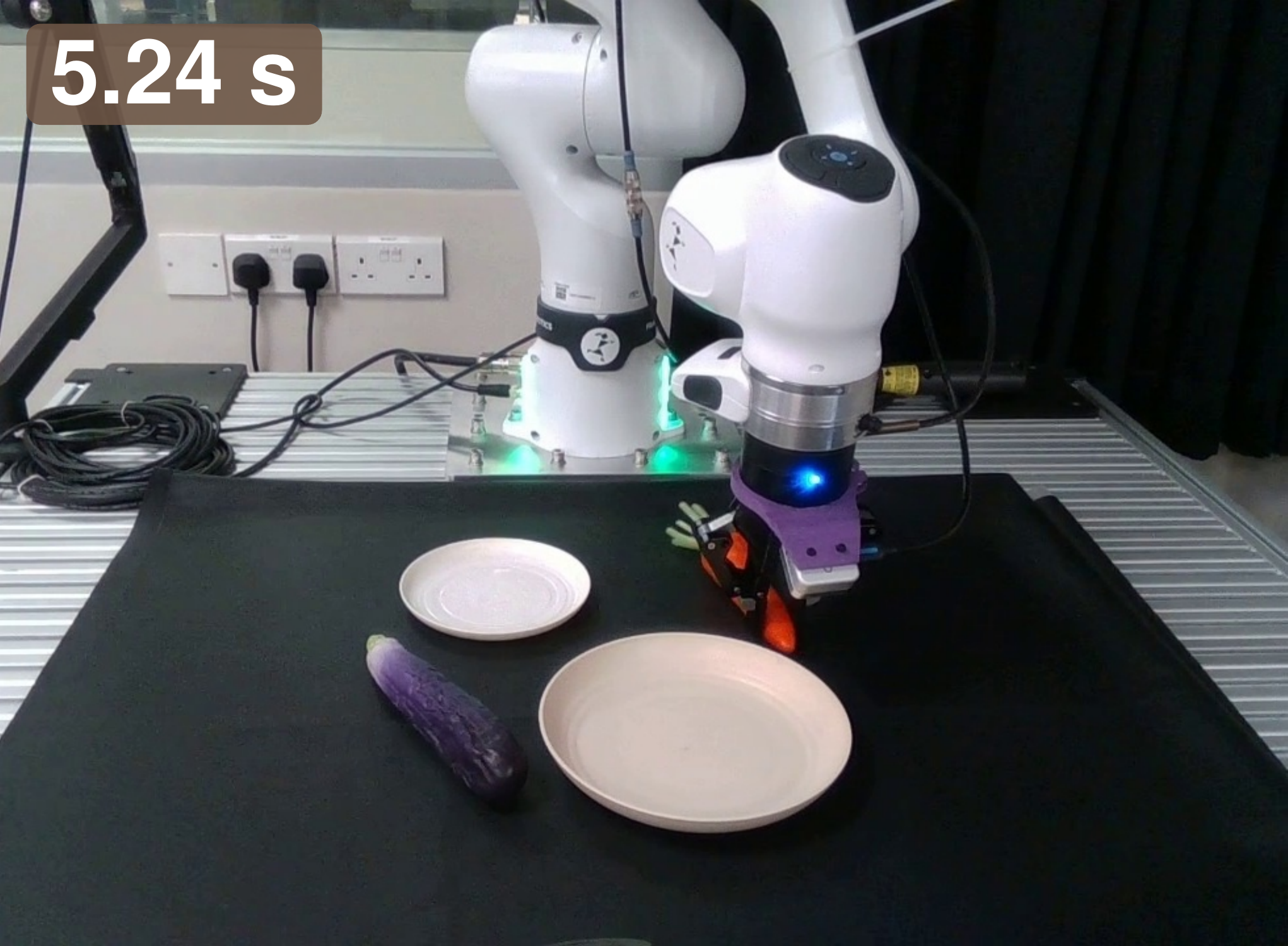}
\caption{Grasp 1.}\end{subfigure}\hfill
\begin{subfigure}{.16\textwidth}\centering
\includegraphics[width=\linewidth]{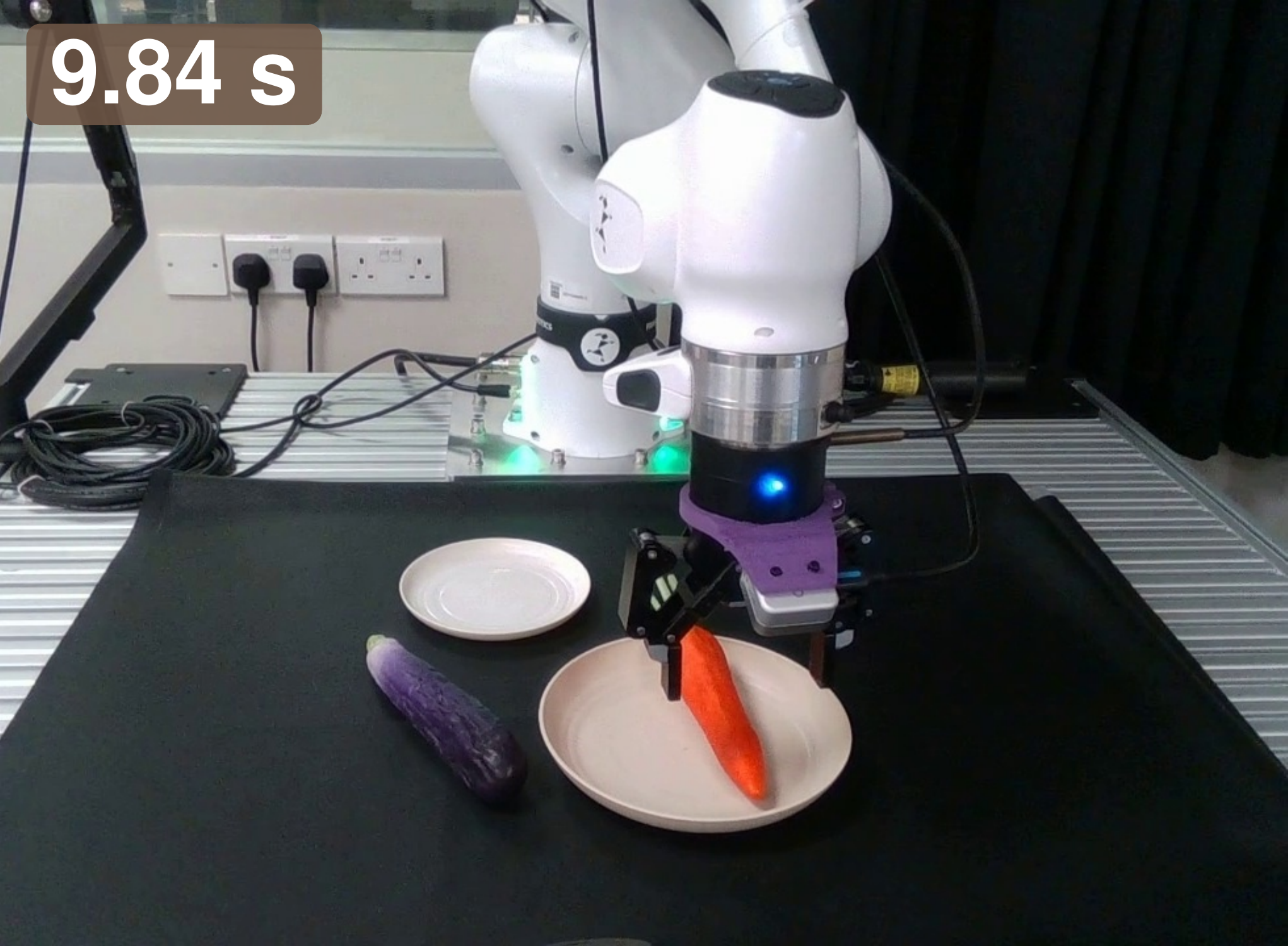}
\caption{Release 1.}\end{subfigure}\hfill
\begin{subfigure}{.16\textwidth}\centering
\includegraphics[width=\linewidth]{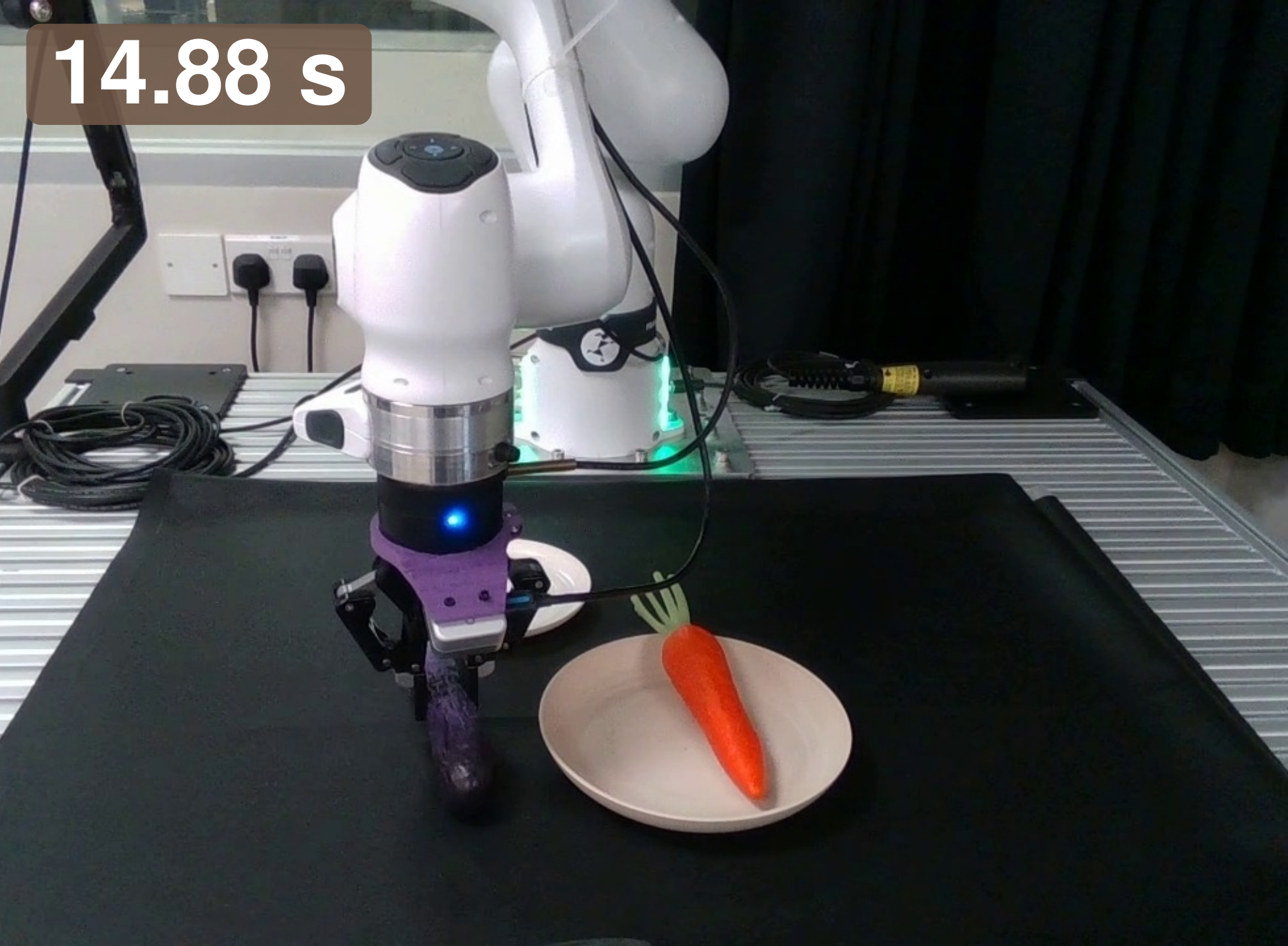}
\caption{Grasp 2.}\end{subfigure}\hfill
\begin{subfigure}{.16\textwidth}\centering
\includegraphics[width=\linewidth]{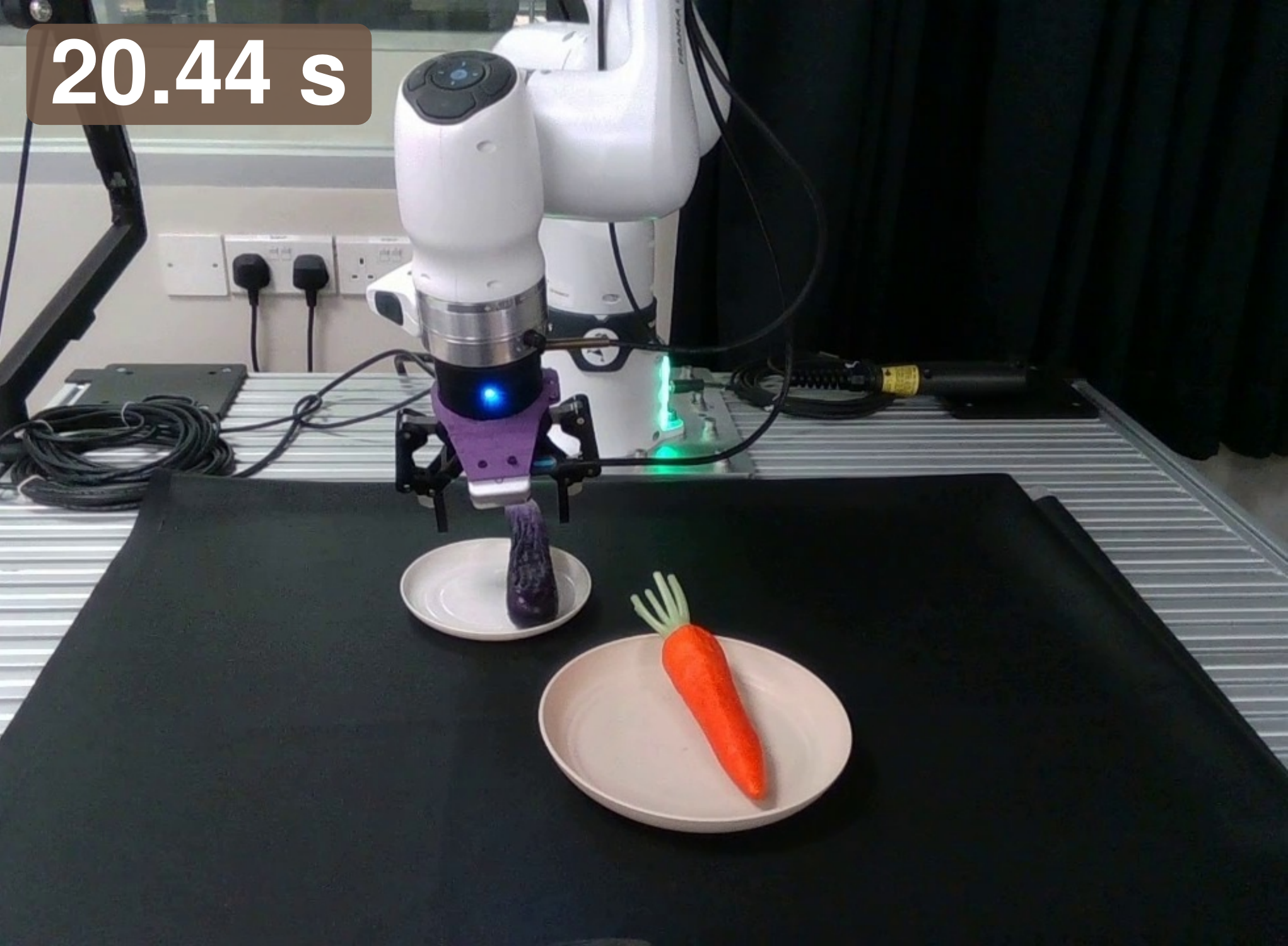}
\caption{Release 2.}\end{subfigure}\hfill
\begin{subfigure}{.16\textwidth}\centering
\includegraphics[width=\linewidth]{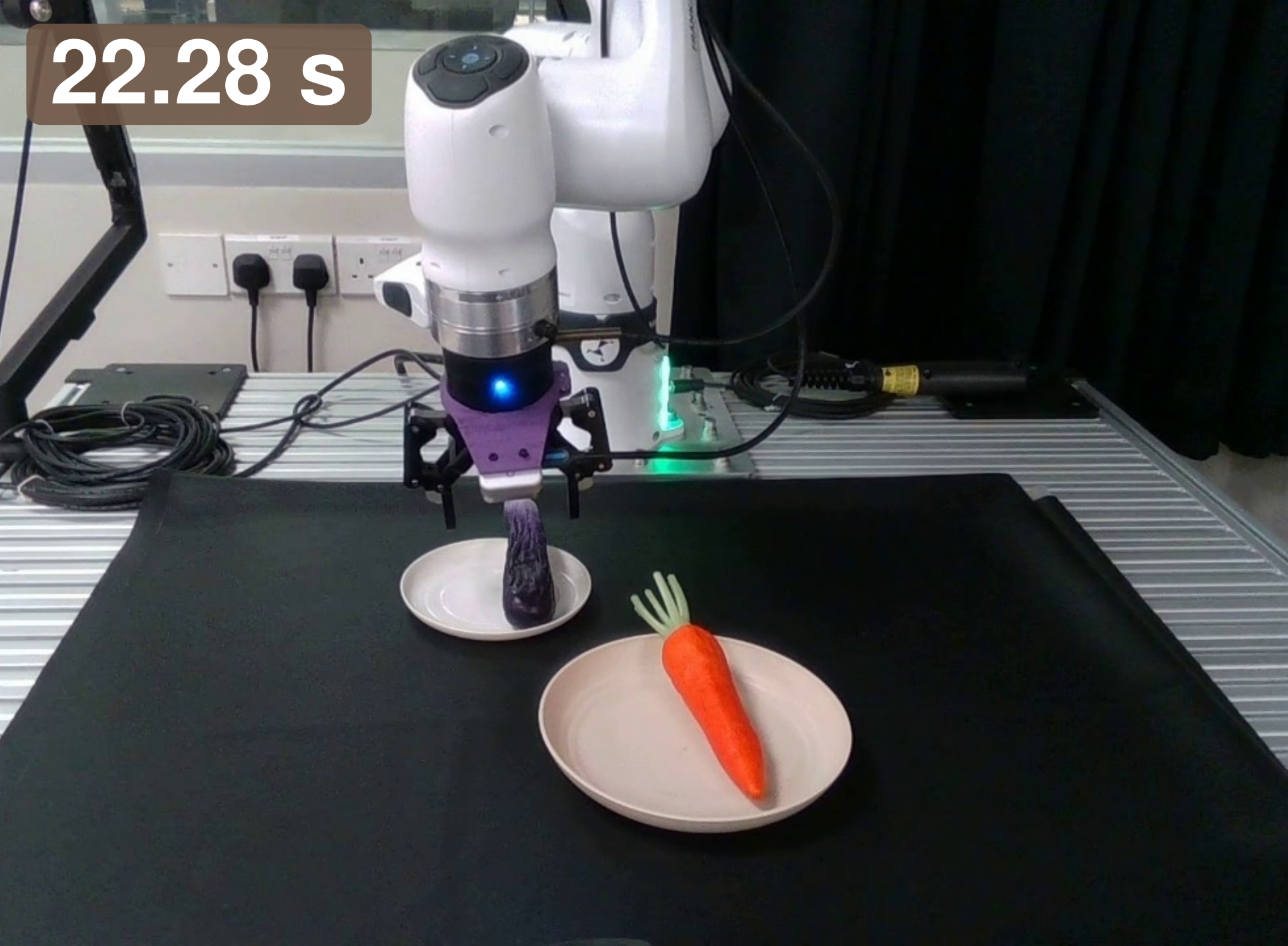}
\caption{Final.}\end{subfigure}\par\vspace{3pt}
\makebox[\textwidth][l]{\footnotesize\textbf{\method}}\par\vspace{2pt}
\begin{subfigure}{.16\textwidth}\centering
\includegraphics[width=\linewidth]{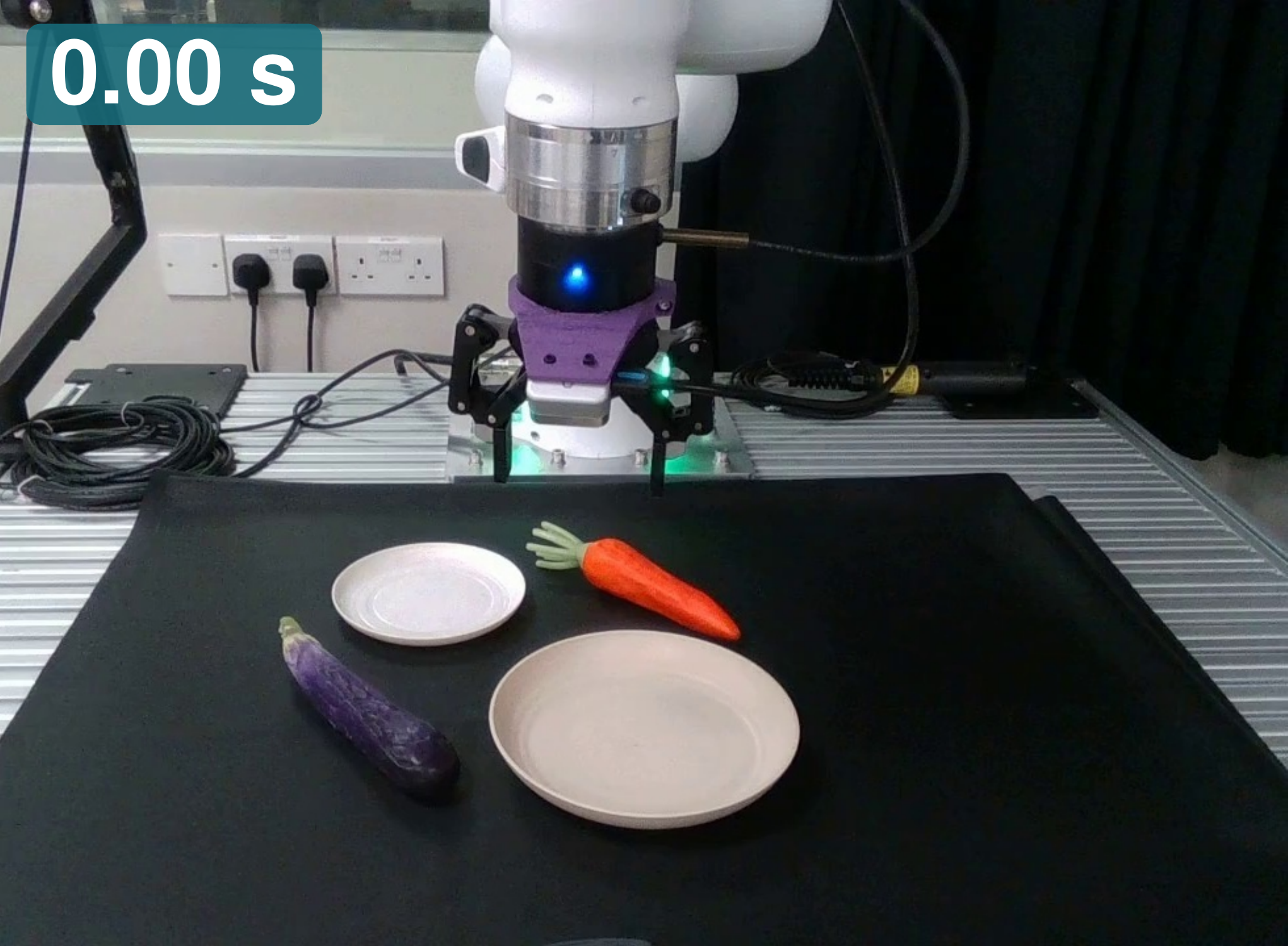}
\caption{Initial.}\end{subfigure}\hfill
\begin{subfigure}{.16\textwidth}\centering
\includegraphics[width=\linewidth]{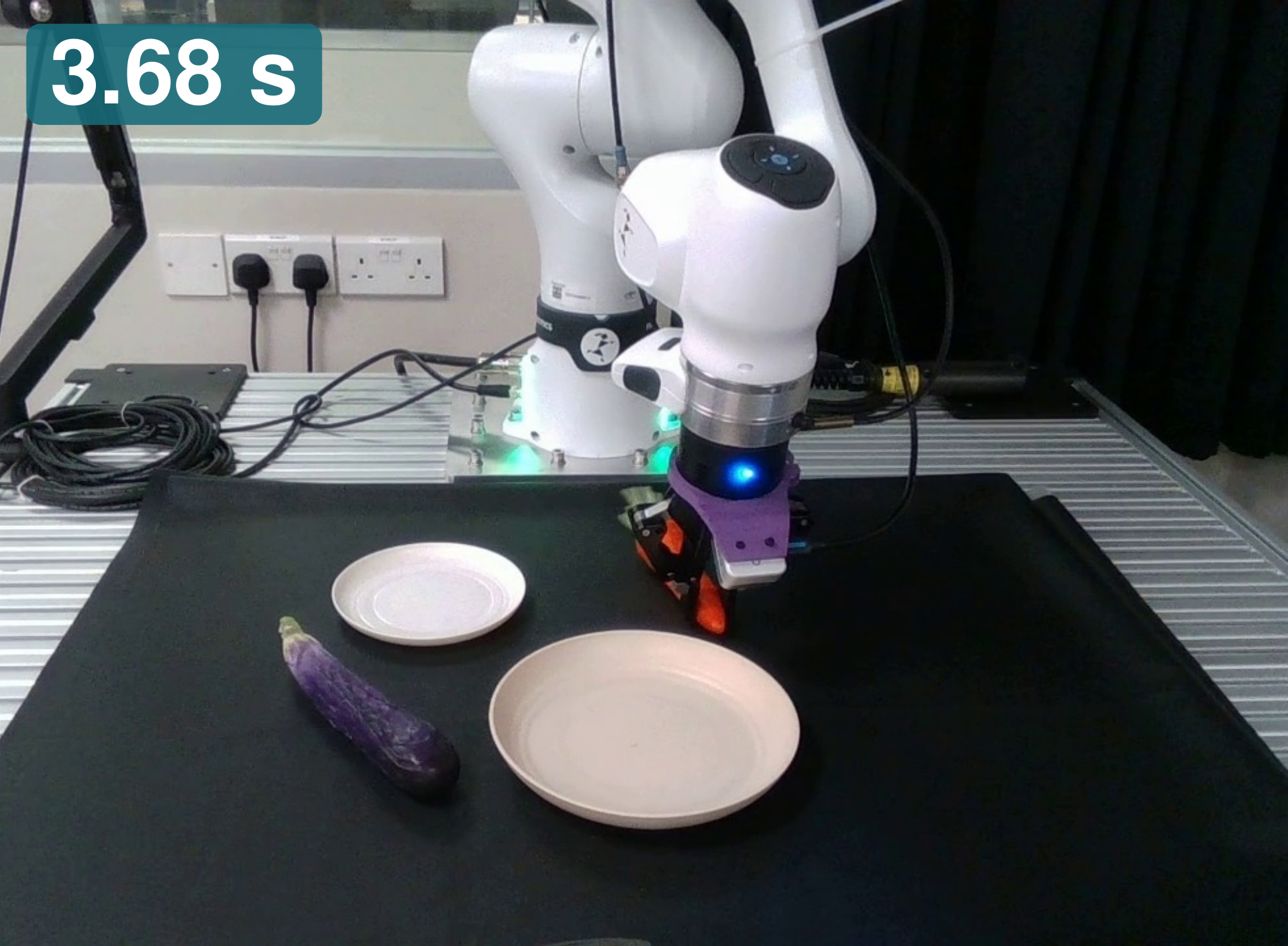}
\caption{Grasp 1.}\end{subfigure}\hfill
\begin{subfigure}{.16\textwidth}\centering
\includegraphics[width=\linewidth]{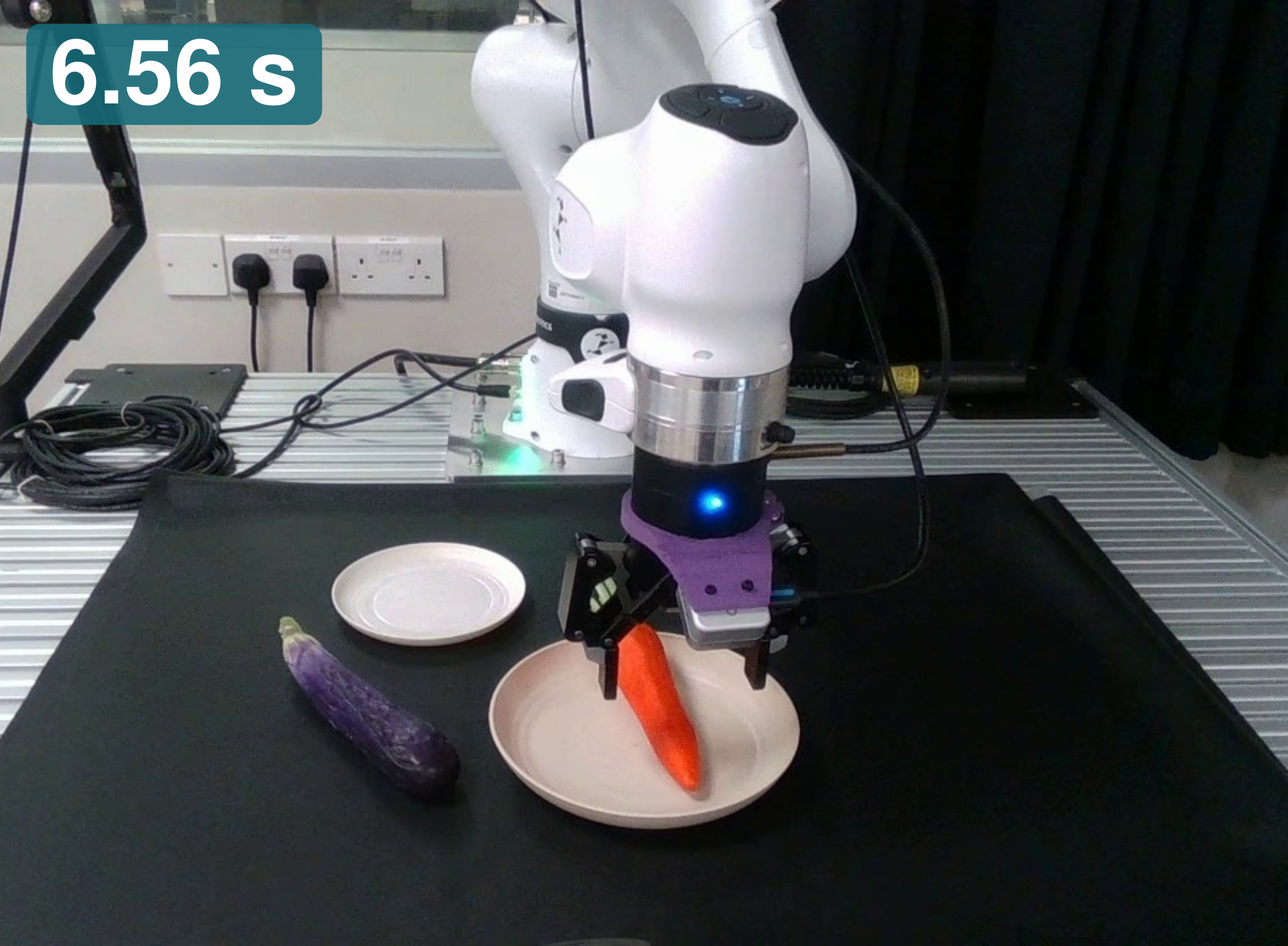}
\caption{Release 1.}\end{subfigure}\hfill
\begin{subfigure}{.16\textwidth}\centering
\includegraphics[width=\linewidth]{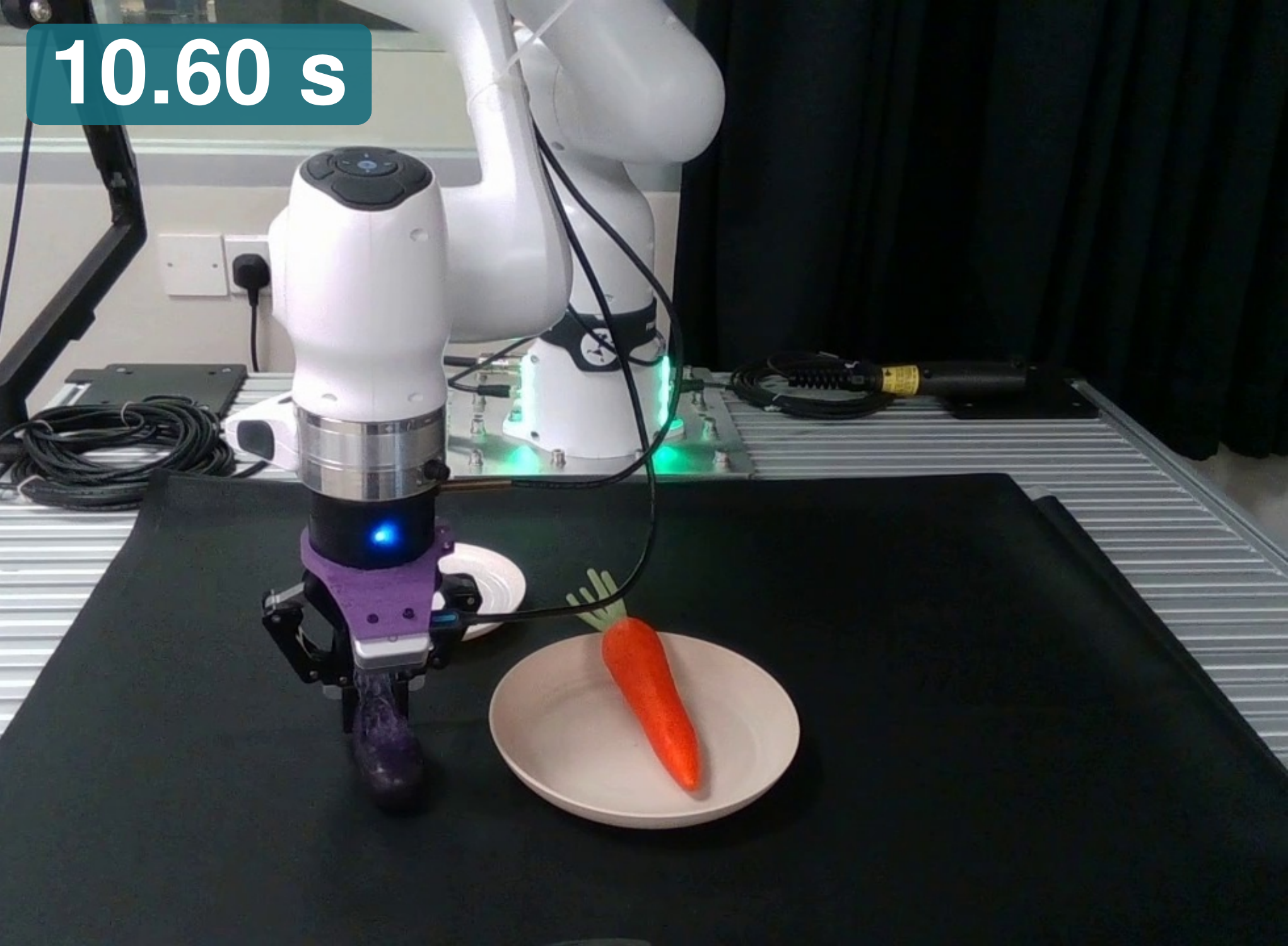}
\caption{Grasp 2.}\end{subfigure}\hfill
\begin{subfigure}{.16\textwidth}\centering
\includegraphics[width=\linewidth]{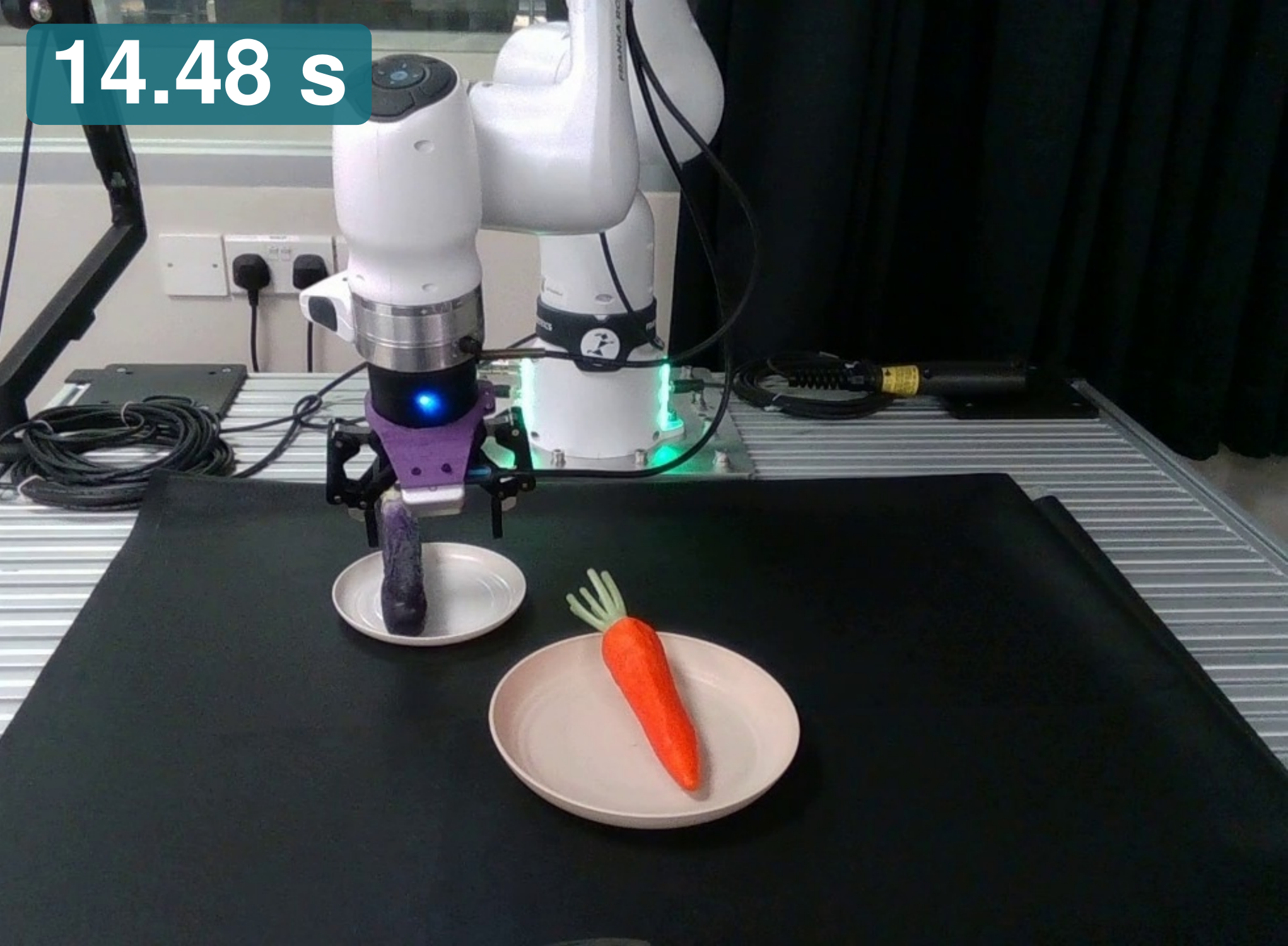}
\caption{Release 2.}\end{subfigure}\hfill
\begin{subfigure}{.16\textwidth}\centering
\includegraphics[width=\linewidth]{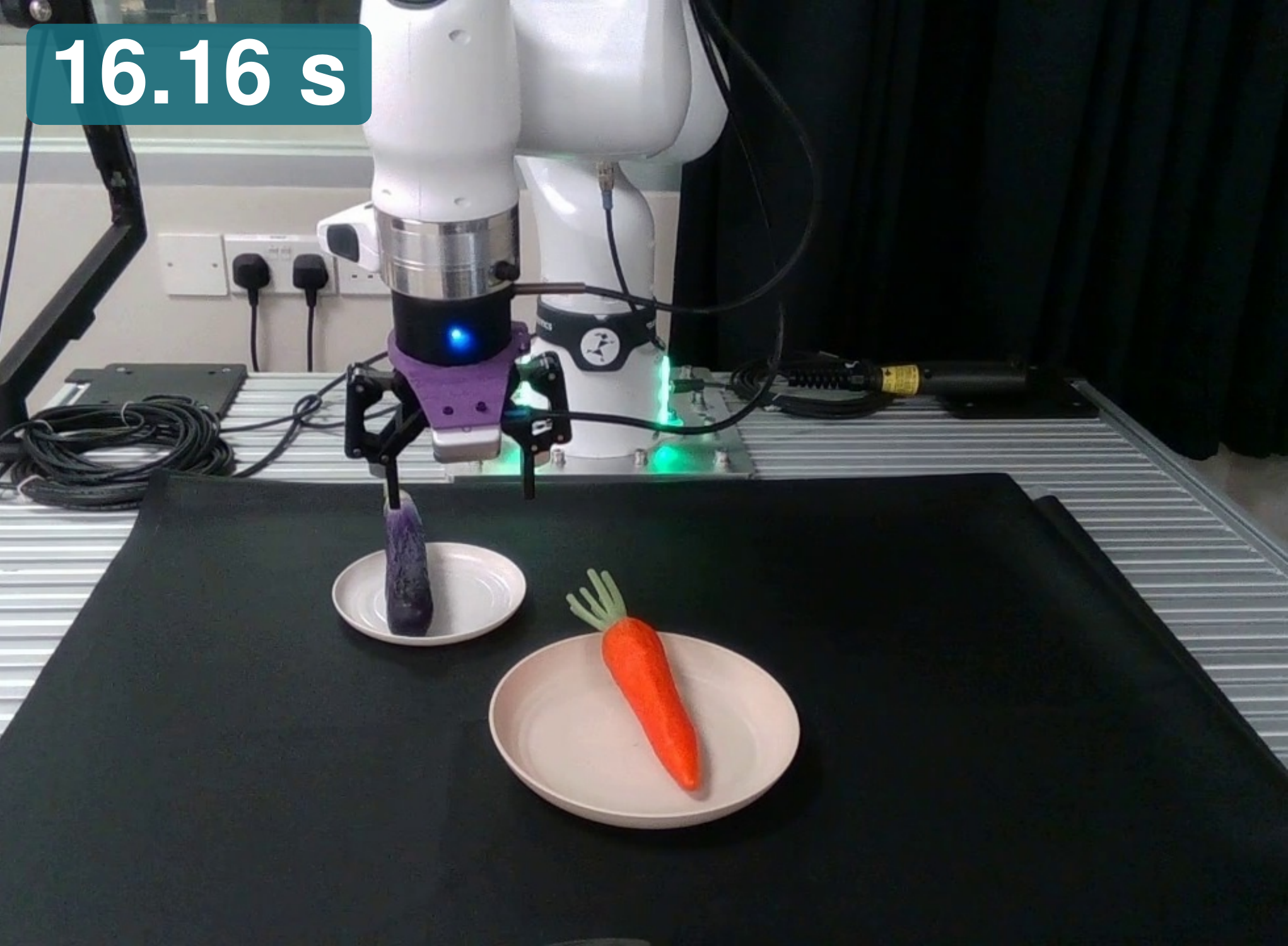}
\caption{Final.}\end{subfigure}\par\smallskip
\caption{\textbf{Vegetable sorting: matched sequence 5.} Prompt: \textit{``pick carrot on big plate and eggplant on small plate.''} Rows compare $\pi_{0.5}$ above and \method below across six event-defined stages under shared framing, with top-left badges indicating elapsed trial time.}
\label{fig:motion_pair_5}
\end{figure}

Figures~\ref{fig:motion_pair_1}--\ref{fig:motion_pair_5} align frames by six discrete event stages (initial frame, first grasp completion, first release completion, second grasp completion, second release completion, and final observation) under uniform framing within a 28\,ms frame-alignment tolerance, while Table~\ref{tab:motion_pairs} summarizes the corresponding unsmoothed active trajectory metrics over the complete motion segments.
\begin{table}[H]\centering\footnotesize
\caption{\textbf{Motion and response in the five matched physical pairs.} Entries compare $\pi_{0.5}$ / \method over complete unsmoothed active trajectories from the 138-trial cohort; Geometry reports the phase-matching RMS distance.}
\label{tab:motion_pairs}
\setlength{\tabcolsep}{3.5pt}
\begin{tabularx}{\textwidth}{lXrrrrr}\toprule
Pair & Task & Geometry (mm) & Speed CV $\downarrow$ & Accel.\ (m/s$^2$) $\downarrow$ & Replans/s $\uparrow$ & Response P50 (ms) $\downarrow$\\\midrule
1 & Cup stacking & 18.0 & 0.513 / \best{0.446} & 0.403 / \best{0.336} & 9.99 / \best{21.72} & 72.31 / \best{12.96}\\
2 & Cup stacking & 30.8 & 0.527 / \best{0.462} & 0.408 / \best{0.317} & 9.55 / \best{21.57} & 72.73 / \best{13.37}\\
3 & Fruit placement & 18.3 & 0.508 / \best{0.469} & 0.410 / \best{0.317} & 10.18 / \best{20.65} & 72.18 / \best{21.66}\\
4 & Fruit placement & 21.8 & 0.508 / \best{0.426} & 0.364 / \best{0.299} & 10.07 / \best{21.67} & 72.30 / \best{13.23}\\
5 & Vegetable sorting & 44.2 & 0.661 / \best{0.523} & 0.393 / \best{0.329} & 10.07 / \best{21.45} & 72.23 / \best{20.88}\\
\bottomrule\end{tabularx}\end{table}

\FloatBarrier
\subsection{Direct Pretrained-Controller Comparison}
\label{sec:pretrained_protocol}
All ten LIBERO-10 tasks are evaluated under official initial-state indices and seed~7 across 500 native trials (520-action budget) and 100 timed trials (20\,Hz dispatch pacing without catch-up under a 26\,s wall-clock ceiling), executing on dedicated RTX~5090 workers under Torch~2.8/CUDA~12.8. Timed response excludes the initial cold plan prepared before the timed loop and measures observation-to-first-action latency for adopted plans, reporting unweighted arithmetic means of per-episode response quantiles alongside distinct node dispatch and hold rates across all 100 timed episodes.

The comparison uses public pretrained weights with the inference configurations specified below. RTC~\citep{black2025rtc} uses ten denoising evaluations with exponential prefix weights, input-vector--Jacobian guidance capped at 5, and open-loop tail continuation while the next plan computes; under the precision-preserving compiled input-gradient backend, it reaches 459/500 native successes (91.80\%) and 93/100 timed successes (93.00\%, 0.00\% hold rate). Latent Bridge~\citep{liu2026latentbridge} uses the released LIBERO-10 R0 checkpoint with two denoising steps and a visual refresh every three plans, reaching 459/500 native successes or 91.80\% in Table~\ref{tab:pretrained_tasks}. \method uses the 12k-update checkpoint with one evaluation and $\rho=0.5$, while VLA-Rolling~\citep{he2026vlarolling} uses its released LeRobot backbone at $H/C=50/10$ with an adaptive budget of one to ten denoising steps.

\paragraph{RTC scheduling and latency coverage.}
RTC's published physical setting uses $H=50$, minimum execution horizon $s_{\min}=25$, five denoising steps, and a 50\,Hz controller~\citep{black2025rtc}. We retain the OpenPI LIBERO checkpoint's $H=10$, $s_{\min}=5$, and ten denoising steps used in the native comparison. Following Algorithm~1, the last ten observed delays are measured in actual controller steps, the next estimate is their maximum, and a completed chunk is published immediately with its elapsed prefix skipped. The estimate is never clipped to the remaining tail, and execution checks $d\leq s\leq H-d$. The policy and renderer run in separate processes on two exclusively allocated RTX~5090 GPUs, one for each role; the other timed controllers retain their previously reported co-located execution. The additional renderer GPU is part of the configuration in Table~\ref{tab:rtc_isolated_execution}; this result is therefore separate from the main single-GPU execution comparison. The change from the earlier run combines GPU isolation, process isolation, and scheduling repairs, so the latency reduction is not an isolated estimate of GPU contention alone.

\begin{table}[!htbp]\centering\footnotesize
\caption{\textbf{Per-task pretrained benchmark successes.} Native columns span 500 starts (50/task); paced columns span the shared 100-start timed assay (10/task). RTC uses separate policy/rendering GPUs; the other paced columns use one GPU. Paced highlighting compares only the single-GPU controllers.}
\label{tab:pretrained_tasks}
\setlength{\tabcolsep}{2pt}
\input{tables/pretrained_tasks.tex}
\end{table}

\begin{table}[!htbp]\centering\small
\caption{\textbf{RTC with isolated rendering.} The policy and renderer each occupy a separate, exclusively allocated RTX~5090. Results span the same 100 official starts at 20\,Hz with a 26\,s budget. This two-GPU configuration is reported separately from the main single-GPU execution table.}
\label{tab:rtc_isolated_execution}
\setlength{\tabcolsep}{7pt}
\begin{tabular}{lr}\toprule
Metric & RTC\\\midrule
Timed success (\%) & 93.00\\
Response P50 / P95 (ms) & 235.25 / 256.49\\
New plans/s & 3.89\\
Distinct action nodes/s & 20.01\\
Hold (\%) & 0.00\\
Mean plan generation (ms) & 208.03\\
\bottomrule\end{tabular}
\end{table}

Only the 100 timed starts were rerun after this scheduling repair; the native weights, outcomes, and independent inference-capacity measurements are unchanged. Mean per-plan closed-loop generation time is 208.03\,ms, compared with 283.33\,ms in the superseded co-located implementation. Mean per-episode response P50/P95 is 235.25/256.49\,ms, with 3.89 newly adopted plans/s, 20.01 distinct action nodes/s, and no holds. Rates use the recorded rollout interval, so endpoint counting can yield a value slightly above the nominal 20\,Hz. All 100 official initial states match the native subset; zero runtime errors occurred. The earlier 55/100 result arose under tail exhaustion and a clipped delay estimate and is retained as implementation-diagnostic evidence rather than the final RTC comparison.

\begin{table}[!htbp]\centering\small
\caption{\textbf{Published timing context.} Source-reported values retain their original hardware and timing definitions.}
\label{tab:published_execution}
\setlength{\tabcolsep}{7pt}
\begin{tabular}{llr}\toprule
Controller & Scope & Latency (ms) $\downarrow$\\\midrule
RTC~\citep{black2025rtc} & Inference & 76.00/97.00\\
Latent Bridge~\citep{liu2026latentbridge} & Control cycle & 46.00\\
EagleVLA~\citep{yang2026eaglevla} & Reaction & 61.10/97.10\\\bottomrule
\end{tabular}
\end{table}

\begin{table}[!htbp]\centering\footnotesize
\caption{\textbf{Additional dispatch and continuity metrics.} Nodes/s and Hold use the 100 timed runs in Table~\ref{tab:related_execution}; native jump measures unclipped boundary translation increments.}
\label{tab:pretrained_full}
\setlength{\tabcolsep}{3pt}
\input{tables/pretrained_baselines_dispatch.tex}
\end{table}

\paragraph{Scheduling controls and VLA-Rolling comparison.}
While Table~\ref{tab:deadline_detail} isolates observation-to-plan-ready scheduling on separate rendering GPUs, Table~\ref{tab:related_execution} measures end-to-end observation-to-first-dispatch latency with co-located rendering and policy inference on one GPU. RTC's separate-renderer result is retained in Table~\ref{tab:rtc_isolated_execution}. Across the completed native and timed comparisons in Table~\ref{tab:pretrained_full}, \method reduces unclipped native boundary translation jumps from 0.1013 for $\pi_{0.5}$ to 0.0771 and cuts timed holds from 34.86\% to 16.80\%. On the LeRobot $H/C=50/10$ base in Table~\ref{tab:rolling_control}, VLA-Rolling averages 6.19 native and 6.17 paced steps to lower paced response latency from 376.6 to 295.2\,ms, yielding 93.4\% native and 87.0\% paced success versus 93.8\% and 88.0\% for its matched baseline. Under the exclusive RTX~5090 resident-input capacity probe across 120 synchronized calls after 15 warmups, Latent Bridge R0 averages 63.06\,ms with 78.29\,ms P95, 15.86 plans/s, and 79.29 actions/s across full refresh cycles, whereas VLA-Rolling averages 166.47\,ms with 202.06\,ms P95, 6.01 plans/s, and 60.07 actions/s.

\begin{table}[!htbp]\centering\footnotesize
\caption{\textbf{Matched backbone control for VLA-Rolling.} Evaluated at $H/C=50/10$ on the released LeRobot base over 500 native and 100 timed starts.}
\label{tab:rolling_control}
\setlength{\tabcolsep}{7pt}
\begin{tabular}{lrrrr}\toprule
Controller & Native (\%) $\uparrow$ & Paced (\%) $\uparrow$ & P50 (ms) $\downarrow$ & Actions/s $\uparrow$\\\midrule
$\pi_{0.5}$ (LeRobot) & \best{93.8} & \best{88.0} & 376.6 & 12.06\\
\rowcolor{tablerow}VLA-Rolling & 93.4 & 87.0 & \best{295.2} & \best{13.20}\\\bottomrule
\end{tabular}
\end{table}

\FloatBarrier
\subsection{Visualizing Real and Predicted KV Memory}
\label{sec:kv_visualization}

\begin{figure}[H]\centering
\kvexampleimages{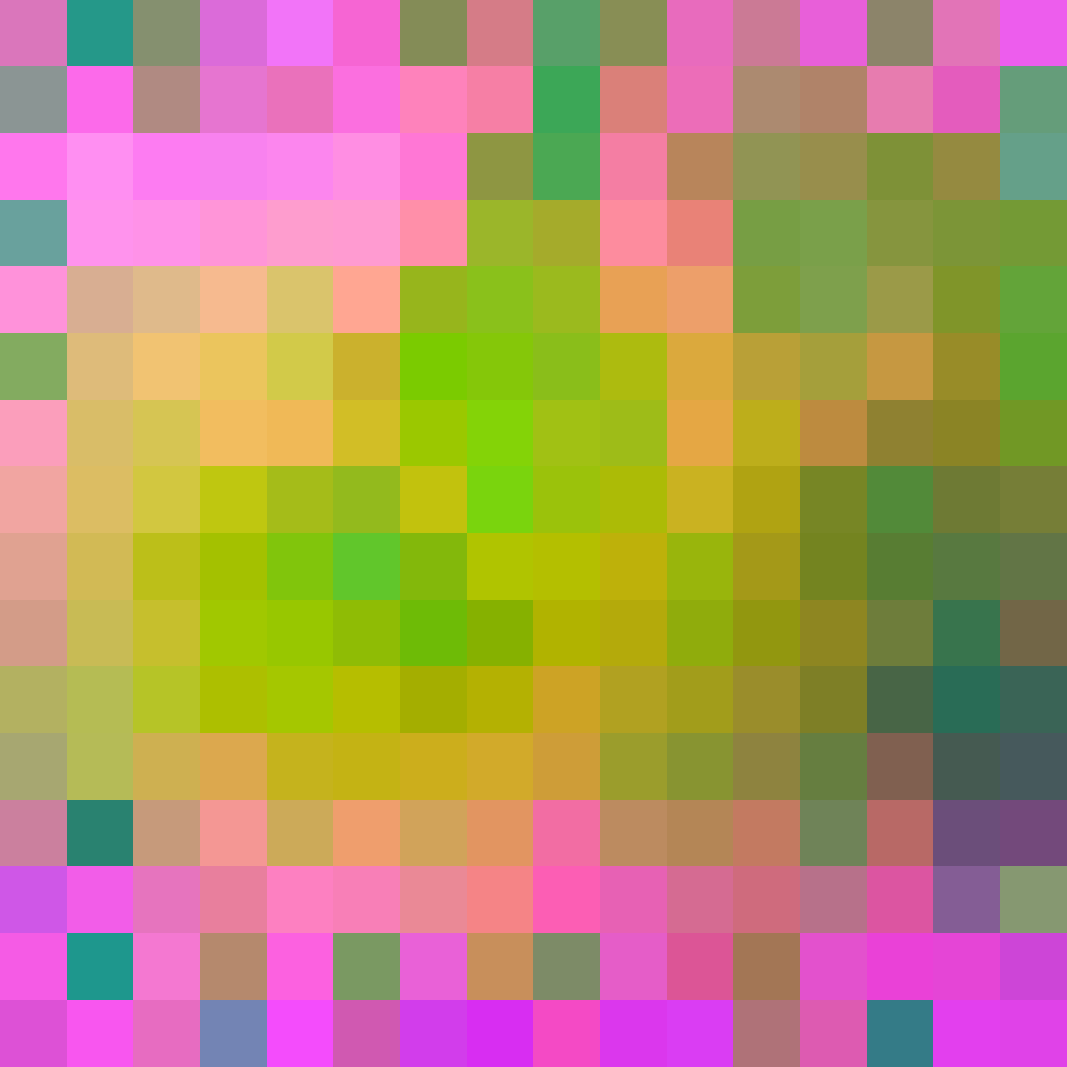}
\caption{\textbf{KV memory visualization: two-mug placement.} Episode 85, retained rows 95--100 (external camera above, wrist camera below). RGB frames retain unmodified camera pixels, whereas latent KV panels use the shared PCA projection defined in Appendix~\ref{sec:kv_visualization}.}
\label{fig:kv_mugs}
\end{figure}

\begin{figure}[H]\centering
\kvexampleimages{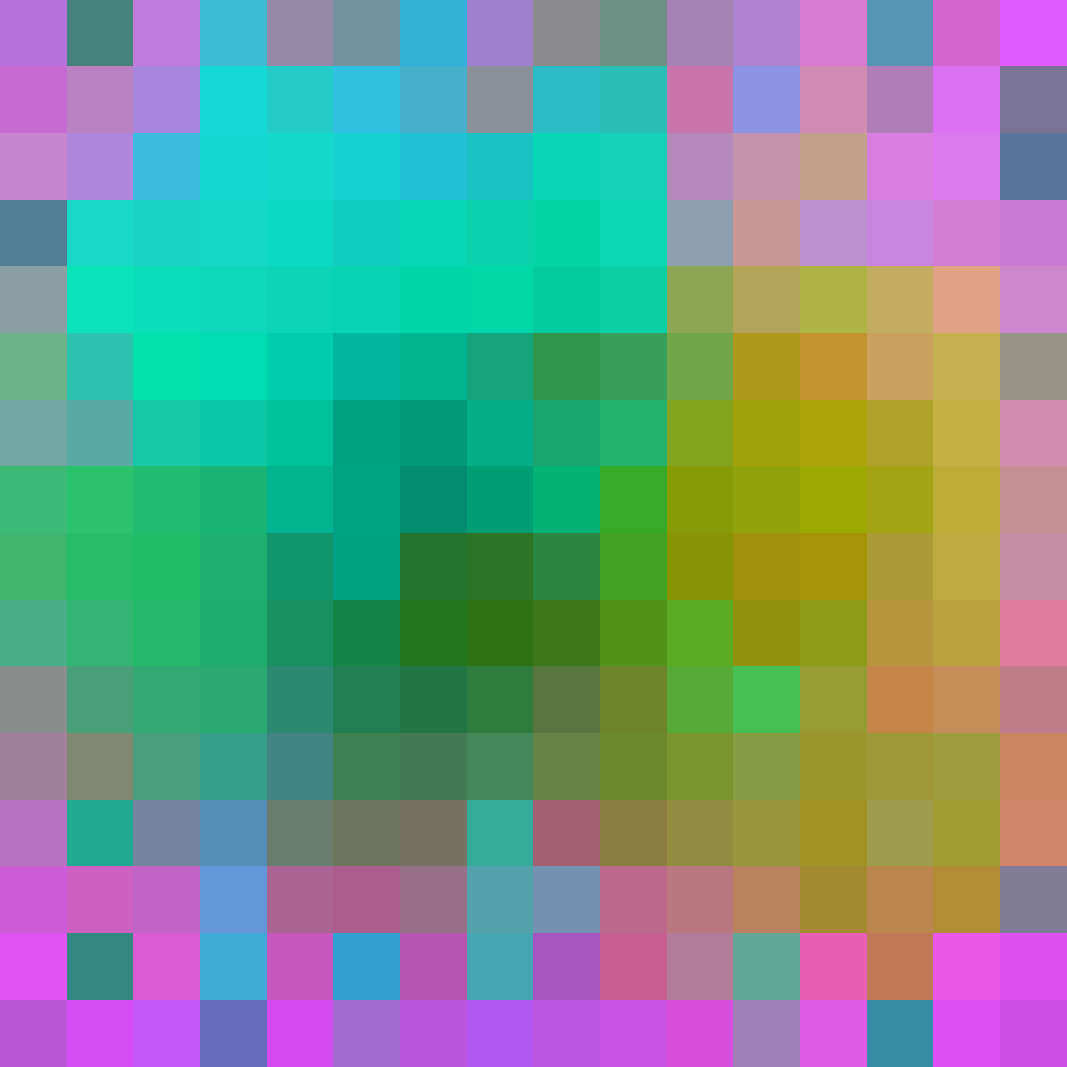}
\caption{\textbf{KV memory visualization: soup and cream cheese into basket.} Episode 148, retained rows 94--99 (external camera above, wrist camera below). Predicted KV memory updates local manipulator and object regions toward fresh KV memory.}
\label{fig:kv_food}
\end{figure}

\begin{figure}[H]\centering
\kvexampleimages{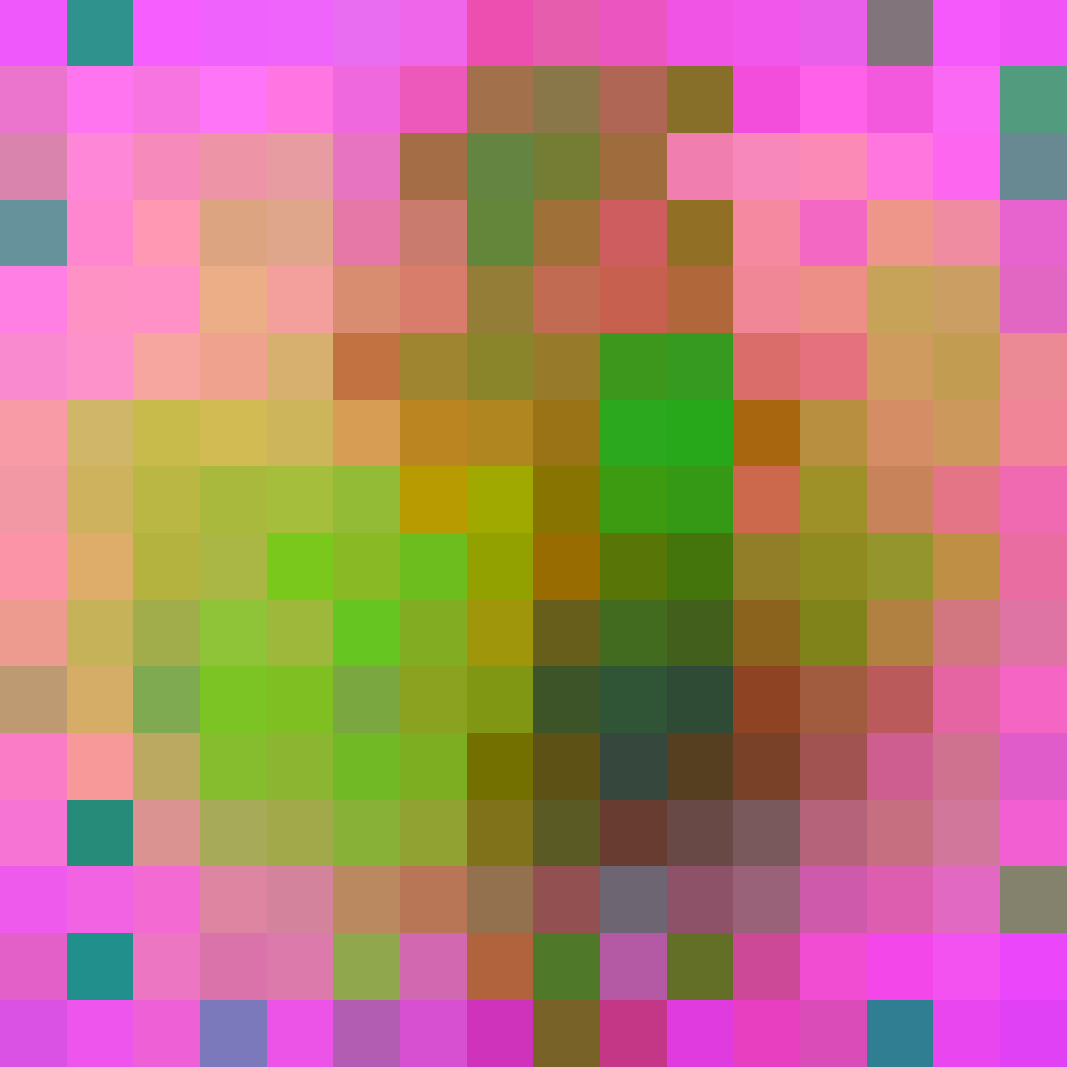}
\caption{\textbf{KV memory visualization: bowl into drawer.} Episode 15, retained rows 78--83 (external camera above, wrist camera below). Both cameras share the common PCA basis and percentile bounds fitted solely on real anchors and targets.}
\label{fig:kv_drawer}
\end{figure}

Figures~\ref{fig:kv_mugs}--\ref{fig:kv_drawer} compare the KV anchor, predicted KV memory, and fresh KV memory for three held-out LIBERO examples. We use the selected 12k updater and fix the examples before inference: the first held-out demonstration of tasks 0, 4 and 8, with an anchor approximately one third into each trajectory and a target five retained rows later. The history contains the intervening recorded states and executed actions. Retained demonstration rows need not correspond to every simulator tick.

For each visual token, we concatenate keys and values from all 18 layers into a 9,216-dimensional vector, normalized by the checkpoint's input KV scales. One PCA basis is fitted to the real anchors and real targets from all three examples and both cameras; predictions do not enter the fit. The first three components explain 25.56\% of the variance. Common component signs and 1st--99th percentile channel limits map them to RGB. Each camera's 256 tokens form a $16\times16$ grid, enlarged by nearest-neighbor replication. Colors thus depict latent features rather than camera RGB.

Table~\ref{tab:kv_examples} measures error to the target in the full visual KV space, using the training loss's delta normalization. Prediction reduces this error in each of the three examples. The images show where the predicted representation changes; the full-dimensional errors quantify its proximity to the target independently of the RGB projection.

\begin{table}[H]\centering\footnotesize
\caption{\textbf{KV prediction error on diagnostic examples.} Mean squared error over 512 visual tokens, both K/V tensors, all 18 layers, and all 256 channels using training delta normalization.}
\label{tab:kv_examples}
\setlength{\tabcolsep}{7pt}
\begin{tabular}{lrrr}\toprule
 & \multicolumn{2}{c}{KV error $\downarrow$} & \\
\cmidrule(lr){2-3}
Example & Stale & Predicted & Reduction (\%) $\uparrow$\\\midrule
Two mugs & 0.14273 & \best{0.13538} & 5.15\\
Soup and cream cheese & 0.32771 & \best{0.23517} & 28.24\\
Bowl into drawer & 0.37498 & \best{0.31968} & 14.75\\\bottomrule
\end{tabular}\end{table}

\end{document}

%% file: tables/pretrained_baselines.tex
\begin{tabular}{lrrrr}\toprule
Controller & Success (\%) $\uparrow$ & Response (ms) $\downarrow$ & Replans/s $\uparrow$ & Actions/s $\uparrow$\\\midrule
$\pi_{0.5}$ & 86.00 & \second{176.73}/189.97 & \second{2.57} & 13.02\\
Latent Bridge R0 & 82.00 & 183.88/\second{189.79} & 2.48 & 12.58\\
VLA-Rolling~\citep{he2026vlarolling} & \second{87.00} & 295.21/345.94 & 1.28 & \second{13.20}\\
\rowcolor{tablerow}\method & \best{91.00} & \best{83.04/90.53} & \best{3.28} & \best{16.60}\\
\bottomrule\end{tabular}

%% file: tables/claim_history_table.tex
\begin{tabular}{lrr}\toprule
History conditioning & KV error $\downarrow$ & Action MSE $\downarrow$\\\midrule
Stale anchor (w/o update) & \FinalStaleKV & \FinalStaleError\\
Age only & 0.37217 & 0.07221\\
Age + endpoint states & 0.35669 & 0.04602\\
Shuffled history & \second{0.32037} & \second{0.01107}\\\midrule
\rowcolor{tablerow}Full causal history & \best{0.31958} & \best{0.00983}\\
\bottomrule\end{tabular}

%% file: tables/pfws_comparison_table.tex
\begin{tabular}{lrrr}\toprule
Metric / policy & w/o PFWS & w/ PFWS & $\Delta$ (pp)\\\midrule
\multicolumn{4}{l}{\textit{Synchronous $\pi_{0.5}$, 1-step; 500 paired starts}}\\
Native success (\%) $\uparrow$ & 90.80 & \best{93.00} & $+2.20$\\
Boundary flips (\%) $\downarrow$ & 2.57 & \best{1.36} & $-1.22$\\
Interior flips (\%) $\downarrow$ & 1.69 & \best{1.39} & $-0.30$\\
Boundary excess (pp; $\to0$) & +0.88 & \best{\textminus0.03} & $-0.91$\\
\midrule\multicolumn{4}{l}{\textit{LIBERO-Plus success (\%); 2,519 paired variants}}\\
$\pi_{0.5}$, 1-step & 78.36 & \best{80.15} & $+1.79$\\
\rowcolor{tablerow}\method, 1-step & 76.38 & \best{78.09} & $+1.71$\\
\bottomrule\end{tabular}

%% file: tables/fr3_final_table.tex
\begin{tabularx}{\textwidth}{llr*{2}{>{\raggedleft\arraybackslash}p{41pt}}*{2}{>{\raggedleft\arraybackslash}X}>{\raggedleft\arraybackslash}p{39pt}>{\raggedleft\arraybackslash}p{45pt}r}\toprule
 & & \multicolumn{1}{r}{Success} & \multicolumn{2}{c}{Active dispatch (Hz) $\uparrow$} & \multicolumn{2}{c}{Response (ms) $\downarrow$} & \multicolumn{2}{c}{EEF motion $\downarrow$} & \multicolumn{1}{r}{Time}\\
\cmidrule(lr){4-5}\cmidrule(lr){6-7}\cmidrule(lr){8-9}
Task & Controller & (\%) $\uparrow$ & Actions & Plans & P50 & P95 & Speed CV & Accel.\ (m/s$^2$) & (s) $\downarrow$\\\midrule
Cup stacking & $\pi_{0.5}$ & 73.91 & 20.00 & 9.98 & 72.34 & 75.03 & 0.530 & 0.390 & 28.97\\
\rowcolor{tablerow} & \method & \best{82.61} & \best{43.22} & \best{21.67} & \best{13.30} & \best{25.50} & \best{0.453} & \best{0.289} & \best{19.63}\\
\midrule
Fruit placement & $\pi_{0.5}$ & 69.57 & 20.08 & 10.07 & 72.30 & 73.42 & 0.487 & 0.348 & 22.87\\
\rowcolor{tablerow} & \method & \best{95.65} & \best{42.41} & \best{21.27} & \best{14.28} & \best{31.75} & \best{0.433} & \best{0.289} & \best{16.62}\\
\midrule
Vegetable sorting & $\pi_{0.5}$ & 73.91 & 20.04 & 10.03 & 72.26 & 73.28 & 0.659 & 0.405 & 23.50\\
\rowcolor{tablerow} & \method & \best{82.61} & \best{43.26} & \best{21.66} & \best{12.82} & \best{23.54} & \best{0.525} & \best{0.332} & \best{17.75}\\
\midrule
All tasks & $\pi_{0.5}$ & 72.46 & 20.06 & 10.03 & 72.30 & 73.47 & 0.531 & 0.387 & 25.16\\
\rowcolor{tablerow} & \method & \best{86.96} & \best{43.13} & \best{21.60} & \best{13.17} & \best{25.53} & \best{0.462} & \best{0.296} & \best{17.93}\\
\bottomrule\end{tabularx}

%% file: tables/final_by_task.tex
\begingroup
\setlength{\tabcolsep}{7pt}
\begin{tabular}{lrr}\toprule
Task & 10-step $\pi_{0.5}$ & \method\\\midrule
Soup / tomato & 46 & \best{49}\\
Cheese / butter & \best{50} & \best{50}\\
Stove + moka pot & 47 & \best{49}\\
Bowl / drawer & \best{50} & 47\\
Two mugs & \best{48} & \best{48}\\
Book / caddy & 43 & \best{44}\\
Mug / pudding & 45 & \best{46}\\
Soup / cheese & 49 & \best{50}\\
Two moka pots & 29 & \best{38}\\
Mug / microwave & \best{49} & 48\\
\midrule\rowcolor{tablerow} Total (/500) & 456 & \best{469}\\\bottomrule\end{tabular}
\endgroup

%% file: tables/final_validation_table.tex
\begingroup
\setlength{\tabcolsep}{7pt}
\begin{tabular}{lrrr}\toprule
Metric & 6k & 12k & 18k\\\midrule
Weighted action MSE & 0.00807 & \second{0.00686} & \best{0.00648}\\
Unweighted action MSE & 0.01158 & \best{0.00983} & \second{0.01033}\\
Gripper MSE & \second{0.02736} & \best{0.02308} & 0.02993\\
\bottomrule\end{tabular}
\endgroup

%% file: tables/final_module_tasks.tex
\begingroup
\setlength{\tabcolsep}{7pt}
\begin{tabular}{lrrrr}\toprule
Task & w/o latent WM & w/o PFWS & Sync. PFWS & \method\\\midrule
0 & 42 & 47 & 48 & \best{49}\\
1 & 46 & \best{50} & \best{50} & 49\\
2 & 29 & \best{50} & 49 & 48\\
3 & 48 & \best{50} & 48 & 48\\
4 & 44 & \best{50} & 49 & \best{50}\\
5 & 45 & 43 & 44 & \best{46}\\
6 & 40 & 44 & \best{46} & 45\\
7 & 48 & 48 & \best{50} & \best{50}\\
8 & 22 & 31 & 31 & \best{34}\\
9 & 37 & 47 & 45 & \best{48}\\
\midrule\rowcolor{tablerow} Total & 401 & 460 & 460 & \best{467}\\\bottomrule\end{tabular}
\endgroup

%% file: tables/final_age_table.tex
\begingroup
\setlength{\tabcolsep}{7pt}
\begin{tabular}{lrrrrr}\toprule
 & & \multicolumn{2}{c}{KV error $\downarrow$} & \multicolumn{2}{c}{Action MSE $\downarrow$}\\
Age & Pairs & Stale & Updated & Stale & Updated\\\midrule
1--4 & 105 & 0.1553 & \best{0.1356} & 0.0171 & \best{0.0025}\\
5--9 & 126 & 0.3953 & \best{0.3116} & 0.0702 & \best{0.0072}\\
10--19 & 40 & 0.5856 & \best{0.4400} & 0.1125 & \best{0.0121}\\
20--29 & 21 & 0.9364 & \best{0.6357} & 0.3208 & \best{0.0284}\\
30--39 & 21 & 0.9037 & \best{0.6083} & 0.3253 & \best{0.0373}\\
40--45 & 7 & 1.0858 & \best{0.7205} & 0.6330 & \best{0.0161}\\
\bottomrule\end{tabular}
\endgroup

%% file: tables/final_plus_table.tex
\textbf{Frozen policy: synchronous inference}\par\smallskip
\begin{tabularx}{\textwidth}{Xrrrrr}\toprule
Perturbation & Episodes & Cold (1 NFE) & Cold (10 NFE) & PFWS (1 NFE) & PFWS (10 NFE)\\\midrule
Background Textures & 289 & 87.89\% (254) & \best{88.93}\% (257) & \best{90.66}\% (262) & 88.58\% (256)\\
Camera Viewpoints & 419 & \best{55.85}\% (234) & \best{54.89}\% (230) & 54.18\% (227) & 54.42\% (228)\\
Language Instructions & 383 & \best{90.08}\% (345) & \best{90.86}\% (348) & 89.03\% (341) & 89.56\% (343)\\
Light Conditions & 274 & 93.43\% (256) & 90.51\% (248) & \best{93.80}\% (257) & \best{93.43}\% (256)\\
Objects Layout & 312 & 86.54\% (270) & \best{86.54}\% (270) & \best{89.10}\% (278) & 83.65\% (261)\\
Robot Initial States & 393 & 73.03\% (287) & 75.32\% (296) & \best{75.57}\% (297) & \best{78.63}\% (309)\\
Sensor Noise & 449 & 73.05\% (328) & 77.51\% (348) & \best{79.51}\% (357) & \best{78.40}\% (352)\\
\midrule\rowcolor{tablerow}Overall & 2,519 & 78.36\% (1,974) & 79.28\% (1,997) & \best{80.15}\% (2,019) & \best{79.60}\% (2,005)\\\bottomrule\end{tabularx}
\par\medskip\textbf{\method: fixed world model, 1 NFE}\par\smallskip
\begin{tabularx}{\textwidth}{Xrrrr}\toprule
Perturbation & Episodes & Without PFWS & With PFWS & Change (pp)\\\midrule
Background Textures & 289 & 91.70\% (265) & \best{93.77}\% (271) & +2.08\\
Camera Viewpoints & 419 & \best{51.07}\% (214) & 49.88\% (209) & -1.19\\
Language Instructions & 383 & 86.95\% (333) & \best{87.99}\% (337) & +1.04\\
Light Conditions & 274 & 84.31\% (231) & \best{88.32}\% (242) & +4.01\\
Objects Layout & 312 & \best{81.09}\% (253) & 79.17\% (247) & -1.92\\
Robot Initial States & 393 & 73.54\% (289) & \best{78.12}\% (307) & +4.58\\
Sensor Noise & 449 & 75.50\% (339) & \best{78.84}\% (354) & +3.34\\
\midrule\rowcolor{tablerow}Overall & 2,519 & 76.38\% (1,924) & \best{78.09}\% (1,967) & +1.71\\\bottomrule\end{tabularx}

%% file: tables/final_timing_table.tex
\begingroup
\setlength{\tabcolsep}{7pt}
\begin{tabular}{llrrr}\toprule
$C$ & Operation & Mean $\downarrow$ & P95 $\downarrow$ & Actions/s $\uparrow$\\\midrule
5 & $\pi_{0.5}$, 10-step & 60.01 & 60.20 & 83.32\\
5 & $\pi_{0.5}$, 1-step & 36.00 & 36.32 & 138.88\\
5 & First cold plan & 35.39 & 35.66 & 141.27\\
5 & \method, isolated & \best{4.79} & \best{4.83} & \best{1043.02}\\
\rowcolor{tablerow}5 & \method, shared & \second{11.91} & \second{13.38} & \second{419.82}\\
\midrule
1 & $\pi_{0.5}$, 10-step & 59.96 & 60.25 & 16.68\\
1 & $\pi_{0.5}$, 1-step & 36.15 & 36.47 & 27.67\\
1 & First cold plan & 35.55 & 35.81 & 28.13\\
1 & \method, isolated & \best{4.78} & \best{4.81} & \best{209.32}\\
\rowcolor{tablerow}1 & \method, shared & \second{11.97} & \second{13.53} & \second{83.51}\\
\bottomrule\end{tabular}
\endgroup

%% file: tables/final_deadline_detail.tex
\begin{tabularx}{\textwidth}{rrXrrrrr}\toprule
Hz & $C$ & Controller / miss rule & Solved $\uparrow$ & Ready (\%) $\uparrow$ & Hold (\%) $\downarrow$ & Capped time (s) $\downarrow$ & Overrun ticks $\downarrow$\\\midrule
20 & 1 & $\pi_{0.5}$, 10-step / hold & 2/10 & 0.00 & 52.05 & 25.70 & 3/5132\\
20 & 1 & $\pi_{0.5}$, 1-step / hold & 0/10 & 0.00 & 50.10 & 26.00 & 0/5190\\
20 & 1 & $\pi_{0.5}$, 10-step / tail & 10/10 & 0.00 & 0.00 & 13.66 & 0/2726\\
20 & 1 & $\pi_{0.5}$, 1-step / tail & 10/10 & 0.00 & 0.00 & 13.30 & 0/2657\\
\rowcolor{tablerow}20 & 1 & \method / hold & 9/10 & 83.92 & 15.40 & 15.35 & 56/3064\\
\midrule
50 & 1 & $\pi_{0.5}$, 10-step / hold & 0/10 & 0.00 & 83.34 & 26.00 & 1/12990\\
50 & 1 & $\pi_{0.5}$, 1-step / hold & 0/10 & 0.00 & 70.02 & 26.00 & 11/12990\\
50 & 1 & $\pi_{0.5}$, 10-step / tail & 4/10 & 0.00 & 15.41 & 21.07 & 2/10528\\
50 & 1 & $\pi_{0.5}$, 1-step / tail & 10/10 & 0.00 & 0.00 & 10.19 & 0/5091\\
\rowcolor{tablerow}50 & 1 & \method / hold & 3/10 & 0.22 & 60.00 & 25.29 & 20/12634\\
\midrule
20 & 5 & $\pi_{0.5}$, 10-step / hold & 9/10 & 0.00 & 25.31 & 16.73 & 4/3342\\
20 & 5 & $\pi_{0.5}$, 1-step / hold & 10/10 & 0.00 & 16.53 & 14.75 & 0/2946\\
20 & 5 & $\pi_{0.5}$, 10-step / tail & 9/10 & 0.00 & 0.00 & 14.00 & 0/2796\\
20 & 5 & $\pi_{0.5}$, 1-step / tail & 10/10 & 0.00 & 0.00 & 11.99 & 0/2394\\
\rowcolor{tablerow}20 & 5 & \method / hold & 9/10 & 54.63 & 9.17 & 14.75 & 54/2944\\
\midrule
50 & 5 & $\pi_{0.5}$, 10-step / hold & 6/10 & 0.00 & 51.82 & 22.45 & 2/11215\\
50 & 5 & $\pi_{0.5}$, 1-step / hold & 8/10 & 0.00 & 37.24 & 17.23 & 0/8611\\
50 & 5 & $\pi_{0.5}$, 10-step / tail & 0/10 & 0.00 & 49.58 & 26.00 & 0/12990\\
50 & 5 & $\pi_{0.5}$, 1-step / tail & 6/10 & 0.00 & 11.59 & 18.93 & 0/9456\\
\rowcolor{tablerow}50 & 5 & \method / hold & 6/10 & 2.34 & 25.41 & 19.54 & 234/9764\\
\midrule
\multicolumn{8}{l}{Additional-backbone references: separate PRO6000 policy/rendering GPUs}\\
20 & 5 & GR00T N1.7 / hold & 10/10 & 0.00 & 16.57 & 14.76 & 0/2951\\
20 & 5 & GR00T N1.7 / tail & 10/10 & 0.00 & 0.00 & 12.38 & 0/2474\\
20 & 5 & OpenVLA-OFT / hold & 9/10 & 0.00 & 16.57 & 15.01 & 0/2999\\
20 & 5 & OpenVLA-OFT / tail & 10/10 & 0.00 & 0.00 & 12.51 & 0/2501\\
\bottomrule\end{tabularx}

%% file: tables/fr3_final_timing.tex
\begin{tabularx}{\textwidth}{ll*{4}{>{\raggedleft\arraybackslash}X}*{2}{>{\raggedleft\arraybackslash}p{53pt}}}\toprule
& & \multicolumn{2}{c}{New actions (Hz) $\uparrow$} & \multicolumn{2}{c}{New plans (Hz) $\uparrow$} & \multicolumn{2}{c}{First-dispatch latency (ms) $\downarrow$}\\
\cmidrule(lr){3-4}\cmidrule(lr){5-6}\cmidrule(l){7-8}
Task & Controller & Active & Wall clock & Active & Wall clock & P50 & P95\\\midrule
Cup stacking & $\pi_{0.5}$ & 20.00 & 17.46 & 9.98 & 8.77 & 72.34 & 75.03\\
\rowcolor{tablerow} & \method & \best{43.22} & \best{35.54} & \best{21.67} & \best{17.80} & \best{13.30} & \best{25.50}\\
\midrule
Fruit placement & $\pi_{0.5}$ & 20.08 & 18.33 & 10.07 & 9.16 & 72.30 & 73.42\\
\rowcolor{tablerow} & \method & \best{42.41} & \best{37.41} & \best{21.27} & \best{18.71} & \best{14.28} & \best{31.75}\\
\midrule
Vegetable sorting & $\pi_{0.5}$ & 20.04 & 17.52 & 10.03 & 8.77 & 72.26 & 73.28\\
\rowcolor{tablerow} & \method & \best{43.26} & \best{35.19} & \best{21.66} & \best{17.63} & \best{12.82} & \best{23.54}\\
\bottomrule\end{tabularx}

%% file: tables/fr3_final_windows.tex
\begingroup
\setlength{\tabcolsep}{7pt}
\begin{tabular}{lrrr}\toprule
Task & Windows & $\geq2$ fast plans & Fraction\\\midrule
Cup stacking & 5,574 & 2,009 & 36.04\%\\
Fruit placement & 4,723 & 1,569 & 33.22\%\\
Vegetable sorting & 4,502 & 1,602 & 35.58\%\\
\midrule
\rowcolor{tablerow}All tasks & 14,799 & 5,180 & 35.00\%\\
\bottomrule\end{tabular}
\endgroup

%% file: tables/fr3_selected_pair_table.tex
\begingroup
\setlength{\tabcolsep}{7pt}
\begin{tabular}{lrr}\toprule
Metric & $\pi_{0.5}$ & \method\\\midrule
Successful-trial time (s) $\downarrow$ & 28.43 & \best{20.37}\\
Speed CV $\downarrow$ & 0.531 & \best{0.502}\\
Acceleration RMS (m/s$^2$) $\downarrow$ & 0.370 & \best{0.291}\\
New plans (Hz) $\uparrow$ & 10.20 & \best{21.59}\\
New action nodes (Hz) $\uparrow$ & 20.38 & \best{43.14}\\
Response P50 (ms) $\downarrow$ & 72.16 & \best{13.50}\\
Response P95 (ms) $\downarrow$ & 73.84 & \best{23.26}\\
\bottomrule\end{tabular}
\endgroup

%% file: tables/pretrained_tasks.tex
\begingroup
\setlength{\tabcolsep}{7pt}
\begin{tabular}{lcccccc}\toprule
 & \multicolumn{2}{c}{Native (50/task)} & \multicolumn{4}{c}{Paced (10/task)}\\
\cmidrule(lr){2-3}\cmidrule(l){4-7}
Task & RTC & LB R0 & $\pi_{0.5}$ & RTC & LB R0 & \method\\\midrule
0 & \best{48} & 46 & \best{10} & 10 & \best{10} & 9\\
1 & 49 & \best{50} & \best{10} & 10 & \best{10} & \best{10}\\
2 & \best{50} & 46 & \best{10} & 10 & \best{10} & \best{10}\\
3 & \best{49} & \best{49} & \best{10} & 8 & 7 & 9\\
4 & \best{48} & 45 & \best{10} & 9 & 8 & \best{10}\\
5 & 43 & \best{46} & \best{10} & 10 & \best{10} & \best{10}\\
6 & 43 & \best{45} & \best{10} & 10 & 8 & 9\\
7 & \best{50} & 49 & 9 & 10 & \best{10} & \best{10}\\
8 & 32 & \best{41} & 0 & 8 & 0 & \best{5}\\
9 & \best{47} & 42 & 7 & 8 & \best{9} & \best{9}\\
\midrule\rowcolor{tablerow} Total & \best{459} & \best{459} & 86 & 93 & 82 & \best{91}\\\bottomrule\end{tabular}
\endgroup

%% file: tables/pretrained_baselines_dispatch.tex
\begingroup
\setlength{\tabcolsep}{7pt}
\begin{tabular}{lrrr}\toprule
Controller & Nodes/s $\uparrow$ & Hold (\%) $\downarrow$ & Native jump $\downarrow$\\\midrule
$\pi_{0.5}$ & \second{13.02} & \second{34.86} & \second{0.10}\\
Latent Bridge R0 & 12.58 & 37.08 & 0.13\\
\rowcolor{tablerow}\method & \best{16.60} & \best{16.80} & \best{0.08}\\
\bottomrule\end{tabular}
\endgroup